\documentclass{article}

\usepackage[main, final]{neurips_2026}
\usepackage{inconsolata}

\usepackage[utf8]{inputenc}
\usepackage[T1]{fontenc}
\usepackage{xcolor}
\definecolor{darkblue}{RGB}{0,32,96}
\usepackage{hyperref}
\hypersetup{
    colorlinks=true,
    linkcolor=darkblue,
    citecolor=darkblue,
    urlcolor=darkblue,
}
\usepackage{url}
\usepackage{booktabs}
\usepackage{longtable}
\usepackage{makecell}
\usepackage{amsfonts}
\usepackage{nicefrac}
\usepackage{microtype}
\usepackage[flushmargin,hang]{footmisc}

\usepackage{graphicx}
\usepackage{wrapfig}
\usepackage{amsthm}
\usepackage{caption}
\usepackage{amsmath}
\usepackage[capitalize,noabbrev]{cleveref}
\usepackage{mleftright}
\usepackage{algorithm}
\usepackage[noend]{algorithmic}

\usepackage[inline]{enumitem}
\usepackage{etoc}
\setitemize{noitemsep,topsep=0pt,parsep=0pt,partopsep=0pt,leftmargin=*}
\setenumerate{noitemsep,topsep=0pt,parsep=0pt,partopsep=0pt,leftmargin=*}
\makeatletter
\renewcommand{\paragraph}{%
  \@startsection{paragraph}{4}{\z@}%
                {0.25ex \@plus 0.1ex \@minus 0.1ex}%
                {-1em}%
                {\normalsize\bf}%
}
\makeatother
\setcitestyle{round}
\usepackage{xspace}
\usepackage{amsfonts}
\usepackage{bbm}
\usepackage{amssymb}

\newtheoremstyle{plaintight}%
{\topsep}
{0pt}
{\itshape}
{}
{\bfseries}
{.}
{ }
{}
\theoremstyle{plaintight}     
\newtheorem{theorem}{Theorem}[section]

\newtheorem{remark}[theorem]{Remark}

\newcommand{\deferredproof}[1]{\noindent{\small $\blacktriangleright$}~Proof in \cref{#1}.}

\crefname{section}{\S}{\S\S}
\Crefname{section}{\S}{\S\S}
\crefname{table}{Tab.}{}
\crefname{figure}{Fig.}{Figs.}
\crefname{equation}{Eq.}{}
\crefname{appendix}{App.}{}
\crefname{thm}{Theorem}{}
\crefname{prop}{Proposition}{}
\crefname{defin}{Definition}{}
\crefname{cor}{Corollary}{}
\crefname{observation}{Observation}{}
\crefname{assumption}{Assumption}{}
\crefformat{section}{\S#2#1#3}
\crefname{footnote}{footnote}{footnotes}
\crefname{algorithm}{Alg.}{}

\definecolor{NavyBlue}{rgb}{0.0, 0.0, 0.5}
\definecolor{OliveGreen}{rgb}{0.33, 0.42, 0.18}
\definecolor{WildStrawberry}{rgb}{1.0, 0.26, 0.64}
\definecolor{Plum}{rgb}{0.56, 0.27, 0.52}
\definecolor{RoyalPurple}{rgb}{0.47, 0.32, 0.66}

\usepackage{mathtools}
\usepackage{ellipsis}
\makeatletter
\renewcommand*{\mathellipsis}{%
  \mathinner{%
    \kern\ellipsisbeforegap%
    {\ldotp}\kern\ellipsisgap%
    {\ldotp}\kern\ellipsisgap%
    {\ldotp}\kern\ellipsisaftergap%
  }%
}
\renewcommand*{\dotsb@}{%
  \mathinner{%
    \kern\ellipsisbeforegap%
    {\cdotp}\kern\ellipsisgap%
    {\cdotp}\kern\ellipsisgap%
    {\cdotp}\kern\ellipsisaftergap%
  }%
}
\renewcommand*{\@cdots}{%
  \mathinner{%
    \kern\ellipsisbeforegap%
    {\cdotp}\kern\ellipsisgap%
    {\cdotp}\kern\ellipsisgap%
    {\cdotp}\kern\ellipsisaftergap%
  }%
}
\renewcommand*{\ellipsis@default}{%
  \ellipsis@before
  \kern\ellipsisbeforegap
  .\kern\ellipsisgap
  .\kern\ellipsisgap
  .\kern\ellipsisgap
  \ellipsis@after\relax}
\renewcommand*{\ellipsis@centered}{%
  \ellipsis@before
  \kern\ellipsisbeforegap
  .\kern\ellipsisgap
  .\kern\ellipsisgap
  .\kern\ellipsisaftergap
  \ellipsis@after\relax}
\AtBeginDocument{%
  \DeclareRobustCommand*{\dots}{%
    \ifmmode\@xp\mdots@\else\@xp\textellipsis\fi}}
\def\ellipsisgap{-.05em}
\def\ellipsisbeforegap{-.05em}
\def\ellipsisaftergap{.05em}
\makeatother

\definecolor{ETHBlue}{RGB}{33,92,175}
\definecolor{ETHGreen}{RGB}{98,115,19}
\definecolor{ETHPurpleDark}{RGB}{140,10,89}
\definecolor{ETHPurple}{RGB}{163,7,116}
\definecolor{ETHGray}{RGB}{111,111,111}
\definecolor{ETHRed}{RGB}{183,53,45}
\definecolor{ETHPetrol}{RGB}{0,120,148}
\definecolor{ETHBronze}{RGB}{142,103,19}

\colorlet{ETHdarkblue}{ETHBlue!80!black}
\colorlet{ETHdarkgreen}{ETHGreen!80!black}
\colorlet{ETHpink}{ETHPurple}
\colorlet{ETHgray}{ETHGray}
\colorlet{ETHred}{ETHRed}
\colorlet{ETHgreenblue}{ETHPetrol}
\colorlet{ETHbrown}{ETHBronze}

\usepackage{upgreek}
\usepackage{pifont}
\usepackage{fontawesome5}
\usepackage{xcolor}

\usepackage{graphicx}

\definecolor{mintgreen}{RGB}{152, 255, 152}

\DeclareUrlCommand\code{\urlstyle{tt}}

\newcommand{\defn}[1]{\textbf{#1}}
\newcommand{\defeq}{\stackrel{\scriptscriptstyle\mathrm{def}}{=}}

\def\1{\mathbf{1}}

\newcommand{\R}[0]{\mathbb{R}}

\DeclareMathOperator*{\E}{\mymacro{\mathbb{E}}}
\DeclareMathOperator*{\Var}{\mymacro{\mathbb{V}}}

\DeclarePairedDelimiter\abs{\lvert}{\rvert}

\newcommand{\set}[0]{\mymacro{S}}
\DeclareMathOperator*{\argmax}{argmax}

\newcommand{\indicator}[1]{\mathbbm{1}\mleft\{#1\mright\}}
\newcommand{\bigo}[0]{\mathcal{O}}
\newcommand{\KL}{\mathrm{KL}}

\newcommand{\CC}[0]{\mymacro{\mathbf{C}}}

\newcommand{\BB}[0]{\mymacro{\mathbf{B}}}
\newcommand{\Capprox}{\mymacro{\widehat{\CC}}}
\newcommand{\Bapprox}{\mymacro{\widehat{\BB}}}
\newcommand{\Lapprox}{\mymacro{\widehat{\LL}}}

\newcommand{\embedy}{\mymacro{{\embed(\yy)}}}
\newcommand{\plm}[0]{\mymacro{\ensuremath{p}}\xspace}
\newcommand{\proposal}[0]{\mymacro{q}}

\newcommand{\localTmodel}[0]{\mymacro{p^{\circ}_{T'}}}
\newcommand{\plmPrefix}[0]{\mymacro{\vec{p}}}
\newcommand{\localTmodelPrefix}[0]{\mymacro{\vec{p}^{\,\circ}_{T'}}}
\newcommand{\globalTmodel}[0]{\mymacro{p^{\bullet}_{T}}}

\newcommand{\incl}[1]{\mymacro{u_{#1}}}
\newcommand{\overlap}[0]{\mymacro{c}}
\newcommand{\falling}[2]{(#1)_{#2}}
\newcommand{\yyPast}[0]{\mymacro{\yy_{<t}}}
\newcommand{\pDPP}[0]{\mymacro{P}}

\newcommand{\U}[0]{\subpopulation} 
\newcommand{\universe}[0]{\mymacro{\mathcal{Y}}}
\newcommand{\subpopulation}[0]{\mymacro{\widehat{\universe}}}

\newcommand{\K}{\mymacro{K}}

\newcommand{\Zhat}{\widehat{\normalizer}}

\newcommand{\mymacro}[1]{{#1}}

\newcommand{\xx}[0]{\mymacro{\ensuremath{\boldsymbol{x}}}\xspace}
\newcommand{\yy}[0]{\mymacro{\ensuremath{\boldsymbol{y}}}\xspace}

\newcommand{\eos}[0]{\mymacro{\ensuremath{\textsc{eos}}}\xspace}

\newcommand{\local}[1]{\mymacro{\breve{#1}}}
\newcommand{\alphabet}[0]{\mymacro{\ensuremath{\Sigma}}\xspace}
\newcommand{\kleene}[1]{\mymacro{#1^{*}}}

\newcommand{\LL}[0]{\mymacro{\mathbf{L}}\xspace}
\newcommand{\simIID}[0]{\mathop{\overset{{\textnormal\scriptsize\mathrm{i.i.d.}}}{\sim}}}
\newcommand{\normalizer}[0]{\mymacro{Z}}

\newcommand{\strings}[0]{\kleene{\alphabet}}

\newcommand{\quality}[0]{\mymacro{\kappa}}
\newcommand{\embed}[0]{\mymacro{\boldsymbol{\upphi}}}
\newcommand{\Embs}[0]{\boldsymbol{\Phi}}

\newcommand{\uu}{\mathbf{u}}
\newcommand{\vv}{\mathbf{v}}

\newcommand{\bb}{\mathbf{b}}

\newcommand{\myalgname}[1]{\texttt{#1}\xspace}
\newcommand{\algname}[0]{\PoolDPP}
\newcommand{\PoolDPP}[0]{\myalgname{PoolDPP}}

\newcommand{\algnameMAP}[0]{\myalgname{PoolDPP-MAP}}
\newcommand{\algnameSamp}[0]{\myalgname{PoolDPP-Sampling}}
\newcommand{\algnameKM}[0]{\myalgname{$\K$-means}}
\newcommand{\algnameLLM}[0]{\myalgname{LLM-Select}}

\newcommand{\SampleDual}[0]{\myalgname{SampleDual}}
\newcommand{\GreedyMAP}[0]{\myalgname{GreedyMAP}}

\newcommand{\fhat}{\widehat{f}}
\newcommand{\fhatom}{\widehat{f}_\omega}
\newcommand{\pihat}{\widehat{\pDPP}_\omega}
\newcommand{\pibar}{\overline{\pDPP}}
\newcommand{\alphaMK}{\alpha_{N,\K}}

\newcommand{\eig}{\mathrm{eig}}
\newcommand{\Ptilde}{\mathbb{P}}%

\newcommand{\justification}[1]{\text{{\scriptsize\color{black!30}\parbox{.3\linewidth}{#1}}}}

\usepackage{thmtools}

\title{A Model of Diverse Sampling from Language Models}

\author{%
  \href{https://manueldeprada.com/}{\textcolor{black}{Manuel Prada-Corral}}\textsuperscript{\mdseries 1,2,3} \quad
  \href{https://yahya010.github.io/}{\textcolor{black}{Yahya Emara}}\textsuperscript{\mdseries 4} \quad
  \href{https://todonnell.github.io/}{\textcolor{black}{Timothy J.\@ O'Donnell}}\textsuperscript{\mdseries 5,6,7,8} \\
  \bf
  \href{https://rycolab.io/authors/ryan/}{\textcolor{black}{Ryan Cotterell}}\textsuperscript{\mdseries 1} \quad
  \href{https://timvieira.github.io/}{\textcolor{black}{Tim Vieira}}\textsuperscript{\mdseries 1} \\[0.5em]
  \textsuperscript{1}ETH Zürich \quad
  \textsuperscript{2}MPI-IS \quad
  \textsuperscript{3}ELLIS Institute Tübingen \quad
  \textsuperscript{4}Cornell University \\
  \textsuperscript{5}McGill University \quad
  \textsuperscript{6}Mila \quad
  \textsuperscript{7}Canada CIFAR AI Chair \quad
  \textsuperscript{8}CHI-FRO \\[0.4em]
  \texttt{\{\href{mailto:manuel.depradacorral@inf.ethz.ch}{manuel.depradacorral}, \href{mailto:ryan.cotterell@inf.ethz.ch}{ryan.cotterell}\}@inf.ethz.ch} \quad \texttt{\href{mailto:ye52@cornell.edu}{ye52@cornell.edu}} \\
  \texttt{\href{mailto:timothy.odonnell@mcgill.ca}{timothy.odonnell@mcgill.ca}} \quad \texttt{\href{mailto:tim.f.vieira@gmail.com}{tim.f.vieira@gmail.com}}
}

\begin{document}
\maketitle

\begin{abstract}    
Language models often produce high-quality individual samples but poor sets, with repeated samples clustering around the same semantic modes. We formalize diverse generation as sampling a size-$\K$ subset of complete strings from a determinantal point process (DPP), where the likelihood encodes item quality, and the embedding geometry encodes repulsion between similar outputs. Exact inference over all strings is intractable, so we propose \algname: draw a finite candidate pool from a tractable proposal, importance-weight the candidates toward the target quality, and run a $\K$-DPP sampler on the induced pool kernel; a greedy MAP variant provides deterministic selection. We prove a finite-sample bound with an explicit constant: the KL divergence between the sampler's output and the target $\K$-DPP decays as $\bigo(1/N)$ in the pool size. Empirically, \algname extends the quality--diversity frontier on open-ended generation and the precision--recall frontier on ambiguous text-to-SQL (Ambrosia), and on mathematical reasoning (GSM8K), it raises diversity at about the same pass@$\K$.
\begin{center}
\raisebox{-0.25em}{\includegraphics[height=1em]{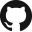}}
\url{https://github.com/rycolab/pool-dpp}
\end{center}
\end{abstract}

\etocdepthtag.toc{mainmatter}

\section{Introduction}
\label{sec:intro}
Search engines gave us pages of results; we browsed and chose. In contrast, language models often give us just one authoritative-sounding answer. 
However, for important questions---where multiple perspectives exist, or the right answer depends on context---we might want that plurality back. For example, when answering ambiguous questions, a model should surface multiple valid interpretations rather than commit to a single one \citep{ambrosia}. 
Pushing the language model's regenerate button 10 times yields 10 distinct outputs, but they often cluster around the same idea, phrased slightly differently.\footnote{Language models trained with reinforcement learning are particularly prone to collapse toward a narrow set of high-likelihood outputs, even when valid responses are broad. This has motivated prior work \citep{kazimi2025dppgrpo, chen2025dqo} to leverage ideas from DPPs to augment reinforcement learning training objectives, promoting diversity and preventing policy collapse.\looseness=-1
}  This makes sense because the model generates each string independently; it has no notion of what it means for a set of 10 strings to be \emph{diverse}.\looseness=-1

\begin{figure}[t!]
\begin{minipage}[c]{0.42\textwidth}
\caption{\emph{Diverse implementations of sorting.} Each point is a sorting-algorithm implementation, embedded by a code language model and projected to 3D with PCA; colors denote algorithm family. The $3$-DPP assigns subset probability proportional to the squared volume of the parallelepiped spanned by the selected quality-scaled embeddings---large when the selected items are individually high-quality (large norms) and mutually dissimilar. The highlighted triple has high probability because it combines implementations from distinct algorithmic strategies, yielding a large spanned volume; by contrast, choosing three variants from the same cluster (e.g., three quicksorts) produces nearly parallel embeddings and thus near-zero volume despite high individual quality.}
\label{fig:sorting-alg}

\hfill
\end{minipage}
\begin{minipage}[c]{0.58\textwidth}
\centering
\includegraphics[width=\linewidth]{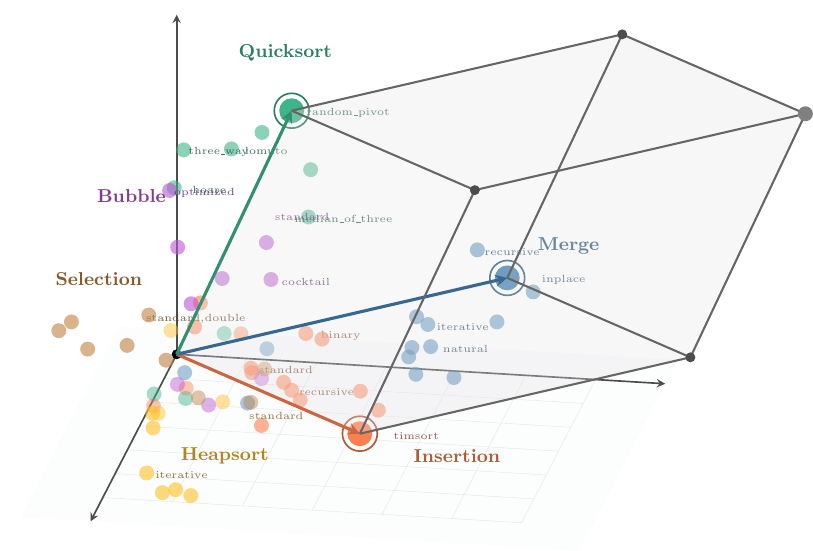}
\end{minipage}
\end{figure}

Why do samples cluster? Because nothing in the sampling process discourages similarity.
Temperature sampling broadens the distribution over next tokens, but it does not model relationships between outputs---two samples may independently land in the same high-probability region.\footnote{Several articles \citep[e.g.,][]{holtzman2020curiouscaseneuraltext,zhang2020tradingdiversityqualitynatural,meister-etal-2023-efficacy} have studied the pros and cons of decoding heuristics, such as temperature sampling, top-$p$ sampling, and their close cousins, including the effects on quality and diversity.\looseness=-1
}
Diverse decoding methods based on beam search \citep{vijayakumar2018diversebs,meister-etal-2021-determinantal} inject heuristics during decoding, but they lack a globally coherent notion of what it means for two strings to be similar or for a set to be diverse. These approaches are attractive because they are fast, but they are documented to produce artifacts such as incoherence or sudden shifts in topic \citep{ippolito-etal-2019-comparison, tevet-berant-2021-evaluating}. Other work has explored adopting sampling-without-replacement schemes in LMs \citep{kool2019stochasticbeamsthemgumbeltopk, meister-etal-2021-conditional}, but the notion of similarity in these models is exact equality. Thus, these methods tend to cluster like independent samples.
To get diverse outputs, we need to define similarity and then explicitly discourage it.

We formalize \emph{diverse sampling} as a probabilistic model over subsets of strings using determinantal point processes \citep[DPPs;][]{kulesza-dpp}.\footnote{DPPs started in quantum physics \citep{Macchi_1975} and later moved to machine learning \citep{kulesza-dpp} with applications to recommendation \citep{Wilhelm2018}, summarization \citep{Gong2014}, and language typology \citep{cotterell2021probabilistic}.\looseness=-1}
The key ingredient is a similarity function: we represent each string by its embedding (e.g., from a pretrained sentence embedder) scaled by its quality, so that the direction encodes meaning and the magnitude encodes quality.
Two strings are similar if their embeddings point in roughly the same direction.
A DPP assigns probability to subsets proportional to the squared volume spanned by their quality-scaled embeddings---large when items are high-quality and point in different directions, near-zero when they are redundant.
This gives us a well-defined distribution over sets: one that assigns high probability to sets of $\K$ strings that are individually good and collectively spread out, with a parameter that controls the balance between quality and diversity. \Cref{fig:sorting-alg} illustrates this in the context of generating meaningfully different sorting algorithm implementations.

Since exact inference in this DPP over all strings is intractable, we propose \algname, a finite-pool approximation. \algname draws $N$ candidate strings from a tractable proposal, importance-weights them toward a globally tempered
quality distribution, and applies a $\K$-DPP on the resulting pool. The same pool kernel supports two inference modes: stochastic $\K$-DPP sampling (\algnameSamp) and greedy MAP selection (\algnameMAP). For the sampling mode, we prove a non-asymptotic approximation guarantee: the KL divergence between the marginal output law of \algnameSamp (its output distribution, averaged over random pools) and the target $\K$-DPP decreases as $\bigo(1/N)$ with the pool size. The bound's constant is explicit, and we evaluate it on each task, on instances built from its pools rather than on the full proposal; the MAP mode has no distributional guarantee. Empirically, \algname extends the quality--diversity frontier on open-ended generation and the precision--recall frontier on ambiguous text-to-SQL. On mathematical reasoning, it raises diversity at about the same pass@$\K$ (within $0.01$; \cref{sec:gsm8k}), and we measure how weakly the embedders separate correct from incorrect reasoning chains.

\section{Language Modeling}
\label{sec:background}
A \defn{language model} $\plm$ is a probability distribution over the set $\strings$ of all strings over an alphabet $\alphabet$. 
Every language model $\plm$ admits the factorization $\plm(\yy) = \plmPrefix(\eos \mid \yy)\prod_{t=1}^{\abs{\yy}} \plmPrefix\mleft(y_t \,\middle|\, \yyPast\mright)$, where $\yyPast \defeq y_1,{\ldots},y_{t-1}$, $\abs{\yy}$ denotes the string's length, and $\eos \notin \alphabet$ is a distinguished \defn{end-of-string marker}.
Here, $\plmPrefix(\yy) \defeq \sum_{\yy' \in \strings} \plm(\yy\yy')$ is the \defn{prefix probability}, the probability that a string drawn from $\plm$ starts with $\yy$. Its conditionals are $\plmPrefix(y \mid \yy) \defeq \plmPrefix(\yy y)/\plmPrefix(\yy)$ for $y \in \alphabet$ and $\plmPrefix(\eos \mid \yy) \defeq \plm(\yy)/\plmPrefix(\yy)$ (if $\plmPrefix(\yy) = 0$, we set $\plmPrefix(y \mid \yy) \defeq \indicator{y = \eos}$ for all $y \in \alphabet \cup \{\eos\}$).
In our experiments, the language model is conditioned on a prompt $\xx \in \strings$: $\plm(\yy \mid \xx) \defeq \plm(\xx\yy)/\plmPrefix(\xx)$, the probability that a string starting with $\xx$ continues with $\yy$ and then ends. We write $\plm(\yy)$ when the prompt is fixed.
This factorization gives a generation procedure, known as \defn{ancestral sampling}: starting from the empty prefix, we sample tokens left-to-right as $y_t \sim \plmPrefix(\cdot \mid \yyPast)$ until $\eos$ is drawn.\looseness=-1

We define \defn{local annealing} by temperature $T'$ as the following change to the conditional distribution, and the \defn{locally annealed language model} as the product of the annealed conditionals:
\par\vspace{-\parskip}\noindent
\begin{minipage}{0.51\linewidth}
\begin{equation}
\label{eq:local-temp}
\!\!\!\!\localTmodelPrefix(y_t \mid \yyPast) \defeq \frac{\plmPrefix(y_t \mid \yyPast)^{1/T'}}{\sum_{y \in \alphabet \cup \{\eos\}} \plmPrefix(y \mid \yyPast)^{1/T'}}\!
\end{equation}
\end{minipage}%
\hfill
\begin{minipage}{0.48\linewidth}
\begin{equation}
\label{eq:local-temp-string}
\!\!\localTmodel(\yy) \defeq \localTmodelPrefix(\eos \mid \yy) \prod_{t=1}^{\abs{\yy}} \localTmodelPrefix(y_t \mid \yyPast)\!\!
\end{equation}
\end{minipage}
\par\smallskip\noindent
for all $y_t \in \alphabet \cup \{ \eos \}$ and $\yy \in \strings$.
Ancestral sampling with the annealed conditionals draws from $\localTmodel$.
As $T' \to 0$, each conditional concentrates on its mode, yielding greedy decoding; in contrast, as $T' \to \infty$, each conditional approaches the uniform distribution over $\alphabet \cup \{ \eos \}$.
High local temperatures increase diversity at the cost of coherence; low temperatures increase coherence at the cost of diversity \citep{deepseekai2025deepseekr1incentivizingreasoningcapability,holtzman2020curiouscaseneuraltext}.
Local annealing is efficient, as it only requires $\bigo(|\alphabet|)$ additional work per time step to adjust its conditional distribution.\looseness=-1

In contrast, \defn{global annealing} by temperature $T$ modifies the probability of entire sequences:
\begin{equation}\label{eq:global-temp}
\globalTmodel(\yy) \defeq \frac{\plm(\yy)^{1/T}}{\sum_{\yy' \in \strings} \plm(\yy')^{1/T}}
\end{equation}
when the sum is finite, which holds for every $T \le 1$, since $\plm(\yy)^{1/T} \le \plm(\yy)$.
Unlike local annealing, which modifies each conditional independently, global annealing linearly scales log-probability differences between sequences: high $T$ compresses these differences, while low $T$ amplifies them.  Global annealing tends to preserve the coherence and fluency of the underlying language model.
\citet[\S 3.2]{kempton2025localnormalizationdistortionthermodynamic} show that local annealing does not generally yield a good approximation to the globally annealed distribution $\globalTmodel$.
Unfortunately, exact sampling from the global distribution is intractable.\footnote{Global annealing $\globalTmodel$ concentrates on the most likely string as $T \!\to\! 0$.  However, finding the most likely string is NP-hard \citep{shimony1994finding,knight1999decoding}. Recent work has explored MCMC approximations to global annealing \citep{karan2025reasoning}.}%
\looseness=-1

\section{Determinantal Point Processes}\label{sec:dpp}

Determinantal point processes (DPPs) are distributions over subsets that naturally encode diversity: items that are similar are unlikely to be selected together.
We use DPPs to define a distribution over sets of strings where high-probability sets contain strings that are both high-quality and well-dispersed.

\paragraph{The many faces of diversity.}
There are many ways to operationalize the notion of a diverse set of strings.
In natural language, we can distinguish semantic diversity (variation in meaning), syntactic diversity (variation in surface form), and stylistic diversity (variation in tone and register).
In specialized domains such as code generation, one can also consider \emph{algorithmic} diversity, i.e., variation in the underlying algorithms implemented by the generated code.
Which notion matters depends on the application.
In this work, we focus on notions of diversity that can be expressed through embeddings. This lets us quantify similarity geometrically and swap in a different embedder depending on the domain.\looseness=-1
\paragraph{Embeddings.}
To measure similarity between strings, we represent each string $\yy$ by an embedding $\embedy \in \R^D$ from a pretrained embedder.
The choice of embedder determines what notion of diversity the DPP captures---semantic, syntactic, or domain-specific.
We try three options across our experiments: a small off-the-shelf semantic embedder (\textbf{MiniLM-L6}), a larger one (\textbf{Qwen3-Embedding-0.6B}), and the generating LM's own mean-pooled last-layer hidden states. On open-ended tasks like FineWeb (\cref{sec:intrinsic}), the two off-the-shelf embedders give very similar quality--diversity Pareto frontiers across language models (\cref{fig:abl_lm_emb}), so we use the smaller MiniLM for these tasks; on more constrained tasks like math reasoning on GSM8K (\cref{sec:gsm8k}), the larger Qwen3-Embedding-0.6B picks up structural distinctions between reasoning chains that MiniLM blurs, and we adopt it as the default. The self-contained LM-hidden-state variant performs on par with the small semantic embedder in several settings but does not consistently improve over it; layer and pooling choices that we did not exhaustively optimize, and a semantic-oriented fine-tune, are a plausible route to closing that gap.\looseness=-1
\paragraph{Building a DPP.}
Given a \defn{base set} $\universe$, a \defn{quality measure} $\kappa\colon \universe \to \R_{\ge 0}$, and a (unit-norm) \defn{embedding} $\embed\colon \universe \to \R^D$ (i.e., $\|\embedy\|_2 = 1$ for all $\yy \in \universe$),\footnote{It is important to normalize the raw embeddings by their $\| \cdot \|_2$, or else an embedding's raw length will confound the quality measure.} we define the (quality--diversity) \defn{kernel matrix} \citep[\S 3.1]{kulesza-dpp}:
\begin{equation}\label{eq:quality-diversity}
\LL(\xx,\yy) \defeq \sqrt{\quality(\xx)}  \embed(\xx)^\top \embed(\yy)  \sqrt{\quality(\yy)}
\end{equation}
Intuitively, in the \defn{quality-scaled embedding} $\sqrt{\quality(\yy)} \embedy$, the direction encodes meaning (for diversity) and the magnitude encodes quality.
Let $\set \subseteq \universe$ be a finite set, and $\LL_\set \defeq [\LL(\xx,\yy)]_{\xx,\yy \in \set} \in \R^{\set \times \set}$ be the kernel matrix indexed by $\set$.
A \defn{determinantal point process} (\defn{DPP}) is a probability distribution over subsets $\set \subseteq \universe$ where the probability of the set $\set$ is proportional to $\det(\LL_\set)$.
This determinant has a geometric interpretation: it equals the squared volume of the parallelepiped spanned by the quality-scaled embeddings $\{\sqrt{\quality(\yy)}  \embedy\}_{\yy \in \set}$ \citep[\S 3]{kulesza-dpp}.
This volume is large when strings are high-quality and point in different directions, and collapses to zero when two strings are nearly identical.  More precisely, under \cref{eq:quality-diversity}, the determinant decomposes as $\det(\LL_\set) = (\prod_{\yy\in\set}\quality(\yy)) \det(\Embs_\set^\top \Embs_\set)$, where $\Embs_\set \defeq [ \embedy ]_{\yy \in \set} \in \R^{D \times \set}$. Here, the quality factor $\prod_{\yy\in\set}\quality(\yy)$ rewards individual quality, while the diversity factor $\det(\Embs_\set^\top \Embs_\set)$ penalizes collective similarity (the squared volume spanned by the embeddings).

\paragraph{$\K$-DPPs.}
Our work uses the $\K$-DPP variant, which introduces a set-size parameter $\K$. This parameter constrains the underlying DPP to generate sets of exactly $\K$ items; thus, it can be used to encourage diversity by controlling the number of elements.  The $\K$-DPP's probability is defined below:\footnote{A $\K$-DPP is equivalent to probabilistically conditioning an ordinary DPP to generate only size-$\K$ sets.}\looseness=-1
\begin{align}
\pDPP(\set) 
\defeq \frac{\det(\LL_\set)\indicator{{|\set|=\K}}}{\normalizer}
\qquad\text{where}\quad
\normalizer
\defeq \sum_{\set \in \binom{\universe}{\K}} 
\det(\LL_{\set})
\label{eq:k-dpp}
\end{align}
and $\binom{\universe}{\K}$ is the set of size-$\K$ subsets from $\universe$. $\pDPP(\set)$ is only well-defined when $0 < \normalizer < \infty$.\footnote{The lower bound holds provided at least one $\K$-element set with nonzero determinant exists. The upper bound holds whenever $\quality_\Sigma \defeq \sum_{\yy\in\strings}\quality(\yy) < \infty$: Hadamard's inequality gives $\det(\LL_\set) \le \prod_{\yy\in\set}\quality(\yy)$, so $\normalizer \le \quality_\Sigma^{\K}/\K!$. In our instantiation, $\quality(\yy) = \plm(\yy)^{1/T}$ (next paragraph), so $\quality_\Sigma$ is the normalizer of the globally annealed model (\cref{eq:global-temp}), which is finite for $T \le 1$; for $T > 1$ it is an assumption on the tail of $\plm$, holding trivially whenever the generation length is capped, as in practice (\cref{app:length-cap}).\looseness=-1
}%

\paragraph{Trading off quality and diversity.}
We also make use of a parameter $T > 0$, the \defn{global temperature}, to trade off quality and diversity.  Specifically, we use
\begin{equation}\label{eq:quality-temp}
\quality(\yy) = \plm(\yy)^{1/T}
\end{equation}
so $T$ sets the target's quality, which connects to globally annealed sampling (\cref{eq:global-temp}).\footnote{Note: Because we use a $\K$-DPP model, we do not need to compute the normalization constant for the globally annealed distribution, as the $\K$-DPP is invariant to rescaling its kernel $\LL$; this is not true of an ordinary DPP.}
Varying $T$ smoothly trades off quality and diversity: smaller $T$ sharpens $\quality$ toward high-likelihood strings (favoring quality); larger $T$ flattens $\quality$ (favoring diversity).  We found the global temperature parameter $T$ essential for calibrating the language model's likelihood to the embedder's volumes, as $T=1$ is rarely optimal.

\paragraph{Inference via the dual.}
The standard $\K$-DPP sampler eigendecomposes the $\universe \times \universe$ kernel $\LL$, which costs $\bigo(N^3)$ for a base set of size $N$ (in \cref{sec:algorithm}, the pool). When the embedding dimension $D$ is much smaller than $N$, the dual representation avoids this. The kernel factors as $\LL = \BB^\top \BB$, where the \defn{feature matrix} $\BB \in \R^{D \times \universe}$ has as columns the quality-scaled embeddings $\sqrt{\quality(\yy)}\embedy$, and it shares its nonzero eigenvalues with the $D \times D$ \defn{dual kernel}\looseness=-1
\begin{equation}\label{eq:dual-kernel}
\CC \defeq \BB \BB^\top = \sum_{\yy \in \strings} \quality(\yy)  \embed(\yy) \embed(\yy)^\top
\end{equation}
The dual sampler \SampleDual of \citet[Alg.~3]{kulesza-dpp} (\cref{alg:dual-kdpp}, described in \cref{app:dual-sampler}) runs the $\K$-DPP sampler on $\CC$ instead of $\LL$, at cost $\bigo(D^2 N + D^3 + D N \K)$, which is linear in $N$ for fixed $D$.

The dual trick needs the base set in memory, but ours is the infinite set $\strings$. \Cref{sec:algorithm} restricts it to a finite pool drawn from a proposal, and \cref{sec:analysis} analyzes the approximation.

\section{Pool-Based Inference}
\label{sec:algorithm}

Exact inference in the $\K$-DPP over $\strings$ is intractable.\footnote{Structured DPPs can be tractable under strong assumptions on $\quality$ and $\embed$ \citep{structured-dpp}, but those assumptions rule out the language models and embedders we use.}
We approximate it by drawing a finite pool of candidate strings from a proposal distribution, importance-weighting the pool toward the quality function, and running DPP inference on the induced finite kernel.

\paragraph{Approximate $\K$-DPP.}
Let $\U \defeq (\yy^{(1)}, {\ldots}, \yy^{(N)})$ with $\yy^{(n)} \simIID \proposal$ be a \defn{pool} of candidates.\footnote{\label{fn:ac}We require $\quality \ll \proposal$, that is, $\proposal(\yy) = 0 \implies \quality(\yy) = 0$ ($\quality$ is absolutely continuous with respect to $\proposal$).}
We define the \defn{pool feature matrix}, which importance-weights the embeddings of the drawn strings:
\begin{equation}\label{eq:B-matrix}
\Bapprox \defeq \left[\sqrt{w^{(n)}/N}  \embed(\yy^{(n)})\right]_{n=1}^{N} \in \R^{D \times N}
\quad\text{where}\quad
w^{(n)} \defeq \frac{\quality(\yy^{(n)})}{\proposal(\yy^{(n)})}
\end{equation}
These weights $w^{(n)}$ correct for the mismatch between the proposal and the target quality distribution $\quality$.
If a string is sampled multiple times, each occurrence contributes its own (identical) column to $\Bapprox$.
Duplicates are thereby handled naturally: any selection containing two copies of the same string has two identical columns and hence a zero determinant, so it receives zero probability, while the copies' weights add up across selections, preserving the string's total quality mass.
\Cref{sec:analysis} makes both effects precise: collisions are exactly what the constant $\alphaMK$ of \cref{eq:PoolDPP-fhat} accounts for in the analysis (\cref{prop:pushforward,prop:PoolDPP-well-behaved}).

The \defn{dual kernel approximation} is the following matrix (which satisfies $\E_\omega[\Capprox]=\CC$):
\begin{equation}
\Capprox \defeq \Bapprox \Bapprox^\top = \frac{1}{N} \sum_{n=1}^{N} w^{(n)}  \embed(\yy^{(n)}) \embed(\yy^{(n)})^\top \in \R^{D \times D}\label{eq:c_hat}
\end{equation}
The \defn{pool $\K$-DPP} is the $\K$-DPP whose base set is the pool indices $\{1, {\ldots}, N\}$ and whose kernel is the \defn{pool kernel} $\Lapprox \defeq \Bapprox^\top\Bapprox \in \R^{N \times N}$, which has the same nonzero eigenvalues as $\Capprox$.

\paragraph{Two modes of inference.}
The pool $\K$-DPP is a finite, tractable distribution; we use it in two ways:\looseness=-1
\begin{itemize}[leftmargin=*]
\item \algnameSamp draws an exact sample from the pool $\K$-DPP by invoking \cref{alg:dual-kdpp} with $\K$ and $\Bapprox$. The base set of this $\K$-DPP is the set of pool \emph{indices} $\{1, {\ldots}, N\}$; given the sampled index set $I$, \algnameSamp returns the labeled strings $\{\yy^{(n)}\}_{n \in I}$. \Cref{prop:pushforward} characterizes the distribution of this string set.
\item \algnameMAP targets the most likely set $\set^\star$ in the pool: the size-$\K$ subset of the pool's strings that maximizes $\det\LL_\set$ (\cref{eq:k-dpp}). This objective depends on the quality and the embeddings of the strings and not on how often the proposal drew them, so \algnameMAP needs no importance correction: its feature matrix has the quality-scaled embeddings $\sqrt{\quality(\yy)} \embed(\yy)$ as columns. Exact maximization is NP-hard \citep{ko1995exact}; the greedy algorithm (\cref{alg:greedy-map}) returns a set $\set^\dagger$ whose determinant is within a multiplicative factor $(\K!)^{2}$ of optimal, $\det \LL_{\set^\dagger} \ge \det \LL_{\set^\star} / (\K!)^{2}$ \citep[Thm.~11]{civril2009maxvol}.
Note that $\set^\star$ is the best size-$\K$ subset \emph{in the pool}, not the intractable population optimum over $\strings$.\looseness=-1
\end{itemize}
The two modes call for different settings of the global temperature $T$ in $\quality(\yy) = \plm(\yy)^{1/T}$: for sampling, the stochasticity of drawing a set of pool indices $I$ with probability $\propto\det\Lapprox_I$ already spreads mass across the pool, so $\quality$ can be sharp (small $T$, $T \in [0.43, 1.22]$ in our figures); for MAP, the deterministic arg-max would otherwise collapse onto near-clones of the highest-quality item, so $\quality$ must be flat (large $T$, $T \in [1, 500]$ in our figures; see \cref{sec:ablations}).%

\paragraph{Instantiation.}
\label{par:instantiation}
Our experiments use the quality function of \cref{eq:quality-temp} and locally annealed ancestral sampling (\cref{eq:local-temp}) as the proposal, $\proposal(\yy) = \localTmodel(\yy)$.
The importance weights then capture the difference between local and global annealing.

In \cref{sec:analysis}, we show that this procedure is equivalent to a Rao--Blackwellized importance sampler, which marginalizes out the randomness in selecting subsets from $\U$, yielding lower-variance tractable estimates with theoretical guarantees.

\section{Algorithm Analysis}
\label{sec:analysis}

\PoolDPP samples from a $\K$-DPP restricted to a finite pool $\U$ instead of the target $\K$-DPP over $\strings$. We bound the resulting bias in the pool size $N$ with a generic tool, the \emph{sketched distribution}. This section states the results; \cref{app:PoolDPP-proofs} has the proofs, and \cref{app:lean} describes their Lean formalization, which is in the code repository linked on the first page.%

\paragraph{Sketched distributions.}
Let $f(\set) \ge 0$ define an unnormalized target $\pDPP(\set) = f(\set)/\normalizer$ on a countable domain $\mathcal{D}$, with $0 < \normalizer = \sum_{\set \in \mathcal{D}} f(\set) < \infty$ throughout (in our case $\mathcal{D} = \binom{\strings}{\K}$ with $f(\set) = \det(\LL_\set) \indicator{|\set|{=}\K}$).
A \defn{sketch}\footnote{We use \emph{sketch} in the sense of randomized sketching in numerical linear algebra \citep{woodruff14sketching}: internal randomness compresses an intractable object into a tractable random surrogate; here, the infinite base set $\strings$ into a finite pool, and the population dual kernel $\CC$ into $\Capprox$ with $\E_\omega[\Capprox] = \CC$.}
is a pair $(\mu, \fhat)$: a distribution $\mu$ over internal randomness $\omega$, together with a (measurable) map $\fhat \colon \omega \mapsto \fhatom$ that produces an unnormalized weight function $\fhatom(\set) \ge 0$ determined by $\omega$ (the subscript in $\fhatom$ records this dependence). Each realization of $\omega$ induces the sketched normalizer $\Zhat \defeq \sum_{\set \in \mathcal{D}} \fhatom(\set)$ and, when $\Zhat > 0$, the sketched distribution $\pihat(\set) \defeq \fhatom(\set)/\Zhat$.
We suppose without loss of generality that $\Zhat > 0$ almost surely (\cref{fn:wlog}).
One run draws one $\omega \sim \mu$ and then $\set \sim \pihat$, so its output follows the \defn{marginal output law}:
\begin{align}\label{eq:marginal-output-law}
\pibar(\set) \defeq \E_\omega\mleft[\pihat(\set)\mright]
\end{align}
which marginalizes the sketch's internal randomness $\omega$.
A sketch is \defn{well-behaved} if $\E_\omega[\fhatom(\set)] = f(\set)$ for every $\set$ and $\E_\omega[\Zhat^{2}] < \infty$.\footnote{\label{fn:wlog}If $\Pr_\omega[\Zhat = 0] > 0$, \algname redraws $\omega$; \cref{app:redraw} shows that the bounds below still hold, with $\Var_\omega(\Zhat)$ taken under $\mu$.}

\begin{restatable}[Variance controls bias]{theorem}{thmVarKL}
\label{thm:var-kl-mixture}
For any well-behaved sketch $(\mu, \fhat)$,
\begin{align}\label{eq:var-kl}
\chi^2(\pDPP \| \pibar) \le \frac{\Var_\omega(\Zhat)}{\normalizer^2}\quad\text{and}
\quad
\KL(\pDPP \| \pibar)
\le
\log\mleft(1 + \frac{\Var_\omega(\Zhat)}{\normalizer^2}\mright)
\le
\frac{\Var_\omega(\Zhat)}{\normalizer^2}
\end{align}
\end{restatable}
\deferredproof{app:proof-var-kl-mixture}

\paragraph{\algname is a sketch.}
We now instantiate the sketch $(\mu, \fhat)$ for \algname.
The internal randomness \emph{is} the pool of \cref{sec:algorithm}: $\omega = (\yy^{(1)}, {\ldots}, \yy^{(N)}) \simIID \proposal$.
The pool supplies the randomness, and the sketch is the deterministic map from a realized pool to the weight function $\fhatom$ that defines $\pihat$; drawing a subset from $\pihat$ is a second, separate stage of randomness.
Writing $[N] \defeq \{1, {\ldots}, N\}$ for the set of pool indices, the sketched weight of a subset $\set \subseteq \strings$ aggregates every way the pool can realize $\set$:\looseness=-1
\begin{align}\label{eq:PoolDPP-fhat}
\fhatom(\set) \defeq \alphaMK
\smashoperator{\sum_{\substack{I \subseteq [N], |I|=\K \\ \{\yy^{(n)} \mid n \in I \} = \set}}}
\det(\Lapprox_I)
\quad\text{where}\
\alphaMK \defeq \frac{N^\K}{(N)_\K},
\ \ (N)_\K \defeq N(N{-}1)\cdots(N{-}\K{+}1)
\end{align}
where $I$ ranges over size-$\K$ sets of pool \emph{indices}, $\Lapprox$ is the pool kernel (\cref{sec:algorithm}), and the constant $\alphaMK = 1/\Pr[\text{$\K$ uniform draws from $[N]$ are all distinct}]$ makes the weights unbiased.\footnote{$\alphaMK$ cancels in $\pihat$ (\cref{prop:pushforward}), but \cref{thm:var-kl-mixture} requires $\E_\omega[\fhatom(\set)] = f(\set)$; see \cref{app:redraw}.\looseness=-1}

\algnameSamp returns the strings of an index set $I$ that \cref{alg:dual-kdpp}, invoked with $\Bapprox$, draws with probability proportional to $\det(\Lapprox_I)$. These strings follow $\pihat$ (\cref{prop:pushforward}): index sets that repeat a string have determinant $0$, and the set-independent constant $\alphaMK$ cancels in $\pihat = \fhatom/\Zhat$. A pool has $\Zhat = 0$ exactly when $\mathrm{rank}(\Capprox) < \K$, and \algname then redraws it (\cref{fn:wlog}).

\paragraph{Inclusion ratios.}
The analysis compares the target with the proposal through \defn{inclusion ratios}: for $\overlap \le \K$ distinct strings,\looseness=-1
\begin{align}\label{eq:incl-def}
\incl{\overlap}(\yy_1, {\ldots}, \yy_\overlap) \defeq \frac{\pDPP(\{\yy_1, {\ldots}, \yy_\overlap\} \subseteq \set)}{\falling{\K}{\overlap} \, \proposal(\yy_1) \cdots \proposal(\yy_\overlap)},
\qquad
\falling{\K}{\overlap} \defeq \K (\K{-}1) \cdots (\K{-}\overlap{+}1)
\end{align}
and $\incl{\overlap} \defeq 0$ on tuples with a repeat and on tuples outside the support of $\proposal$ (there, $\quality = 0$ because $\quality \ll \proposal$, so the target never contains them).
The ratio $\incl{\overlap}$ has mean $1$ under $\proposal^{\otimes \overlap}$ and is unchanged by rescaling $\quality$. We write $\rho \defeq \E_{\yy \sim \proposal}[\incl{1}(\yy)^2]$ for the second moment of $\incl{1}$. For any $\K \ge 1$, $N \ge \K$, and proposal $\proposal$ with $\quality \ll \proposal$, the sketch is unbiased, $\E_\omega[\fhatom(\set)] = \det(\LL_\set) \indicator{|\set|=\K}$, and if $\rho < \infty$, then $\E_\omega[\Zhat^2] < \infty$, so \PoolDPP is well-behaved (\cref{prop:PoolDPP-well-behaved}).

\paragraph{Finite-sample bound.}
Combining \cref{thm:var-kl-mixture,prop:PoolDPP-well-behaved} reduces bounding the KL to bounding the \defn{relative variance} $\Var_\omega(\Zhat)/\normalizer^2$.
Since $\Zhat$ is a $U$-statistic of order $\K$ (\cref{eq:ustat}), its variance has an exact formula \citep{hoeffding1948class}, whose terms are variances of the inclusion ratios $\incl{\overlap}$ of \cref{eq:incl-def}.

\begin{restatable}[Finite-sample bound]{theorem}{thmZfree}
\label{thm:zfree}
For any $\K \ge 1$, $N \ge \K$, and proposal $\proposal$ with $\quality \ll \proposal$,
\begin{align}\label{eq:zfree}
\frac{\Var_\omega(\Zhat)}{\normalizer^{2}}
=
\sum_{\overlap=1}^{\K} a_\overlap  \Var_{\proposal^{\otimes \overlap}}\mleft(\incl{\overlap}\mright),
\qquad
a_\overlap \defeq \frac{\binom{\K}{\overlap}\binom{N-\K}{\K-\overlap}}{\binom{N}{\K}}
\end{align}
If $\rho < \infty$, then for any fallback
\begin{align}\label{eq:kl-closed-form}
\KL(\pDPP \| \pibar) \le \log\bigl(e^{\theta} - 1 + a_0\bigr) \le \theta \defeq{} \frac{\K^{2} \, \rho}{N - \K + 1}
\end{align}
{where $a_0 = \binom{N-\K}{\K}/\binom{N}{\K}$ is the formula of \cref{eq:zfree} at $\overlap = 0$.}
\end{restatable}
\deferredproof{app:zfree}

\noindent The coefficients $a_\overlap$ are the hypergeometric law of the overlap of two independent uniform $\K$-subsets of $[N]$, so the bound decays as $1/N$ (\cref{eq:hoeffding-bounds}). The $\overlap=1$ term measures how far $\pi_1(\yy) \defeq \pDPP(\yy \in \set)/\K$, the law of a uniformly chosen member of a target set, is from the proposal: since $\E_\proposal[\incl{1}] = 1$, $\rho = 1 + \Var_\proposal(\incl{1}) = 1 + \chi^2(\pi_1 \| \proposal)$.

Applying the same bound to every term of \cref{eq:zfree} gives a sharper bound, \cref{cor:kl-sum}, which we use in the numerical evaluation (\cref{app:zfree-eval}).
\paragraph{Bound scope.}
\Cref{thm:zfree} bounds the KL divergence between the target and the law of the single set that \algnameSamp emits, at every pool size.
\algnameMAP is a deterministic function of the pool and inherits no distributional guarantee; its only guarantee is the $(\K!)^2$ volume factor of the greedy algorithm relative to the best set \emph{in the pool} (\cref{sec:algorithm}).
The bound depends on the target only through the inclusion probabilities $\pDPP(\yy \in \set)$. For a language model, these need $\CC$ summed over all strings, which we cannot compute, so \cref{app:zfree-eval} evaluates the bound on a finite instance built from our pools: the strings that the pools drew, with their observed frequencies as the proposal. These instances hold a median proposal mass of $0.68$ on Ambrosia, $0.0017$ on GSM8K, and $0.0003$ on FineWeb. On them, the bound is small on Ambrosia and on FineWeb at $T{=}1$ and exceeds $1$ nat on GSM8K at $N{=}100$. These values are the bound on the instance, not a bound on the sampler on the full proposal.

\paragraph{Rao--Blackwell interpretation.}
\algname is the Rao--Blackwellization of the naive sampler that draws $\K$ strings i.i.d.\ from the proposal, with relative variance at least a factor $N/\K$ lower (\cref{rem:rao-blackwell}).

\section{Experiments}
\label{sec:experiments}

Unless otherwise noted, we sample sets of size $\K{=}10$ (\cref{fig:abl_k_size} ablates this choice). \algnameSamp is the mode that \cref{thm:zfree} covers, while \algnameMAP is its deterministic counterpart to be compared against the deterministic pool rerankers.

We report results primarily for \textbf{Gemma-2-9B-Instruct}, paired with \textbf{MiniLM-L6} on FineWeb and Ambrosia and \textbf{Qwen3-Embedding-0.6B} on GSM8K (\cref{sec:dpp}). We also evaluated \textbf{Qwen3-8B} and \textbf{Llama-3.1-8B-Instruct} on the FineWeb intrinsic experiment: \cref{fig:abl_lm_emb} compares Gemma and Qwen3-8B with both embedders, and the three-model Pareto comparison with MiniLM-L6 is in \cref{fig:abl_multimodel}. The qualitative ordering of methods is the same on every model. The LM-hidden-state embedding variant is discussed in \cref{sec:dpp}, and \cref{sec:ablations} sweeps the pool size $N$, subset size $\K$, and global temperature $T$, and reports per-stage wall-clock time (\cref{tab:efficiency-fixed,tab:efficiency-scaling}). Pool construction dominates runtime; \algnameMAP and \algnameSamp selection cost is negligible compared to LM sampling.

\begin{figure}[t]
\begin{minipage}[c]{0.6\textwidth}
\centering
\includegraphics[width=\columnwidth]{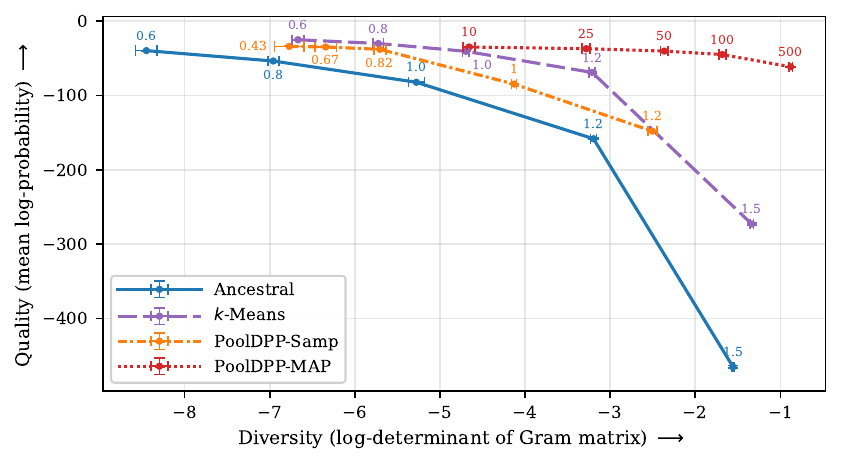}
\end{minipage}
\begin{minipage}[c]{0.4\textwidth}
\caption{\textbf{Intrinsic evaluation (FineWeb).} Quality vs.\ diversity; higher is better. Curves sweep the local sampling temperature (ancestral sampling, $\K$-means) or the global temperature $T$ (\algnameSamp, \algnameMAP). \algnameSamp lies below the $\K$-means frontier at intermediate diversity and produces different sets across runs. Error bars are $95\%$ $t$-intervals over $10$ pools (Gemma-2-9B, MiniLM-L6, $\K{=}10$, $N{=}400$)}
\label{fig:fineweb_pareto}
\end{minipage}
\end{figure}

We report \algname in both inference modes (\algnameSamp and \algnameMAP). We compare against the following baselines:
\begin{enumerate}
\item \textbf{Temperature sampling}: Ancestral sampling with local annealing at various temperatures.
\item \textbf{Naive DPP}: Samples a pool at $T'{=}1$, then applies DPP selection using the proposal likelihood $\plm(\yy)$ as the quality score. It isolates the DPP without the global temperature correction.
\item\label{item:divbs} \defn{Diverse beam search (DivBS)} \citep{vijayakumar2018diversebs}: A decoding heuristic that penalizes sibling beams for producing similar tokens.
\item\label{item:detbs} \defn{Determinantal beam search (DetBS)} \citep{meister-etal-2021-determinantal}: A beam-search variant that replaces the top-$\K$ step with a \emph{local} $\K$-DPP MAP query over a string-subsequence-kernel similarity matrix. We report both DivBS and DetBS in \cref{app:beam_baselines}; neither matches sampling-based methods on quality--diversity, so we omit them from \cref{fig:fineweb_pareto}.
\item \textbf{Set-level baselines}: Stochastic beam search (SBS; \citealp{kool2019stochasticbeamsthemgumbeltopk}) and conditional Poisson SBS (CPSBS; \citealp{meister-etal-2021-conditional}), which sample sets without replacement during decoding, and two rerankers on the same pools and embeddings as \algname: minimum Bayes risk (MBR; \citealp{bickel2015mathematical,kumar-byrne-2004-minimum}) and maximal marginal relevance (MMR; \citealp{carbonell1998mmr}). \Cref{app:baselines} reports them on all three tasks.
\item \textbf{$\K$-means reranking} (\algnameKM): The deterministic pool-reranking baseline. We sample the same pool $\U$ as for \algname, cluster the embeddings $\{\embedy\}_{\yy \in \U}$ into $\K$ clusters with $\K$-means \citep{arthur2007k-meanspp}, and return the highest-likelihood item per cluster. $\K$-medoids returns nearly identical subsets, so we report only $\K$-means.
\item \textbf{LLM-based selection} (\algnameLLM): The pool-reranking baseline that replaces the DPP with a language model. We present 40 pool items to Qwen3-8B and ask it to select $\K{=}10$ diverse completions, without access to log-probabilities or log-determinants. Its result is reported in \cref{sec:ablations}.
\end{enumerate}
Each prompt is evaluated on independently sampled pools (10 on FineWeb, 100 on Ambrosia and GSM8K), and error bars are 95\% confidence intervals over pools for every method. We do not run naive DPP on Ambrosia since $461$ of the $10{,}000$ (prompt, pool) pairs have $\Zhat = 0$ and the results are unreliable.
The methods differ only in their remaining sources of randomness: \algnameKM and \algnameMAP are deterministic given a pool, so their variability is driven by pool sampling alone, whereas \algnameSamp additionally redraws its DPP sample on each call.\looseness=-1

\subsection{Experiment 1: Intrinsic Evaluation}
\label{sec:intrinsic}

We first evaluate whether \algname improves the intrinsic quality--diversity Pareto frontier. 
We use 50 open-ended prompts from FineWeb \citep{fineweb} and sample $\K=10$ completions per prompt. 
For each prompt $\xx$, we sample $M = 10$ sets $\set^{(1)}, {\ldots}, \set^{(M)}$ and report quality as mean LM log-probability and diversity as the log-determinant of the normalized embedding Gram matrix, averaged over the $50$ prompts:
\begin{align}
\label{eq:intrinsic-metrics}
\text{Diversity}
&=
\frac{1}{M}\sum_{m=1}^M
\log\det\!\left(\Embs_{\set^{(m)}}^\top\Embs_{\set^{(m)}}\right),
&
\text{Quality}
&=
\frac{1}{M}\sum_{m=1}^M
\frac{1}{\K}\sum_{\yy\in\set^{(m)}}\log\plm(\yy\mid\xx)
\end{align}

Each curve in \cref{fig:fineweb_pareto} is traced by varying one parameter: for ancestral sampling and \algnameKM, we vary the local sampling temperature; for \algnameSamp and \algnameMAP, we draw the pool at local sampling temperature $1$ and vary the global temperature $T$.

\paragraph{Results.}
\Cref{fig:fineweb_pareto} shows that \algname improves the intrinsic quality--diversity frontier. 
Ancestral sampling traces the expected baseline: increasing local temperature improves diversity but sharply degrades quality. 
\algnameKM provides a strong deterministic pool-reranking baseline. \algnameMAP reaches a higher log-determinant with almost no quality loss, while \algnameKM keeps the absolute higher quality at a lower log-determinant. 
\algnameSamp does not dominate \algnameKM on the intrinsic objective, which is expected: \algnameMAP directly optimizes the determinant objective over the pool, whereas \algnameSamp samples from the corresponding pool distribution. 
The sampling mode instead provides stochastic diverse sets; across repeated draws from the same pool, $20$--$60\%$ of selected items differ, while \algnameKM and \algnameMAP are deterministic given the pool. 

We omit beam-search baselines from the plot because they fall outside the displayed quality--diversity range; full numbers are in \cref{app:beam_baselines}, and additional baselines and ablations are in \cref{sec:ablations}.

\subsection{Experiment 2: Ambiguity Resolution (Ambrosia)}
\label{sec:ambrosia}

We evaluate whether the intrinsic improvements translate to meaningful coverage of semantically distinct outputs. The \textbf{Ambrosia} dataset \citep{ambrosia} provides natural language queries with annotated ambiguities---each query admits multiple valid SQL interpretations corresponding to different readings of the ambiguous text.

The Ambrosia dataset contains queries exhibiting scope, attachment, and vagueness ambiguities. Each example includes human-annotated SQL queries representing distinct valid interpretations, along with a database for execution. Since each query has up to 3 annotated interpretations, we sample $\K{=}3$ SQL queries per prompt.
 For instance, given the query \textit{``Show me employees with managers in the NYC office''}, two valid interpretations arise:
\begin{enumerate}[nosep,leftmargin=*]
    \item Employees whose \emph{managers} are located in NYC:\\ \texttt{\small WHERE manager.office = 'NYC'}
    \item Employees \emph{themselves} in NYC who have managers:\\ \texttt{\small WHERE employee.office = 'NYC' AND manager\_id IS NOT NULL}
\end{enumerate}
A successful diverse sampler should generate SQL queries that cover both interpretations, rather than producing minor syntactic variations of just one.

\begin{figure}[ht]
\centering
\begin{minipage}[c]{0.65\textwidth}
\includegraphics[width=\columnwidth]{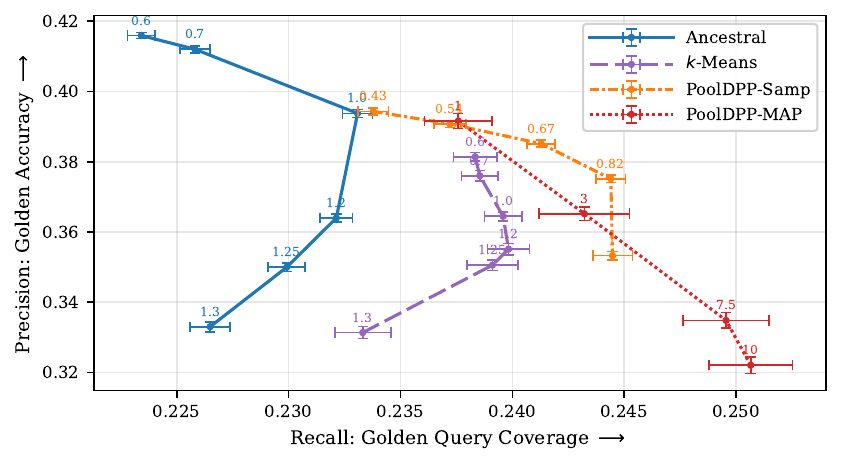}
\end{minipage}
\hfill
\begin{minipage}[c]{.34\textwidth}
\caption{\textbf{Extrinsic evaluation (Ambrosia, $\K{=}3$, $N{=}200$).} Recall vs.\ precision; higher is better. Precision counts a query only if its whole execution result matches a golden result. \algnameMAP sweeps the global temperature $T$ at local sampling temperature $1.5$ and holds the high-recall end of the frontier above $\K$-means; \algnameSamp lies above $\K$-means; ancestral sampling at low local temperature holds the high-precision end. Error bars are $95\%$ $t$-intervals over $100$ pools (Gemma-2-9B, MiniLM-L6).
}
\end{minipage}
\label{fig:ambrosia_pareto}
\end{figure}

\paragraph{Metrics.} We measure \textbf{recall} (\emph{golden-query coverage}) as the fraction of annotated golden intents matched by the $\K$ selected SQL queries, and \textbf{precision} (\emph{golden-query accuracy}) as the fraction of selected queries whose \emph{entire} execution result equals a golden query's result. Both metrics are computed by executing the generated SQL and comparing execution results.

\paragraph{Results.} \Cref{fig:ambrosia_pareto} shows that \algname pushes the recall--precision Pareto frontier above ancestral and the pool-reranking baseline.
\algnameMAP at $T{=}1$ reaches precision $0.392$ at recall $0.238$, above the best $\K$-means precision, and its sweep holds the high-recall end of the frontier. Lowering $T$ trades recall for precision. \algnameSamp lies above $\K$-means across the sweep. Ancestral sampling at local temperature ${\le}0.7$ holds the high-precision, low-recall end: at high local temperature, ancestral sampling produces malformed SQL that fails to execute, so its recall gains stall, while quality reweighting lets the DPP gain recall without that collapse; with a flat quality term (large $T$), the DPP loses precision before it gains recall.\looseness=-1
\Cref{app:qualitative} shows example sets.

\subsection{Experiment 3: Math Reasoning (GSM8K)}
\label{sec:gsm8k}

We evaluate whether modeling diversity improves performance on mathematical reasoning, where generating multiple solution paths can increase the probability of finding a correct answer. We use the GSM8K dataset \citep{gsm8k}, sampling $\K{=}10$ reasoning chains per problem. We measure \textbf{diversity} as the average log-determinant of the embedding Gram matrix and report two accuracy metrics: \textbf{pass@$\K$}, the fraction of problems where at least one generated chain is correct, and \textbf{per-chain accuracy}, the fraction of generated chains that are correct. The two metrics together separate a method that improves coverage by spreading mass across the answer space (high pass@$\K$, possibly lower mean) from one that improves average chain quality.

\begin{figure}[ht]
\begin{minipage}[c]{0.6\textwidth}
\centering
\includegraphics[width=\columnwidth]{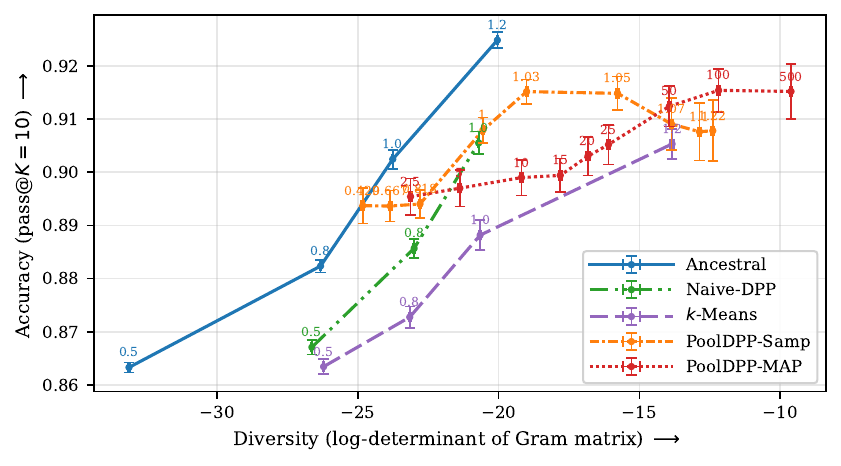}
\end{minipage}
\begin{minipage}[c]{0.4\textwidth}
\caption{\textbf{Math reasoning (GSM8K).} Diversity (log-det Gram matrix) vs.\ pass@$\K{=}10$. Point labels are the local sampling temperature for ancestral sampling, naive DPP, and $\K$-means, and the global temperature $T$ for \algnameSamp and \algnameMAP. \algnameSamp and \algnameMAP hold local sampling temperature $1.2$. At its largest $T$, each is more diverse than any plotted ancestral or $\K$-means setting (log-det $-12.4$ and $-9.6$, against $-20.0$ and $-13.8$). Error bars are $95\%$ $t$-intervals over $100$ pools.
}
\label{fig:gsm8k_pareto}
\end{minipage}
\end{figure}

\paragraph{Results.}
 \Cref{fig:gsm8k_pareto} shows that \algname raises diversity past the plotted ancestral settings and gives up at most $0.01$ pass@$\K$. The best ancestral pass@$\K$ is $0.925 \pm 0.002$, at local sampling temperature $1.2$ and log-det $-20.0$. \algnameSamp reaches $0.915 \pm 0.003$ ($T{=}1.05$, log-det $-15.8$), and \algnameMAP reaches $0.915 \pm 0.005$ ($T{=}500$, log-det $-9.6$). At equal diversity, neither mode has a higher pass@$\K$ than ancestral sampling, so the gain is in diversity, not in coverage.

Pure-diversity selection ($T{=}500$) lowers per-chain accuracy ($0.502$ vs.\ $0.682$), much more than it lowers pass@$\K$ ($0.915$ vs.\ $0.925$), which separates coverage from per-chain quality.
\algnameKM, has the lowest pass@$\K$ of the four methods in \cref{tab:gsm8k_meank_passk}, below both \algname modes, suggesting that DPP repulsion generalizes better to downstream metrics than cluster-based selection. See \cref{app:qualitative:math-reasoning} for example reasoning chains showing how diversity-targeted selection covers competing interpretations of ambiguous problem statements.\looseness=-1

\section{Conclusion}
We presented a method that natively models diversity in language model outputs using determinantal point processes.
On FineWeb, determinantal repulsion alone (naive DPP) already improves on ancestral sampling at matched temperature (log-det $-6.3$ vs.\ $-8.2$ at local temperature $0.6$, with quality $-40.1$ vs.\ $-39.4$), and sweeping the global temperature $T$ traces the rest of the frontier.
On FineWeb, the combined approach extends the quality--diversity frontier (through \algnameMAP; \algnameSamp lies below the $\K$-means frontier at intermediate diversity, \cref{fig:fineweb_pareto}), and on Ambrosia, the precision--recall frontier; on GSM8K, it extends only the diversity axis, whether or not the Qwen3-Embedding-0.6B embedder is given a task instruction (\cref{app:embedder}).

\paragraph{Limitations.}
\algname selects from a pool, so it cannot return an output the proposal never samples: on Ambrosia, the pools contain at most $42\%$ of the gold interpretations on average, and both \algname modes recover $52$--$53\%$ of their pools' coverage (\cref{app:pool-ceiling}). What counts as diverse is fixed by the embedder; the paired-edit probe of \cref{app:embedder} is a cheap check of whether an embedder separates the distinctions a task cares about. The best global temperature $T$ differs across tasks and between the two modes, so choosing an operating point requires task labels or a target diversity level. On FineWeb and GSM8K, the pools approximate the global-temperature target poorly away from $T{=}1$ (ESS below $\K$ for most prompts, \cref{sec:ablations}). Finally, the bound of \cref{cor:kl-sum} is evaluated only on instances built from our pools, so its values are not a bound on the sampler's KL for our runs (\cref{sec:analysis}).

\begin{ack}
The authors thank \href{https://chanr0.github.io/}{Robin Shing Moon Chan}, \href{https://benlipkin.github.io/}{Ben Lipkin}, and other members of the \href{https://genlm.org/}{GenLM consortium} for helpful discussions that improved this work.  We also thank the anonymous reviewers for their valuable feedback, including suggesting the $\K$-means baseline and asking us to evaluate our earlier KL bound numerically, which led to the sharper bound of \cref{cor:kl-sum}. Manuel was supported by Fundación Barrié and the Max Planck ETH Center for Learning Systems during parts of this work.\looseness=-1
\end{ack}

\paragraph{AI-use disclosure.} We used agentic coding utilities to edit and draft parts of the text, to write analysis code, to double-check proofs, and to help write the Lean formalization of \cref{app:lean}. The authors reviewed every change and take full responsibility for this paper's content.\looseness=-1

\bibliographystyle{plainnat}
\bibliography{main}

\appendix
\onecolumn

\etocdepthtag.toc{appendix}
\etocsettagdepth{mainmatter}{none}
\etocsettagdepth{appendix}{subsection}
\etocsetnexttocdepth{subsection}
\etocsettocstyle
{\section*{Contents of the Appendix}}
{}
\tableofcontents

\clearpage
\section{Deferred Proofs}
\label{app:PoolDPP-proofs}

This appendix gives the proofs deferred from the main text. We use the main-text notation: $\omega \sim \mu$ is the sketch's internal randomness (for \algname, the pool), $\fhatom$ and $\Zhat$ are the random unnormalized weight and normalizer, $\pihat = \fhatom/\Zhat$ is the sketched distribution, and $\pibar$ is the marginal output law of \cref{eq:marginal-output-law}. As there, $\Zhat > 0$ almost surely; \cref{fn:wlog} reduces the general case to this one.\looseness=-1

\subsection{Lean formalization}
\label{app:lean}
We formalized \cref{thm:var-kl-mixture,thm:zfree,cor:kl-sum} and \cref{prop:PoolDPP-well-behaved} in Lean~4 with Mathlib, and the code is available in the repository. 

\subsection{Redrawing degenerate pools and the constant \texorpdfstring{$\alphaMK$}{alpha}}
\label{app:redraw}
\paragraph{Redraw.}
If $\Pr_\omega[\Zhat = 0] > 0$, then $\fhatom \equiv 0$ on those $\omega$ (the weights are nonnegative), and \algname redraws $\omega$. For a well-behaved sketch, $\gamma \defeq \Pr_\omega[\Zhat > 0]$ is positive, since $\E_\omega[\Zhat] = \normalizer > 0$. So redrawing until $\Zhat > 0$, which is rejection sampling, ends with probability $1$, and it runs the sketch $(\mu_+, \fhat)$, where $\mu_+$ is $\mu$ conditioned on $\Zhat > 0$. This sketch has $\Zhat > 0$ almost surely, it is well-behaved for $f/\gamma$, which defines the same target $\pDPP$, and $\E_{\mu_+}[\Zhat^2]/(\normalizer/\gamma)^2 = \gamma\,\E_\mu[\Zhat^2]/\normalizer^2$. Any other rule for these $\omega$ (a \emph{fallback}: some measurable distribution over sets) has an output law at least $\gamma$ times the redraw law $\pibar_+$ pointwise, so compared with $\pibar_+$, it multiplies $1 + \chi^2$ by at most $1/\gamma$ and adds at most $\log(1/\gamma)$ to $\KL$. Either way, \cref{thm:var-kl-mixture} for $(\mu_+, \fhat)$ gives its bounds for $(\mu, \fhat)$, with $\Var_\omega(\Zhat)$ under $\mu$.
With embeddings in general position, which holds generically for neural embedders, $\Zhat > 0$ exactly when the pool contains at least $\K$ distinct strings of positive quality. On Ambrosia, a prompt whose pool is degenerate takes its pool from the next of our pools that is not.

A pool has $\Zhat = 0$ exactly when its weighted embeddings span fewer than $\K$ directions (\cref{prop:pushforward}). On Ambrosia, $25$ of $10{,}000$ (prompt, pool) pairs at local sampling temperature $1$ are degenerate.

\paragraph{The constant $\alphaMK$.}
Although $\alphaMK$ cancels in $\pihat$ (\cref{prop:pushforward}), \cref{thm:var-kl-mixture} requires $\E_\omega[\fhatom(\set)] = f(\set)$, and $\alphaMK$ is the constant that gives it (\cref{prop:PoolDPP-well-behaved}). The $1/N$ in each column of $\Bapprox$ (\cref{eq:B-matrix}) normalizes as though all $N^\K$ ordered index tuples contributed to \cref{eq:PoolDPP-fhat}, but only the $(N)_\K$ tuples with distinct indices do, since a repeated index repeats a column and gives determinant $0$. Thus, $1/\alphaMK$ is the probability that $\K$ uniform draws from $[N]$ are all distinct, and $\alphaMK \to 1$ as $N \to \infty$.

\subsection{Proof of \cref{thm:var-kl-mixture}}
\label{app:proof-var-kl-mixture}

\thmVarKL*

The bound is general---it does not require i.i.d.\ proposals---and reduces to Theorem 2.1.1 of \citet{huggins-masters-thesis} when specialized to that case.

\begin{proof}\vphantom{.}
\begin{enumerate}
\item \label{step:marginal-is-pibar} Define the joint distribution $\Ptilde$ of a pair $(\omega, \boldsymbol{S})$ by first drawing the sketch randomness $\omega \sim \mu$ and then drawing $\boldsymbol{S} \sim \pihat(\cdot)$; throughout, $\boldsymbol{S}$ denotes this random outcome and $\set$ a fixed value. Equivalently,
\begin{align}\label{eq:joint-dist-def}
\Ptilde(\omega, \set) = \mu(\omega) \pihat(\set)
\end{align}
Under $\Ptilde$, the marginal law of $\boldsymbol{S}$ is exactly $\pibar$:
\begin{align}\label{eq:marginal-is-pibar}
\Ptilde(\boldsymbol{S} = \set) = \E_\omega\mleft[\pihat(\set)\mright] = \pibar(\set)
\end{align}

\item Define the nonnegative random variable
\begin{align}\label{eq:A-def}
A(\omega) \defeq \frac{\Zhat}{\normalizer}
\end{align}
Note that $A$ depends only on $\omega$. Using unbiasedness of $\fhatom$,
\begin{subequations}\label{eq:E-Zhat-equals-Z}
\begin{align}
\E_\omega[\Zhat]
&= \E_\omega\mleft[\sum_{\set} \fhatom(\set)\mright]  \\
&= \sum_{\set} \E_\omega[\fhatom(\set)] && \justification{Tonelli, since $\fhatom(\set) \ge 0$} \\
&= \sum_{\set} f(\set) && \justification{$\fhatom$ is unbiased (well-behaved)} \\
&= \normalizer
\end{align}
\end{subequations}
hence $\E_\omega[A(\omega)] = 1$.

\item \label{step:A-indicator-equals-pi} Crucially, for any fixed $\set$,
\begin{subequations}
\begin{align}
&\E_{(\omega, \boldsymbol{S}) \sim \Ptilde}\mleft[A(\omega) \indicator{\boldsymbol{S} = \set}\mright] \notag \\
&= \E_{\omega \sim \mu}\mleft[A(\omega) \pihat(\set)\mright] && \justification{given $\omega$, $\boldsymbol{S} \sim \pihat$} \\
&= \E_\omega\mleft[\frac{\Zhat}{\normalizer} \cdot \frac{\fhatom(\set)}{\Zhat}\mright] && \justification{definitions of $A$ and $\pihat$} \\
&= \frac{1}{\normalizer} \E_\omega[\fhatom(\set)] && \justification{canceling $\Zhat > 0$} \\
&= \frac{f(\set)}{\normalizer} && \justification{unbiasedness of $\fhatom$} \\
&= \pDPP(\set) \label{eq:A-indicator-equals-pi}
\end{align}
\end{subequations}

\item For any $\set$ with $\pibar(\set) > 0$, by definition of conditional expectation,
\begin{align}\label{eq:ratio-as-condexp}
\E_{(\omega, \boldsymbol{S}) \sim \Ptilde}\mleft[A(\omega) \,\middle|\, \boldsymbol{S} = \set\mright]
&= \frac{\E_{(\omega, \boldsymbol{S}) \sim \Ptilde}\mleft[A(\omega) \indicator{\boldsymbol{S} = \set}\mright]}{\Ptilde(\boldsymbol{S} = \set)}
= \frac{\pDPP(\set)}{\pibar(\set)}
\end{align}
where the numerator uses step~\ref{step:A-indicator-equals-pi} and the denominator uses \cref{eq:marginal-is-pibar} from step~\ref{step:marginal-is-pibar}.

\item Let $r(\set) \defeq \pDPP(\set)/\pibar(\set)$ be the likelihood ratio; it is defined $\pibar$-almost everywhere, since \eqref{eq:A-indicator-equals-pi} forces $\pDPP(\set) = 0$ whenever $\pibar(\set) = 0$, that is, $\pDPP \ll \pibar$. Then, $\E_{\pibar}[r] = \sum_{\set \in \mathcal{D}} \pDPP(\set) = 1$ and
\begin{align}\label{eq:chisq-as-variance}
\chi^2(\pDPP \| \pibar)
&= \E_{\pibar}\mleft[(r-1)^2\mright]
= \E_{\pibar}[r^2] - 1
\end{align}

\item By \eqref{eq:ratio-as-condexp} and Jensen's inequality applied to $g(x) = x^2$,\footnote{Conditional Jensen: for any random variables $X, Y$, $\E[(\E[X \mid Y])^2] \le \E[\E[X^2 \mid Y]] = \E[X^2]$.}
\begin{align}\label{eq:jensen-bound}
\E_{\set \sim \pibar}[r(\set)^2]
&= \E_{\set \sim \pibar}\mleft[\bigl(\E_{\Ptilde}[A(\omega) \mid \boldsymbol{S} = \set]\bigr)^2\mright] \notag \\
&\le \E_{\Ptilde}[A(\omega)^2] && \justification{Jensen on $g(x) = x^2$, tower property} \notag \\
&= \E_\omega\mleft[\Big(\frac{\Zhat}{\normalizer}\Big)^{2}\mright] && \justification{$A$ depends only on $\omega$} \notag \\
&= \frac{\E_\omega[\Zhat^2]}{\normalizer^2}
= 1 + \frac{\Var_\omega(\Zhat)}{\normalizer^2} && \justification{$\E_\omega[\Zhat] = \normalizer$ from step 2}
\end{align}
Substituting \eqref{eq:jensen-bound} into \eqref{eq:chisq-as-variance} gives the $\chi^2$ bound in \eqref{eq:var-kl}.

\item For the KL bound, we use $\KL(\pDPP \| \pibar) \le \log(1 + \chi^2(\pDPP \| \pibar))$:\footnote{For distributions $P, Q$ on a countable set with $r(\set) \defeq P(\set)/Q(\set)$, Jensen on the concave $\log$ gives $\KL(P \| Q) = \sum_\set Q(\set) r(\set) \log r(\set) \le \log \sum_\set Q(\set) r(\set)^2 = \log(1 + \chi^2(P \| Q))$.}
\begin{align}\label{eq:kl-renyi-bound}
\KL(\pDPP \| \pibar)
&\le \log \E_{\pibar}[r^2]
= \log\bigl(1 + \chi^2(\pDPP \| \pibar)\bigr)
\end{align}
Combining \eqref{eq:kl-renyi-bound} with \eqref{eq:jensen-bound} gives the first KL inequality in \eqref{eq:var-kl}; the second follows from $\log(1+x) \le x$.\looseness=-1
\end{enumerate}
\end{proof}

\clearpage
\subsection{Proof of \cref{prop:pushforward}}
\label{app:proof-pushforward}

\begin{restatable}[\algnameSamp samples the sketched distribution]{proposition}{propPushforward}
\label{prop:pushforward}
Fix a pool $\omega$. Then, $\Zhat = \alphaMK \cdot e_\K(\eig(\Capprox))$, so $\Zhat = 0$ exactly when $\mathrm{rank}(\Capprox) < \K$. If $\Zhat > 0$, then the string set $\set = \{\yy^{(n)}\}_{n \in I}$ returned by \algnameSamp satisfies $\set \sim \pihat$.
\end{restatable}
\begin{proof}
Throughout the proof, the pool $\omega = (\yy^{(1)}, {\ldots}, \yy^{(N)})$ is fixed, and all remaining randomness is due to \cref{alg:dual-kdpp}. For the sampling claim, let $\Zhat > 0$; the derivation of \cref{eq:Zhat-esp} below does not use this.
Invoked with the feature matrix $\Bapprox \in \R^{D \times N}$, \cref{alg:dual-kdpp} is an exact sampler for the $\K$-DPP whose base set is the set of column indices $[N]$ and whose kernel is $\Lapprox = \Bapprox^\top \Bapprox$ \citep[Alg.~3]{kulesza-dpp}; it returns an index set $I \in \binom{[N]}{\K}$ with probability (we write $\Pr_{\Lapprox}$ for this law)
\begin{align}\label{eq:index-kdpp}
\Pr_{\Lapprox}\mleft[I\mright] = \frac{\det(\Lapprox_I)}{\sum_{I' \in \binom{[N]}{\K}} \det(\Lapprox_{I'})}
\end{align}
By the standard identity relating the sum of principal $\K \times \K$ minors of a symmetric matrix to the elementary symmetric polynomial of its eigenvalues \citep[\S 5.2.2]{kulesza-dpp}, and because $\Lapprox = \Bapprox^\top \Bapprox$ and $\Capprox = \Bapprox \Bapprox^\top$ share the same nonzero eigenvalues, the denominator of \cref{eq:index-kdpp} equals $e_\K(\eig(\Lapprox)) = e_\K(\eig(\Capprox))$.

Now push \cref{eq:index-kdpp} forward through the labeling map $\iota(I) \defeq \{\yy^{(n)} \mid n \in I\}$, which sends the sampled index set to the string set that \algnameSamp returns. Two observations:
\begin{enumerate}
\item \emph{Collisions carry no mass.} If $I$ contains indices $n \neq n'$ with $\yy^{(n)} = \yy^{(n')}$, then columns $n$ and $n'$ of $\Bapprox$ are identical (the weight $w$ and the embedding $\embed$ depend only on the string), so $\det(\Lapprox_I) = 0$ and $\Pr_{\Lapprox}[I] = 0$. Hence, every index set with positive probability satisfies $|\iota(I)| = \K$.
\item \emph{The index sum is partitioned.} Each index set $I \in \binom{[N]}{\K}$ with $|\iota(I)| = \K$ appears in the inner sum of \cref{eq:PoolDPP-fhat} for exactly one string set, namely $\set = \iota(I)$; index sets with $|\iota(I)| < \K$ appear in none, and contribute $0$ wherever they would appear by observation 1.
\end{enumerate}
Therefore, for any $\set \in \binom{\strings}{\K}$,
\begin{align}
\Pr_{\Lapprox}\mleft[\iota(I) = \set\mright]
= \frac{\sum_{I : \iota(I) = \set} \det(\Lapprox_I)}{\sum_{I' \in \binom{[N]}{\K}} \det(\Lapprox_{I'})}
= \frac{\alphaMK^{-1} \fhatom(\set)}{\alphaMK^{-1} \Zhat}
= \pihat(\set)
\end{align}
where the numerator uses the definition of $\fhatom$ in \cref{eq:PoolDPP-fhat}, and the denominator uses
\begin{align}\label{eq:Zhat-esp}
\Zhat
=\!\! \sum_{\set \in \binom{\strings}{\K}} \!\! \fhatom(\set)
= \alphaMK\!\! \sum_{\set \in \binom{\strings}{\K}} \sum_{I : \iota(I) = \set} \det(\Lapprox_I)
= \alphaMK\!\! \sum_{I \in \binom{[N]}{\K}} \!\! \det(\Lapprox_I)
= \alphaMK \cdot e_\K(\eig(\Capprox))
\end{align}
with the third equality by observation 2. This proves the sampling claim and the identity for $\Zhat$, which holds for every pool. Since $\Capprox$ is positive semidefinite, its eigenvalues are nonnegative, so $e_\K(\eig(\Capprox)) = 0$ exactly when fewer than $\K$ of them are positive, that is, when $\mathrm{rank}(\Capprox) < \K$. Note that \cref{eq:Zhat-esp} also shows that $\Zhat$ is computable in $\bigo(D^3 + D\K)$ time from the eigenvalues of $\Capprox$ (see \cref{esp}), although \algname needs only whether it is zero.
\end{proof}

\clearpage
\subsection{Proof of \cref{prop:PoolDPP-well-behaved}}
\label{app:proof-PoolDPP-well-behaved}
\begin{restatable}[\PoolDPP is well-behaved]{proposition}{propSketchWellBehaved}
\label{prop:PoolDPP-well-behaved}
For any $\K \ge 1$, $N \ge \K$, and proposal $\proposal$ with $\quality \ll \proposal$, the construction in \cref{eq:PoolDPP-fhat} is unbiased: $\E_\omega[\fhatom(\set)] = \det(\LL_\set) \indicator{|\set|=\K}$.
If, additionally, $\rho < \infty$, then $\E_\omega[\Zhat^2] < \infty$ and \PoolDPP is well-behaved.
\end{restatable}
\begin{proof}
The sketch randomness is $\omega = (\yy^{(1)}, {\ldots}, \yy^{(N)})$ with $\yy^{(n)} \simIID \proposal$. We verify the two requirements of being well-behaved (\cref{thm:var-kl-mixture}'s preconditions): unbiasedness and a finite second moment.

\begin{enumerate}
\item \emph{Unbiasedness: $\E_\omega[\fhatom(\set)] = f(\set)$ for every $\set$.}

If $|\set| \neq \K$, both sides vanish: $f(\set) = 0$ by the size indicator, and $\fhatom(\set) = 0$ because any index set $I$ with $|I| = \K$ whose image is $\set$ must repeat a string, which repeats a column of $\Bapprox$ and forces $\det(\Lapprox_I) = 0$.

So fix $\set = \{\yy_1, {\ldots}, \yy_\K\} \in \binom{\strings}{\K}$, a set of $\K$ \emph{distinct} strings, whence $f(\set) = \det(\LL_\set)$.
Fix also an index set $I \in \binom{[N]}{\K}$ and consider the event $\{\yy^{(n)} \mid n \in I\} = \set$.
Because $|I| = |\set| = \K$, this event holds if and only if the $\K$ i.i.d.\ draws at the positions of $I$ realize the $\K$ elements of $\set$ under some bijection; there are $\K!$ bijections, each occurring with probability $\prod_{j=1}^{\K} \proposal(\yy_j)$, so
\begin{align}\label{eq:event-prob}
\Pr_\omega\mleft[\{\yy^{(n)} \mid n \in I\} = \set\mright] = \K! \prod_{j=1}^{\K} \proposal(\yy_j)
\end{align}
On this event, $\det(\Lapprox_I)$ takes the same value under every bijection: the columns of $\Bapprox$ indexed by $I$ are $\{\sqrt{w(\yy_j)/N} \embed(\yy_j)\}_{j=1}^{\K}$ up to a permutation, which leaves the Gram determinant unchanged, and factoring $\sqrt{w(\yy_j)/N}$ out of each row and column,
\begin{align}\label{eq:det-on-event}
\det(\Lapprox_I)
= \det\mleft(\mleft[\tfrac{\sqrt{w(\yy_i) w(\yy_j)}}{N} \embed(\yy_i)^\top \embed(\yy_j)\mright]_{i,j=1}^{\K}\mright)
= N^{-\K} \prod_{j=1}^{\K} w(\yy_j) \cdot \det(\Embs_\set^\top \Embs_\set)
\end{align}
Combining \cref{eq:event-prob,eq:det-on-event} with the cancellation $\proposal(\yy) w(\yy) = \quality(\yy)$ and the determinant decomposition $\det(\LL_\set) = \prod_{j=1}^{\K} \quality(\yy_j) \cdot \det(\Embs_\set^\top \Embs_\set)$ from \cref{sec:dpp},
\begin{align}\label{eq:per-index-set}
\E_\omega\mleft[\det(\Lapprox_I) \indicator{\{\yy^{(n)} \mid n \in I\} = \set}\mright]
&= \K! \prod_{j=1}^{\K} \proposal(\yy_j) N^{-\K} \prod_{j=1}^{\K} w(\yy_j) \cdot \det(\Embs_\set^\top \Embs_\set) \notag \\
&= \frac{\K!}{N^{\K}} \det(\LL_\set)
\end{align}
This value does not depend on $I$, and there are $\binom{N}{\K}$ index sets, so
\begin{align}\label{eq:PoolDPP-unbiased-derivation}
\E_\omega[\fhatom(\set)]
&= \alphaMK \sum_{I \in \binom{[N]}{\K}} \E_\omega\mleft[\det(\Lapprox_I) \indicator{\{\yy^{(n)} \mid n \in I\} = \set}\mright] && \justification{def.\ of $\fhatom$, \cref{eq:PoolDPP-fhat}} \notag \\
&= \alphaMK \cdot \binom{N}{\K} \cdot \frac{\K!}{N^{\K}} \cdot \det(\LL_\set) && \justification{\cref{eq:per-index-set}, $\binom{N}{\K}$ index sets} \notag \\
&= \alphaMK \cdot \frac{(N)_\K}{N^{\K}} \cdot \det(\LL_\set) && \justification{$\binom{N}{\K} \cdot \K! = (N)_\K$} \notag \\
&= \det(\LL_\set)
= f(\set) && \justification{$\alphaMK = N^{\K}/(N)_\K$}
\end{align}
The counting in \cref{eq:PoolDPP-unbiased-derivation} is exactly the mismatch described in \cref{sec:analysis}: the raw index sum recovers only the fraction $(N)_\K / N^{\K}$ of the mass that the $1/N$-normalization of $\Bapprox$ presumes---the collision-free fraction of ordered index tuples---and $\alphaMK$ restores the remainder.

Summing \cref{eq:PoolDPP-unbiased-derivation} over $\set$ then gives $\E_\omega[\Zhat] = \sum_{\set \in \mathcal{D}} f(\set) = \normalizer$.

\item \emph{Finite second moment: $\E_\omega[\Zhat^2] < \infty$.}

\Cref{thm:zfree} derives the finite second moment from  $\rho < \infty$, which holds whenever $\E_\proposal[w^2] < \infty$ (\cref{app:zfree}). With $\rho < \infty$, the hypothesis of the proposition, the proof of \cref{eq:kl-closed-form} also gives the explicit bound $\E_\omega[\Zhat^2]/\normalizer^2 \le e^{\theta} - 1 + a_0$, with $\theta = \K^2 \rho/(N - \K + 1)$.
\end{enumerate}

\end{proof}

\clearpage
\subsection{Proof of \cref{rem:rao-blackwell}}
\label{app:proof-rao-blackwell}

\begin{remark}[Rao--Blackwellization interpretation]
\label{rem:rao-blackwell}
The sketch in \cref{eq:PoolDPP-fhat} admits a Rao--Blackwell interpretation.
Consider the naive importance sampler that draws $\K$ strings $\yy_1, {\ldots}, \yy_\K$ i.i.d.\ from the proposal, i.e., one draw of a $\K$-tuple from the product distribution $\proposal^{\otimes \K}$, and estimates $f(\set)$ by
$\widetilde{f}(\set) \defeq \frac{1}{\K!} \prod_{j=1}^{\K} w(\yy_j) \cdot \det(\Embs_\set^\top \Embs_\set) \cdot \indicator{\{\yy_1, {\ldots}, \yy_\K\} = \set}$,
which is unbiased by the same cancellation $\proposal \cdot w = \quality$ as in \cref{prop:PoolDPP-well-behaved}.
One way to realize the $\K$ i.i.d.\ draws is to draw the pool of $N \ge \K$ i.i.d.\ samples first, and then read off the strings at $\K$ pool positions chosen uniformly at random \emph{without replacement}.
The sketched weight $\fhatom(\set)$ is precisely the conditional expectation of $\widetilde{f}(\set)$ given the pool, with $\alphaMK$ emerging as the ratio between the $1/N^\K$ normalization inside $\Bapprox$ and the $1/(N)_\K$ probability of each ordered position choice.
The Rao--Blackwell theorem then guarantees variance reduction: $\Var_\omega(\fhatom(\set)) \le \Var_{\proposal^{\otimes \K}}(\widetilde{f}(\set))$ for every $\set$.
\Cref{thm:zfree} quantifies the gain: the naive sampler is \algname with $N = \K$, where $a_\K = 1$ and the relative variance is $\Var_{\proposal^{\otimes \K}}(\incl{\K})$; in a pool of size $N$, this term enters with weight $a_\K = 1/\binom{N}{\K}$. By \cref{eq:hoeffding-bounds}, the relative variance at pool size $N$ is at most $(\K/N)\Var_{\proposal^{\otimes \K}}(\incl{\K})$, a factor of at least $N/\K$ below the naive sampler's.\looseness=-1
\end{remark}

\begin{proof}
Let $\omega = (\yy^{(1)}, {\ldots}, \yy^{(N)}) \simIID \proposal$ be the pool, and---independently of the pool---draw an ordered tuple $(n_1, {\ldots}, n_\K)$ of \emph{distinct} positions, uniformly among the $(N)_\K$ such tuples.
Reading off $\widetilde{\yy}_j \defeq \yy^{(n_j)}$ yields $\K$ strings that are marginally i.i.d.\ from $\proposal$: the strings at any $\K$ fixed distinct positions of an i.i.d.\ pool are i.i.d., and the positions are chosen independently of the pool.
This is therefore a valid realization of the draw from $\proposal^{\otimes \K}$ used by the naive sampler of \cref{rem:rao-blackwell}.

Fix $\set \in \binom{\strings}{\K}$ and take the conditional expectation of the naive weight $\widetilde{f}(\set)$ over the position choice, holding the pool fixed.
Grouping the $(N)_\K$ ordered position tuples by their underlying index set $I \in \binom{[N]}{\K}$---$\K!$ orderings each, all yielding the same summand---we obtain
\begin{subequations}\label{eq:rb-verification}
\begin{align}
\E\mleft[\widetilde{f}(\set) \,\middle|\, \omega\mright]
&= \frac{1}{(N)_\K} \sum_{(n_1, {\ldots}, n_\K)} \frac{1}{\K!} \prod_{j=1}^{\K} w^{(n_j)} \notag \\
&\qquad\qquad \cdot \det(\Embs_\set^\top \Embs_\set) \cdot \indicator{\{\yy^{(n_1)}, {\ldots}, \yy^{(n_\K)}\} = \set} \\
&= \frac{1}{(N)_\K} \sum_{I \in \binom{[N]}{\K}} \prod_{n \in I} w^{(n)} \notag \\
&\qquad\qquad \cdot \det(\Embs_\set^\top \Embs_\set) \cdot \indicator{\{\yy^{(n)} \mid n \in I\} = \set} \\
&= \frac{N^{\K}}{(N)_\K} \sum_{\substack{I \in \binom{[N]}{\K} \\ \{\yy^{(n)} \mid n \in I\} = \set}} \det(\Lapprox_I) && \justification{\cref{eq:det-on-event}} \\
&= \alphaMK \smashoperator{\sum_{\substack{I \subseteq [N], |I|=\K \\ \{\yy^{(n)} \mid n \in I \} = \set}}} \det(\Lapprox_I)
\ \ = \fhatom(\set) && \justification{def.\ of $\fhatom$, \cref{eq:PoolDPP-fhat}}
\end{align}
\end{subequations}
So the sketched weight is precisely the conditional expectation of the naive weight given the pool, with $\alphaMK = N^{\K}/(N)_\K$ arising as the ratio between the $N^{-\K}$ normalization inside $\det(\Lapprox_I)$ (via the $1/N$ factor in each column of $\Bapprox$) and the $1/(N)_\K$ probability of each ordered position choice.
Since conditioning preserves the mean and cannot increase variance (law of total variance), $\E_\omega[\fhatom(\set)] = \E_{\proposal^{\otimes \K}}[\widetilde{f}(\set)] = f(\set)$ and $\Var_\omega(\fhatom(\set)) \le \Var_{\proposal^{\otimes \K}}(\widetilde{f}(\set))$, as claimed.
\end{proof}

\clearpage
\subsection{Proof of \cref{thm:zfree}}
\label{app:zfree}

Throughout, $\falling{n}{k} \defeq n(n{-}1)\cdots(n{-}k{+}1)$.

\paragraph{Overlap intuition.}
For $\K=2$, $\Zhat$ averages contributions from all pool pairs.
The pairs $\{1,2\}$ and $\{3,4\}$ use independent draws.
The pairs $\{1,2\}$ and $\{1,3\}$ share only the first draw: once it is fixed, their remaining draws are independent, so their covariance is the variance of their common conditional mean.
For general $\K$, the same calculation groups pairs of subsets by how many positions they share.

\paragraph{Remarks on the statement.}
The coefficients $a_\overlap$ are the hypergeometric law of the overlap $|I \cap J|$ between two independent uniform $\K$-subsets of $[N]$, with mean $\K^2/N$. Since $\Zhat$ is an average of one symmetric function over all size-$\K$ subsets of pool positions, and the pool is i.i.d., two terms covary only through the positions they share; the pool size enters only through the probability of each overlap.
Since $\Var_{\proposal^{\otimes \overlap}}(\incl{\overlap})/\overlap$ is nondecreasing in $\overlap$ \citep{hoeffding1948class}, at every pool size
\begin{align}\label{eq:hoeffding-bounds}
\frac{\K^2}{N} \Var_\proposal(\incl{1})
\le
\frac{\Var_\omega(\Zhat)}{\normalizer^2}
\le
\frac{\K}{N} \Var_{\proposal^{\otimes \K}}(\incl{\K})
\end{align}
and the lower bound is attained up to $\bigo(1/N^2)$ as $N \to \infty$ (\cref{app:zfree}).
At $\K = 1$, \cref{eq:zfree} is the importance-sampling relative variance $\Var_\proposal(w)/(N \E_\proposal[w]^2)$. In general, $\incl{1} = \pi_1/\proposal$ is at most $C_0 w$ for a constant $C_0$ set by the spectrum of $\CC$, so $\rho \le C_0^2\, \E_\proposal[w^2]$ (\cref{app:zfree}).
To interpret the numerator divided by $\falling{\K}{\overlap}$, draw a target set and choose $\overlap$ of its members uniformly without replacement, in order: the ratio is the probability of obtaining the tuple $(\yy_1,{\ldots},\yy_\overlap)$.
Thus, $\incl{\overlap}$ compares this target tuple distribution with $\overlap$ independent proposal draws, has mean $1$, and is unchanged by rescaling $\quality$.

\thmZfree*
\begin{proof}
We prove the identity \labelcref{eq:zfree} first, and then the bound \labelcref{eq:kl-closed-form}.

\paragraph{Proof of the identity \labelcref{eq:zfree}.}
\textbf{Subset average.}
Write $\boldsymbol{Y}_I=(\yy^{(n)})_{n\in I}$ and $\Embs=[\embed(\yy_1),{\ldots},\embed(\yy_\K)]$.
Pulling the factor $\sqrt{w(\yy^{(n)})/N}$ of each column of $\Bapprox$ out of each determinant in \cref{eq:PoolDPP-fhat} and using
$\alphaMK N^{-\K}=1/[\binom{N}{\K}\K!]$ gives
\begin{align}
h(\yy_1,{\ldots},\yy_\K)
&\defeq \frac{1}{\K!}\prod_{j=1}^{\K}w(\yy_j)\det(\Embs^\top\Embs)
\notag \\
\Zhat
&=\frac{1}{\binom{N}{\K}}\sum_{I\subseteq[N],|I|=\K}h(\boldsymbol{Y}_I)
\label{eq:ustat}
\end{align}
The function $h$ is symmetric and vanishes on repeated strings, because a repeated string repeats a column of $\Embs$.

\textbf{Conditional mean.}
Fix distinct $\yy_1,{\ldots},\yy_\overlap$ and let $\set_0=\{\yy_1,{\ldots},\yy_\overlap\}$.
Averaging over the remaining $\K-\overlap$ independent proposal draws cancels their proposal factors, since $\proposal w=\quality$.
Each completed set containing $\set_0$ has $(\K-\overlap)!$ orderings of its remaining members, so
\begin{align}
\E_{\boldsymbol{Y}_{\overlap+1}, {\ldots}, \boldsymbol{Y}_\K \simIID \proposal}\mleft[h(\yy_1, {\ldots}, \yy_\overlap, \boldsymbol{Y}_{\overlap+1}, {\ldots}, \boldsymbol{Y}_\K)\mright]
&= \frac{(\K-\overlap)!}{\K!\prod_{j\le\overlap}\proposal(\yy_j)}
\sum_{\set \supseteq T, |\set| = \K} \det(\LL_\set)
\notag \\
&=\normalizer\incl{\overlap}(\yy_1,{\ldots},\yy_\overlap)
\label{eq:proj-is-incl}
\end{align}
Both sides are zero on repeated strings.
The ordered-tuple interpretation of \cref{eq:incl-def} gives $\E_{\proposal^{\otimes \overlap}}[\incl{\overlap}]=1$, hence $\E_{\proposal^{\otimes \K}}[h]=\normalizer$.

\textbf{Covariances and overlap counts.}
If $I\cap J=\emptyset$, the contributions are independent and their covariance is zero.
Otherwise, condition on their $\overlap=|I\cap J|$ shared draws.
The remaining draws are independent, and both conditional means equal $\normalizer\incl{\overlap}(\boldsymbol{Y}_{I\cap J})$ by \cref{eq:proj-is-incl}, giving
\begin{align}
\mathrm{Cov}\bigl(h(\boldsymbol{Y}_I),h(\boldsymbol{Y}_J)\bigr)
= \Var_{\boldsymbol{Y}_{I \cap J}}\bigl(\E[h(\boldsymbol{Y}_I) \mid \boldsymbol{Y}_{I \cap J}]\bigr)
=\normalizer^2\Var_{\proposal^{\otimes \overlap}}(\incl{\overlap})
\end{align}
For each $I$, choose $\overlap$ shared positions from $I$ and $\K-\overlap$ other positions from its complement: there are $\binom{\K}{\overlap}\binom{N-\K}{\K-\overlap}$ choices of $J$.
Expanding the variance in \cref{eq:ustat} and grouping these pairs therefore gives
\begin{align}
\frac{\Var_\omega(\Zhat)}{\normalizer^2}
&=\frac{\sum_{I,J}\mathrm{Cov}(h(\boldsymbol{Y}_I),h(\boldsymbol{Y}_J))}{\binom{N}{\K}^2\normalizer^2}
\notag \\
&=\sum_{\overlap=1}^{\K}a_\overlap\Var_{\proposal^{\otimes \overlap}}(\incl{\overlap})
\end{align}
where $I$ and $J$ each range over $\binom{[N]}{\K}$.
This is Hoeffding's variance formula for $U$-statistics \citep{hoeffding1948class}, with the conditional means given by \cref{eq:proj-is-incl}.
 {This computation assumes $\E_{\proposal^{\otimes \K}}[\incl{\K}^2] < \infty$, which makes every covariance finite, since $\incl{\overlap}$ is a conditional mean of $\incl{\K}$ and so $\Var_{\proposal^{\otimes \overlap}}(\incl{\overlap}) \le \Var_{\proposal^{\otimes \K}}(\incl{\K})$. If $\E_{\proposal^{\otimes \K}}[\incl{\K}^2] = \infty$, both sides are $+\infty$: since $h \ge 0$, $\E_\omega[\Zhat^2] \ge \binom{N}{\K}^{-2} \sum_{I \in \binom{[N]}{\K}} \E_{\boldsymbol{Y}_I}[h(\boldsymbol{Y}_I)^2] = \infty$, and $a_\K = 1/\binom{N}{\K} > 0$.}

\textbf{Bounds from Hoeffding's ordering.} \citet{hoeffding1948class} shows that $\Var_{\proposal^{\otimes \overlap}}(\incl{\overlap})/\overlap$ is nondecreasing in $\overlap$, so $\overlap \Var_\proposal(\incl{1}) \le \Var_{\proposal^{\otimes \overlap}}(\incl{\overlap}) \le (\overlap/\K)\Var_{\proposal^{\otimes \K}}(\incl{\K})$. Since $\sum_{\overlap=0}^{\K} \overlap a_\overlap = \K^2/N$, the mean of the hypergeometric overlap, \cref{eq:zfree} gives \cref{eq:hoeffding-bounds}. As $N \to \infty$ for a fixed instance, $a_1 = \K^2/N + \bigo(1/N^2)$ and $a_\overlap = \bigo(N^{-\overlap})$, so the lower bound is attained up to $\bigo(1/N^2)$ when the $\Var_{\proposal^{\otimes \overlap}}(\incl{\overlap})$ are finite.

\paragraph{Proof of the bound \labelcref{eq:kl-closed-form}.}
A $\K$-DPP is a determinantal measure conditioned on $|\set| = \K$;
determinantal measures are strongly Rayleigh, the property survives
conditioning on the number of elements, and strongly Rayleigh measures are
negatively associated \citep{borcea2009negative,anari2016montecarlo}.
We use negative association only for inclusion events, and the paragraph ``Negative correlation without strongly Rayleigh measures'' below proves that case directly, with linear algebra and Newton's inequality.
These results are stated for finite base sets, and we extend the inequality below to $\strings$ by truncation. Fix a finite set $\set_0$ of strings. Restricted to the first $n$ strings of a fixed enumeration of $\strings$ and renormalized, $\pDPP$ is a $\K$-DPP on a finite base set once these $n$ strings include $\set_0$ and some $\K$-set with $\det(\LL_\set) > 0$, so the inequality holds for the restriction. As $n \to \infty$, the unnormalized inclusion sums $\sum_{\set \supseteq \set_1} \det(\LL_\set)$, for $\set_1 = \set_0$ and for $\set_1 = \{\yy\}$, and the normalizer $\normalizer$ increase to their values under $\pDPP$ by monotone convergence. Since $0 < \normalizer < \infty$, the inequality passes to the limit.
Negative association, applied to the increasing indicators $\indicator{\yy \in \set}$ on disjoint
coordinates and iterated, gives
$\pDPP(\set_0 \subseteq \set) \le \prod_{\yy \in \set_0} \pDPP(\yy \in \set)$, hence
\begin{align}\label{eq:incl-product}
\incl{\overlap}(\yy_1, {\ldots}, \yy_\overlap)
\le
\frac{\K^{\overlap}}{\falling{\K}{\overlap}} \prod_{j=1}^{\overlap} \incl{1}(\yy_j)
\end{align}

On tuples with a repeat, $\incl{\overlap} = 0$ and \cref{eq:incl-product} is trivial. Both sides are nonnegative, so squaring \cref{eq:incl-product} and taking the expectation over $\overlap$ independent draws from $\proposal$ gives
\begin{align}\label{eq:second-moment}
\E_{\proposal^{\otimes \overlap}}\bigl[\incl{\overlap}^2\bigr]
\le
\Bigl(\frac{\K^{\overlap}}{\falling{\K}{\overlap}}\Bigr)^{2} \rho^{\overlap}
\end{align}
and, because $\incl{\overlap}$ has mean $1$, $\Var_{\proposal^{\otimes \overlap}}(\incl{\overlap}) \le (\K^{\overlap}/\falling{\K}{\overlap})^2 \rho^{\overlap} - 1$;
every term of
\cref{eq:zfree} is then finite, so $\E_\omega[\Zhat^2] < \infty$ and \PoolDPP is
well-behaved.
For the closed form, expanding the binomials gives
$a_\overlap = \binom{\K}{\overlap}\falling{\K}{\overlap}
\falling{N-\K}{\K-\overlap}/\falling{N}{\K}
\le \binom{\K}{\overlap}\falling{\K}{\overlap}/\falling{N}{\overlap}$,
so with  $\theta \defeq \K^2 \rho/(N - \K + 1)$,
\begin{align}
a_\overlap \E_{\proposal^{\otimes \overlap}}\bigl[\incl{\overlap}^2\bigr]
\le
\frac{\binom{\K}{\overlap} \K^{2\overlap} \rho^{\overlap}}{\falling{\K}{\overlap} \falling{N}{\overlap}}
=
\frac{\K^{2\overlap} \rho^{\overlap}}{\overlap! \falling{N}{\overlap}}
\le
\frac{\theta^{\overlap}}{\overlap!}
\end{align}
using  $\binom{\K}{\overlap} = \falling{\K}{\overlap}/\overlap!$ and
$\falling{N}{\overlap} \ge (N - \K + 1)^\overlap$.
 {Summing, and keeping the $-1$ of each $\Var_{\proposal^{\otimes \overlap}}(\incl{\overlap}) = \E_{\proposal^{\otimes \overlap}}[\incl{\overlap}^2] - 1$ with $\sum_{\overlap=0}^{\K} a_\overlap = 1$, gives $1 + \Var_\omega(\Zhat)/\normalizer^2 \le a_0 + \sum_{\overlap=1}^{\K} \theta^\overlap/\overlap! \le e^{\theta} - 1 + a_0$, and \cref{thm:var-kl-mixture} gives $\KL \le \log(e^\theta - 1 + a_0) \le \theta$, since $a_0 \le 1$.}
\end{proof}

\paragraph{\texorpdfstring{ Finiteness of $\rho$.}{Finiteness of m1.}}
Let $\lambda_1, {\ldots}, \lambda_D$ and $\vv_1, {\ldots}, \vv_D$ be the eigenpairs of $\CC = \sum_\yy \quality(\yy) \embed(\yy)\embed(\yy)^\top$, and $\lambda_{-d}$ the eigenvalues with $\lambda_d$ omitted. Writing the primal eigenvectors of $\LL = \BB^\top\BB$ through the dual ones gives the $\K$-DPP marginal formula
\begin{align}\label{eq:marginal-formula}
\pDPP(\yy \in \set) = \quality(\yy) \sum_{d=1}^{D} \langle \vv_d, \embed(\yy)\rangle^2 \frac{e_{\K-1}(\lambda_{-d})}{e_\K(\lambda)}
\end{align}
Because $\|\embed(\yy)\|_2 = 1$ and the $\vv_d$ are an orthonormal basis, the weights $\langle \vv_d, \embed(\yy)\rangle^2$ sum to $1$, so
$\incl{1}(\yy) = \pDPP(\yy \in \set)/(\K \proposal(\yy)) \le w(\yy) \max_d e_{\K-1}(\lambda_{-d})/(\K e_\K(\lambda))$.
 {Hence, $\rho \le \E_\proposal[w^2] \bigl(\max_d e_{\K-1}(\lambda_{-d})/(\K e_\K(\lambda))\bigr)^2$. Here, $e_\K(\lambda) = \normalizer > 0$ (by Cauchy--Binet, $\normalizer = \sum_{|\set| = \K} \det(\LL_\set) = e_\K(\lambda)$), so $\rho$ is finite whenever $\E_\proposal[w^2] < \infty$, in particular whenever the weights are bounded. The bound is invariant to rescaling $\quality$, as $\rho$ is. On the infinite base set $\strings$, \cref{eq:marginal-formula} and $\normalizer = e_\K(\lambda)$ follow from their finite versions by the truncation argument above: $\CC$ is finite because $\sum_\yy \quality(\yy) = \E_\proposal[w] \le \E_\proposal[w^2]^{1/2}$, the truncated dual kernels increase to $\CC$, and both sides converge.}

\paragraph{Negative correlation without strongly Rayleigh measures.}
The proof of \cref{eq:kl-closed-form} uses negative association only for inclusion events, through $\pDPP(\set_0 \subseteq \set) \le \prod_{\yy \in \set_0} \pDPP(\yy \in \set)$, and this inequality has a short self-contained proof that uses only linear algebra and Newton's inequality. We give it because the cited theorem is more than we need: it rests on the theory of real stable polynomials. By the truncation argument, it suffices to take a finite base set $V$. Write $\boldsymbol{\uppsi}(\yy) \defeq \sqrt{\quality(\yy)} \embed(\yy)$, so that $\LL(\xx, \yy) = \boldsymbol{\uppsi}(\xx)^\top \boldsymbol{\uppsi}(\yy)$, and let $\normalizer_\K(V)$ be the sum of $\det(\LL_\set)$ over $\K$-sets $\set \subseteq V$.
(i) \emph{Normalizers.} Let $\CC_V \defeq \sum_{\yy \in V} \boldsymbol{\uppsi}(\yy)\boldsymbol{\uppsi}(\yy)^\top$. Expanding $\det(\mathbf{I} + x \LL_V)$ in principal minors and using $\det(\mathbf{I} + \mathbf{M}\mathbf{M}') = \det(\mathbf{I} + \mathbf{M}'\mathbf{M})$ gives $\sum_\K \normalizer_\K(V) x^\K = \det(\mathbf{I} + x \LL_V) = \det(\mathbf{I} + x \CC_V)$.
(ii) \emph{Two strings.} Fix $\boldsymbol{a} \ne \boldsymbol{b}$ in $V$ and let $V' = V \setminus \{\boldsymbol{a}, \boldsymbol{b}\}$. Rotating every $\boldsymbol{\uppsi}(\yy)$ by the same orthogonal matrix leaves $\LL$ unchanged, so we may take $\CC_{V'} = \mathrm{diag}(\eta)$ with $\eta \ge 0$. Let $\mathbf{s} = \boldsymbol{\uppsi}(\boldsymbol{a})$, $\mathbf{t} = \boldsymbol{\uppsi}(\boldsymbol{b})$, $g_d = e_{\K-1}(\eta_{-d})$, and $\delta_{dd'} = e_{\K-2}(\eta_{-dd'})$, where $\eta_{-d}$ omits $\eta_d$. By (i), the sums of $\det(\LL_\set)$ over $\K$-sets $\set \subseteq V$ containing neither string, only $\boldsymbol{a}$, only $\boldsymbol{b}$, and both are read off from the coefficients of $x^\K$ in $\det(\mathbf{I} + x \CC_W)$ for $W = V'$, $V' \cup \{\boldsymbol{a}\}$, $V' \cup \{\boldsymbol{b}\}$, and $V$, respectively. Here, $\CC_V = \mathrm{diag}(\eta) + \mathbf{s} \mathbf{s}^\top + \mathbf{t} \mathbf{t}^\top$, and the matrix determinant lemma (for $x \ge 0$, $\mathbf{I} + x \mathrm{diag}(\eta)$ is invertible) gives the four sums as
\begin{align}
\normalizer_\emptyset = e_\K(\eta), \quad \normalizer_a = \sum_{d \in [D]} s_d^2 \gamma_d, \quad \normalizer_b = \sum_{d \in [D]} t_d^2 \gamma_d, \quad \normalizer_{ab} = \frac{1}{2}\sum_{d,d' \in [D]} (s_d t_{d'} - s_{d'} t_d)^2 \delta_{dd'}
\end{align}
The inequality $\pDPP(\boldsymbol{a}, \boldsymbol{b} \in \set) \le \pDPP(\boldsymbol{a} \in \set) \pDPP(\boldsymbol{b} \in \set)$ reads $\normalizer_{ab}(\normalizer_\emptyset + \normalizer_a + \normalizer_b + \normalizer_{ab}) \le (\normalizer_a + \normalizer_{ab})(\normalizer_b + \normalizer_{ab})$, that is, $\normalizer_\emptyset \normalizer_{ab} \le \normalizer_a \normalizer_b$. Expanding,
\begin{align}
\normalizer_a \normalizer_b - \normalizer_\emptyset \normalizer_{ab} = \Bigl(\sum_{d \in [D]} s_d t_d \gamma_d\Bigr)^2 + \frac{1}{2}\sum_{d,d' \in [D]} (s_d t_{d'} - s_{d'} t_d)^2 \bigl(\gamma_d \gamma_{d'} - \normalizer_\emptyset \delta_{dd'}\bigr)
\end{align}
For $d \ne d'$ and $\nu = \eta_{-dd'}$, we have $g_d = e_{\K-1}(\nu) + \eta_{d'} e_{\K-2}(\nu)$, likewise for $g_{d'}$, and $\normalizer_\emptyset = e_\K(\nu) + (\eta_d + \eta_{d'}) e_{\K-1}(\nu) + \eta_d \eta_{d'} e_{\K-2}(\nu)$, so $g_d g_{d'} - \normalizer_\emptyset \delta_{dd'} = e_{\K-1}(\nu)^2 - e_\K(\nu) e_{\K-2}(\nu) \ge 0$ by Newton's inequality. Hence, $\normalizer_\emptyset \normalizer_{ab} \le \normalizer_a \normalizer_b$. (For $\K \le 1$, $\pDPP(\boldsymbol{a}, \boldsymbol{b} \in \set) = 0$.)
This step is where the two routes meet. The strongly Rayleigh route gets the two-string inequality from real stability of the generating polynomial $\sum_{\set \subseteq V} \det(\LL_\set) \prod_{\yy \in \set} z_\yy$ in all variables at once. Here, one variable suffices: setting $z_\yy = x$ for every $\yy \in V'$ gives $\det(\mathbf{I} + x \CC_{V'}) = \prod_{d \in [D]} (1 + x \eta_d)$, whose roots are real, and Newton's inequality holds for the coefficients of any polynomial with real roots. The rest of $\normalizer_a \normalizer_b - \normalizer_\emptyset \normalizer_{ab}$ is a sum of squares.
(iii) \emph{Conditioning.} For any $R \subseteq V \setminus \{\boldsymbol{a}\}$, the Schur complement gives $\det(\LL_{R \cup \{\boldsymbol{a}\}}) = \|\boldsymbol{\uppsi}(\boldsymbol{a})\|^2 \det(\LL'_R)$, where $\LL'$ is the kernel of the vectors $\boldsymbol{\uppsi}(\yy)$ projected onto the orthogonal complement of $\boldsymbol{\uppsi}(\boldsymbol{a})$. So given $\boldsymbol{a} \in \set$, the set $\set \setminus \{\boldsymbol{a}\}$ is a $(\K-1)$-DPP of the same form on $V \setminus \{\boldsymbol{a}\}$.
(iv) \emph{Induction on $|\set_0|$.} We induct for every $\K$, finite $V$, and Gram kernel at once, since (iii) changes all three. For $|\set_0| \le 1$, the claim is an equality. For $\boldsymbol{a} \notin \set_0$ with $\pDPP(\boldsymbol{a} \in \set) = 0$, both sides of the step below are $0$; otherwise, $\boldsymbol{\uppsi}(\boldsymbol{a}) \ne 0$, so (iii) applies, and the induction hypothesis for the conditional $(\K-1)$-DPP of (iii), and then (ii), give
$\pDPP(\set_0 \cup \{\boldsymbol{a}\} \subseteq \set) = \pDPP(\boldsymbol{a} \in \set) \pDPP(\set_0 \subseteq \set \mid \boldsymbol{a} \in \set) \le \pDPP(\boldsymbol{a} \in \set) \prod_{\yy \in \set_0} \pDPP(\yy \in \set \mid \boldsymbol{a} \in \set) \le \prod_{\yy \in \set_0 \cup \{\boldsymbol{a}\}} \pDPP(\yy \in \set)$.
{We have not found this argument in the literature.}

\clearpage
\subsection{Proof of \cref{cor:kl-sum}}
\label{app:proof-kl-sum}

Because the target's inclusion events are negatively correlated, $\E_{\proposal^{\otimes \overlap}}[\incl{\overlap}^2] \le (\K^{\overlap}/\falling{\K}{\overlap})^2\, \rho^{\overlap}$ for every $\overlap$ (\cref{eq:second-moment}, \cref{app:zfree}), with equality at $\overlap = 1$. Applying this to every term of \cref{eq:zfree}, with the exact coefficients $a_\overlap$, gives a sharper upper bound.
\begin{restatable}[Sharper bound]{corollary}{corKLSum}
\label{cor:kl-sum}
Under the conditions of \cref{thm:zfree}, if $\rho < \infty$, then
\begin{align}\label{eq:kl-sum}
\KL(\pDPP \| \pibar)
&\le
\log \sum_{\overlap=0}^{\K} a_\overlap \Bigl(\frac{\K^{\overlap}}{\falling{\K}{\overlap}}\Bigr)^{2} \rho^{\overlap}
\notag \\
&=
\log\mleft(1
+ a_1 \Var_\proposal(\incl{1})
+ \sum_{\overlap=2}^{\K} a_\overlap
\mleft(\Bigl(\frac{\K^{\overlap}}{\falling{\K}{\overlap}}\Bigr)^{2} \rho^{\overlap} - 1\mright)\mright)
\end{align}
where $a_0 = \binom{N-\K}{\K}/\binom{N}{\K}$ is the formula of \cref{eq:zfree} at $\overlap = 0$, so the sum is an expectation over the hypergeometric overlap $\overlap$.
The right-hand side is at most $\log(e^{\theta} - 1 + a_0)$, the bound \labelcref{eq:kl-closed-form}, since the proof of \cref{thm:zfree} bounds the term at each $\overlap \ge 1$ by $\theta^{\overlap}/\overlap!$.
\end{restatable}
\begin{proof}
The $\overlap = 1$ ratio is the one-point inclusion marginal against the
proposal, $\incl{1}(\yy) = \pDPP(\yy \in \set)/(\K \proposal(\yy))$, computable, on a finite base set,
from one $D \times D$ eigendecomposition of $\CC$ via \cref{eq:marginal-formula}.
 Bounding every term of \cref{eq:zfree} by \cref{eq:second-moment}, an equality at $\overlap = 1$, and using $\sum_{\overlap=0}^{\K} a_\overlap = 1$ yields  \cref{cor:kl-sum}:
\begin{align}
\KL(\pDPP \| \pibar)
\le
\log\mleft(1
+ a_1 \Var_\proposal(\incl{1})
+ \sum_{\overlap \ge 2} a_\overlap
\mleft(\Bigl(\frac{\K^{\overlap}}{\falling{\K}{\overlap}}\Bigr)^{2} \rho^{\overlap} - 1\mright)\mright)
\end{align}
and, because \cref{eq:zfree} is an equality,  the relative variance is at least $(\K^2/N)\Var_\proposal(\incl{1})$ (\cref{eq:hoeffding-bounds}), which exceeds the $\overlap = 1$ term $a_1 \Var_\proposal(\incl{1})$: at a pool size where $\log(1 + \K^2 \Var_\proposal(\incl{1})/N)$ exceeds a target, the bound $\log(1 + \Var_\omega(\Zhat)/\normalizer^2)$ exceeds it too. This is a statement about the bound, not about the KL, which can be smaller.
Two special cases: at $\K = 1$, $a_1 = 1/N$, \cref{eq:zfree} is the classical
importance-sampling relative variance, and the bound above is an equality, since only the $\overlap = 1$ term remains;
for the proposal matching the target's marginals,
$\proposal(\yy) \propto \pDPP(\yy \in \set)$, the leading term vanishes
identically ($\incl{1} \equiv 1$), so what the leading term rewards is matching
the target's \emph{inclusion marginals}, not its quality.
\end{proof}

\clearpage
\section{Evaluating the Bound on Pool-Built Instances}
\label{app:zfree-eval}%

\paragraph{\texorpdfstring{ The bound on instances built from our pools.}{The bound on instances built from our pools.}}
We evaluate the  bounds on instances built from the pools of
\cref{sec:experiments} (base set the pooled distinct strings, $\proposal$
their empirical frequency, and quality $\plm^{1/T}$ of
\cref{eq:quality-temp}). On instances small enough to enumerate every pool, the two sides of \cref{eq:zfree} agree to 12 significant digits.
 Each instance has the real embeddings, spectrum, and weight dispersion at the experimental $\K$, $D$, and $N$, but its base set is the observed support, the distinct strings in the union of the pools, whose exact proposal mass is $0.68$ on Ambrosia, $0.0017$ on GSM8K, and $0.0003$ on FineWeb (medians over prompts, from the proposal's log-probabilities of the drawn strings), so almost every string of the proposal is missing on GSM8K and FineWeb.  The columns of \cref{tab:zfree-bounds} therefore evaluate the bound on the instance at the experimental scale. They are not a bound on the KL of the sampler on the full proposal, and they can understate it: for $T > 1$ at local sampling temperature $1$, the weights $w = \plm^{1/T - 1}$ are largest on low-probability strings, which the instance mostly lacks, and including them would raise $\rho$. The Monte Carlo estimate draws pools from the  instance and evaluates $\Var_\omega(\Zhat)/\normalizer^2$ directly; at the sharp FineWeb setting ($T{=}0.82$),  its estimate of $\E_\omega[\Zhat]/\normalizer$, which should be $1$, is $0.49$, so that estimate is unreliable.

\begin{table}[h]
\centering\small
\begin{tabular}{lcccccc}
\toprule
& \multicolumn{2}{c}{\makecell{Ambrosia\\$\K{=}3$, $N{=}200$}}
& \multicolumn{2}{c}{\makecell{GSM8K\\$\K{=}10$, $N{=}100$}}
& \multicolumn{2}{c}{\makecell{FineWeb\\$\K{=}10$, $N{=}400$}} \\
\cmidrule(lr){2-3}\cmidrule(lr){4-5}\cmidrule(lr){6-7}
& $T{=}0.67$ & $T{=}1$ & $T{=}1$ & $T{=}1.05$ & $T{=}0.82$ & $T{=}1$ \\
\midrule
$\Var_\proposal(\incl{1})$ & 1.50 & 0.367 & 3.47 & 941 & 95.8 & 0.0650 \\
 &  &  &  &  &  &  \\
$\rho = 1 + \Var_\proposal(\incl{1})$ & 2.50 & 1.37 & 4.47 & 942 & 96.8 & 1.06 \\
leading term $a_1 \Var_\proposal(\incl{1})$ (exact) & 0.0661 & 0.0162 & 1.42 & 384 & 19.5 & 0.0132 \\
lower bound \cref{eq:hoeffding-bounds} & 0.0675 & 0.0165 & 3.47 & 941 & 23.9 & 0.0162 \\
$\widehat V$: $\Var_\omega(\Zhat)/\normalizer^{2}$ (MC) & 0.0672 & 0.0167 & 9.67 & (1500) & (295) & 0.0165 \\
\midrule
$\log(1+\widehat V)$ (nats) & 0.0651 & 0.0165 & 2.37 & (7.31) & (5.69) & 0.0163 \\
 &  &  &  &  &  &  \\
bound \cref{eq:kl-sum} (nats) & 0.0698 & 0.0175 & 4.23 & 53.9 & 17.3 & 0.0233 \\
 &  &  &  &  &  &  \\
 &  &  &  &  &  &  \\
closed form \cref{eq:kl-closed-form} (nats) & 0.0731 & 0.0194 & 4.91 & 1036 & 24.8 & 0.0835 \\
\bottomrule
\end{tabular}
\caption{Bounds on the KL, in nats, evaluated on the instances built from the pools under the kernel of \cref{eq:quality-temp}, at two values of $T$ per task (medians over prompts). The instance of each prompt keeps only the strings the pools drew, with proposal mass $0.68$ on Ambrosia, $0.0017$ on GSM8K, and $0.0003$ on FineWeb (medians over prompts); the values are bounds on the instance, not on the sampler's KL on the full proposal. The  bound \labelcref{eq:kl-sum} is within $1.06$--$1.07\times$ of $\log(1+\widehat V)$ at $\K{=}3$ and  $1.5$--$2.0\times$ at $\K{=}10$, excluding the parenthesized column. $\log(1+\widehat V)$ is a Monte Carlo estimate of the smallest value that the variance argument of \cref{thm:var-kl-mixture} can give. Parenthesized values  are unreliable: there, the Monte Carlo estimate of $\E_\omega[\Zhat]/\normalizer$, which should be $1$, is $0.49$ on FineWeb and below $10^{-20}$ on GSM8K. The $\rho$ row and the lower-bound row follow exactly from the median of $\Var_\proposal(\incl{1})$.}

\label{tab:zfree-bounds}
\end{table}

\begin{table}[h]
\centering\small
\begin{tabular}{lcc}
\toprule
& $N$ for the  lower bound & $N$ sufficient \\
\midrule
Ambrosia $\K{=}3$, $T{=}0.67$ & 129 & 142 \\
Ambrosia $\K{=}3$, $T{=}1$ & 31 & 43 \\
GSM8K $\K{=}10$, $T{=}1$ & 3300 & 3492 \\
GSM8K $\K{=}10$, $T{=}1.05$ & 895145 & 941451 \\
FineWeb $\K{=}10$, $T{=}0.82$ & 91087 & 95816 \\
FineWeb $\K{=}10$, $T{=}1$ & 62 & 149 \\
\bottomrule
\end{tabular}
\caption{Pool size for a bound of $0.1$ nats on the KL (medians over prompts): the smallest pool at which  the lower bound of \cref{eq:hoeffding-bounds} allows it, and a pool at which \labelcref{eq:kl-sum}  reaches it. At $T{=}1$, the bound reaches $0.1$ nats with pools of $43$ strings on Ambrosia, $149$ on FineWeb, and $3{,}492$ on GSM8K; at the other settings, it needs $142$ strings on Ambrosia ($T{=}0.67$), $95{,}816$ on FineWeb ($T{=}0.82$), and $941{,}451$ on GSM8K ($T{=}1.05$).}

\label{tab:zfree-required-N}
\end{table}

\paragraph{\texorpdfstring{ Running GSM8K with larger pools.}{Running GSM8K with larger pools.}}
The pool size trades accuracy against cost, so we ran GSM8K  with pools of $N{=}2{,}500$. When we extrapolate the  instance built from the $N{=}100$ pools through the $N$-independent $\Var_\proposal(\incl{1})$, the  bound falls below $0.1$ nats at  $N \approx 3{,}500$ for $T{=}1$, while $T{=}1.05$ needs $N \approx 941{,}000$; at $N{=}2{,}500$, it is $0.140$ and $21.5$ nats:

\begin{table}[h]
\centering\small
\setlength{\tabcolsep}{3pt}%
\begin{tabular}{lcccc}
\toprule
 bound \labelcref{eq:kl-sum} in nats (GSM8K, $\K{=}10$) & $N{=}100$ & $500$ & $1{,}000$ & $2{,}500$ \\
\midrule
$T{=}1$ & 4.23 & 0.725 & 0.355 & 0.140 \\
$T{=}1.05$ & 53.9 & 37.4 & 30.5 & 21.5 \\
\bottomrule
\end{tabular}
\caption{ The bound \labelcref{eq:kl-sum} on the pool-built instance, in nats, as a function of the pool size (medians over
prompts; \cref{fig:bound-vs-N}). All columns extrapolate the  instance built from the $N{=}100$ pools through the $N$-independent $\Var_\proposal(\incl{1})$.}

\label{tab:bound-vs-N}
\end{table}

\begin{figure}[h]
\centering
\includegraphics[width=0.55\textwidth]{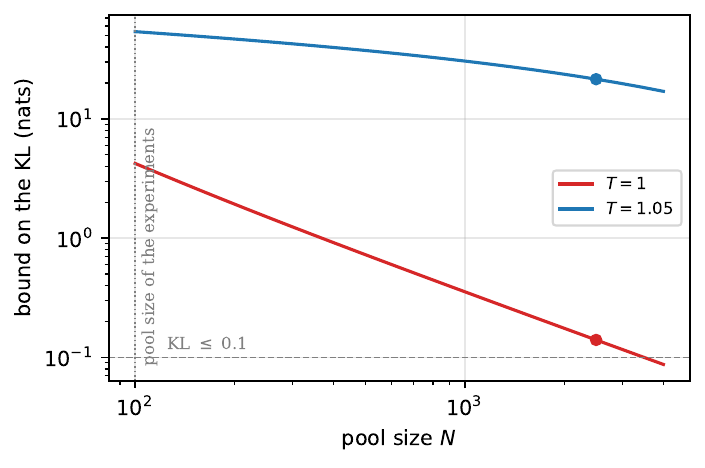}
\caption{ \textbf{The bound \labelcref{eq:kl-sum} on the pool-built instance vs.\ pool size $N$} on GSM8K
(medians over prompts; dots mark the  bound at $N{=}2{,}500$, the pool size of the separate run below; like the curves, they are extrapolated from the $N{=}100$ pools).}

\label{fig:bound-vs-N}
\end{figure}

\clearpage
\section{Additional Algorithms}

For completeness, we include pseudocode for dual sampling.
\paragraph{\SampleDual.}\label{app:dual-sampler}
\SampleDual takes $\K$ and $\BB$ and returns an exact sample from the $\K$-DPP; see \citet{kulesza-dpp} for proofs of correctness. Phase 1 selects a size-$\K$ subset $\mathcal{J}$ of eigenvector indices of $\CC$ with the elementary symmetric polynomials of its eigenvalues.\footnote{\label{esp}The elementary symmetric polynomials $e_k(\lambda_1, {\ldots}, \lambda_n) \defeq \sum_{1 \le i_1 < \cdots < i_k \le n} \lambda_{i_1} \cdots \lambda_{i_k}$ can be computed in $\bigo(D \K)$ time by a dynamic program \citep[Alg.~7]{kulesza-dpp}.} Phase 2 samples $\K$ items: at each step an item is chosen with probability proportional to how much its embedding lies in the subspace spanned by the selected eigenvectors, and the subspace is then projected away from the chosen item. This projection is what makes similar items compete. Forming $\CC$ costs $\bigo(D^2 N)$, the eigendecomposition $\bigo(D^3)$, Phase 1 is negligible, and Phase 2 costs $\bigo(D N \K)$. The primal eigendecomposition is faster for small pools (\cref{app:cost}).

\begin{algorithm}[h]
\footnotesize
\caption{$\SampleDual$: Draws exact samples from the $\K$-DPP using the dual representation trick; pseudocode adapted from \citet[Alg.~3]{kulesza-dpp}.
The base set $\universe$ is the set of \emph{column indices} of the input feature matrix $\BB$. In particular, when \algname invokes it with $\Bapprox$ (\cref{eq:B-matrix}), the base set is the pool indices $\{1, {\ldots}, N\}$, and the returned subset is an index set that \algname maps to its labeled strings (\cref{prop:pushforward}).
}
  \label{alg:dual-kdpp}
  \begin{algorithmic}[1]
    \STATE \textbf{Input:} Subset size $\K$, feature matrix $\BB \in \R^{D \times \universe}$ with $\mathrm{rank}(\BB) \ge \K$.
    \STATE \textbf{Output:} Subset $\set \in \binom{\universe}{\K}$.
    \STATE $\CC \gets \BB \BB^\top$
    \STATE Eigendecompose $\CC$ with eigenvalues $\lambda_1, {\ldots}, \lambda_D$ and eigenvectors $\vv_1, {\ldots}, \vv_D$
    \STATE \textit{// Phase 1: Select $\K$ eigenvectors using elementary symmetric polynomials ($e_k(\cdot)$; \cref{esp}).}
    \STATE $\mathcal{J} \leftarrow \emptyset$
    \FOR{$d = D, {\ldots}, 1$}
      \STATE Let $\lambda_{j_1}, {\ldots}, \lambda_{j_r}$ be the eigenvalues s.t.\ $j < d, j \notin \mathcal{J}$
      \STATE $\xi \leftarrow \lambda_d \cdot \frac{e_{\K - |\mathcal{J}| - 1}(\lambda_{j_1}, {\ldots}, \lambda_{j_r}) }{ e_{\K - |\mathcal{J}|}(\lambda_d, \lambda_{j_1}, {\ldots}, \lambda_{j_r})}$
      \STATE Add $d$ to $\mathcal{J}$ with probability $\xi$
      \STATE \textbf{if} $|\mathcal{J}| = \K$: \textbf{break}
    \ENDFOR
    \STATE \textit{// Phase 2: Sample $\K$ items from the selected eigenspace.}
    \STATE $\set \leftarrow \emptyset$, \quad $\mathcal{V} \gets \{ \frac{1}{\sqrt{\vv_j^\top \CC \vv_j}} \vv_j \mid j \in \mathcal{J} \}$
    \WHILE{$|\set| < \K$}
      \STATE Sample $\yy$ with $\mathrm{Pr}(\yy) = \frac{1}{|\mathcal{V}|} \sum_{\vv \in \mathcal{V}} (\vv^\top \BB_{:,\yy})^2$      
      \STATE $\set \leftarrow \set \cup \{\yy\}$; $\bb \gets \BB_{:, \yy}$
      \STATE Pick any vector $\vv_0 \in \mathcal{V}$ with $\vv_0^\top \bb \ne 0$
      \STATE $\mathcal{V} \gets \{ \vv - \frac{\vv^\top \bb}{\vv_0^\top \bb} \vv_0 \mid \vv \in \mathcal{V} - \{ \vv_0 \} \}$
      \STATE Orthonormalize $\mathcal{V}$ with respect to the inner product $\langle \uu, \vv \rangle \defeq \uu^\top \CC \vv$
    \ENDWHILE
    \STATE \textbf{return} $\set$
  \end{algorithmic}
\end{algorithm}

\begin{algorithm}[h]
\footnotesize
\caption{$\GreedyMAP$: greedy approximation to the $\K$-DPP MAP, with the multiplicative volume guarantee of \citet[Thm.~11]{civril2009maxvol}. Schur-complement greedy expressed in the feature-matrix basis (cf.\ \citealp{chen-2018-fast-greedy-dpp-map} for the kernel-basis Cholesky form). Maintains, for each pool item, the squared distance of its embedding to the span of the currently chosen items, and at each step picks the item with the largest such distance. Runs in $\bigo(D N \K)$ time. As in \cref{alg:dual-kdpp}, the base set $\universe$ is the set of column indices of $\BB$; when the algorithm is invoked with $\Bapprox$, the returned subset is a set of pool indices, which \algname maps to its labeled strings.}
\label{alg:greedy-map}
\begin{algorithmic}[1]
  \STATE \textbf{Input:} Subset size $\K$, feature matrix $\BB \in \R^{D \times \universe}$ with $\mathrm{rank}(\BB) \ge \K$.
  \STATE \textbf{Output:} Subset $\set \in \binom{\universe}{\K}$.
  \STATE $\set \gets \emptyset$
  \FOR{$\yy \in \universe$}
    \STATE $d_\yy \gets \|\BB_{:,\yy}\|^2$ \hfill\textit{// squared residual norm $= \LL_{\yy\yy}$}
  \ENDFOR
  \WHILE{$|\set| < \K$}
    \STATE $\yy^\star \gets \argmax_{\yy \notin \set} d_\yy$ \hfill\textit{// largest remaining contribution to $\log\det$}
    \STATE $\set \gets \set \cup \{\yy^\star\}$
    \STATE $\bb \gets \BB_{:,\yy^\star} / \sqrt{d_{\yy^\star}}$ \hfill\textit{// unit residual direction}
    \FOR{$\yy \notin \set$}
      \STATE $\alpha \gets \langle \bb, \BB_{:,\yy}\rangle$
      \STATE $\BB_{:,\yy} \gets \BB_{:,\yy} - \alpha \bb$ \hfill\textit{// project out chosen direction}
      \STATE $d_\yy \gets d_\yy - \alpha^2$ \hfill\textit{// shrink residual norm}
    \ENDFOR
  \ENDWHILE
  \STATE \textbf{return} $\set$
\end{algorithmic}
\end{algorithm}

\clearpage
\section{Experimental Ablations}
\label{sec:ablations}

Unless otherwise stated, all ablations use the FineWeb intrinsic setup from \cref{sec:intrinsic}; results use one pool and, for stochastic methods, $10$ selection runs.

\begin{figure}[ht]
\centering
\includegraphics[width=.86\columnwidth]{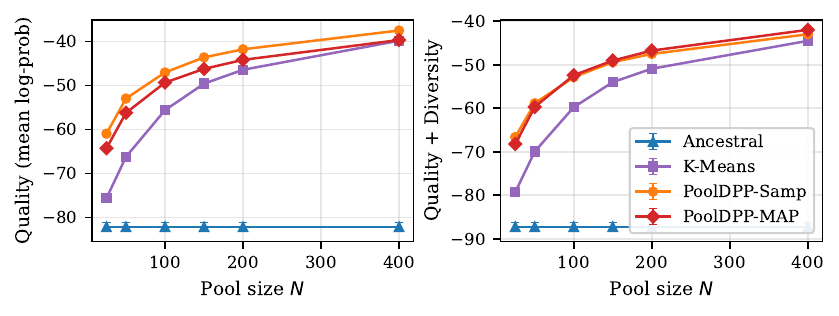}
\caption{\textbf{Pool size $N$ ablation} (FineWeb, $\K=10$, local sampling temperature $1$, \algnameSamp at $T{=}0.82$, \algnameMAP at $T{=}50$). Mean log-probability and the full objective, its sum with the log-determinant, rise with $N$ at a decreasing rate.}
\label{fig:abl_pool_size}
\end{figure}

\paragraph{Pool size $N$.}
We sweep $N \in \{25,50,100,150,200,400\}$ at $\K=10$ and local sampling temperature $1$, with \algnameSamp at global temperature $T{=}0.82$ and \algnameMAP at $T{=}50$ (\cref{fig:abl_pool_size}). \Cref{thm:zfree} predicts that \algnameSamp's approximation error decreases as $\bigo(1/N)$. Since the exact KL to the population $\K$-DPP is intractable, we instead report the sum of the selected set's mean log-probability and log-determinant, together with mean log-probability. This sum rises with $N$ at a decreasing rate and has not saturated at $N{=}400$

\begin{figure}[ht]
\centering
\includegraphics[width=.86\columnwidth]{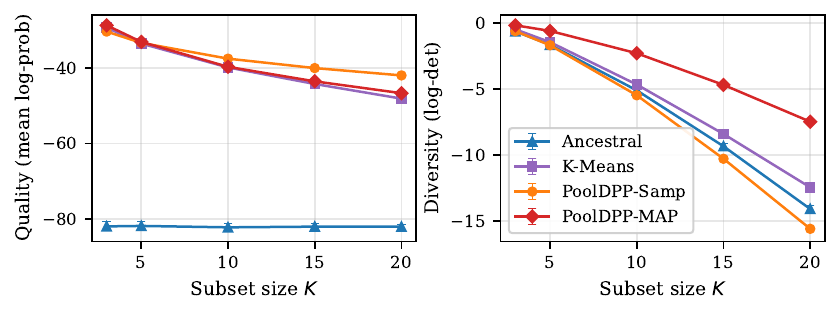}
\caption{\textbf{Subset size $\K$ ablation} (FineWeb, $N=400$, local sampling temperature $1$, \algnameSamp at $T{=}0.82$, \algnameMAP at $T{=}50$). Quality falls with $\K$ for the pool-based methods, and the log-determinant falls for every method.}
\label{fig:abl_k_size}
\end{figure}

\paragraph{Subset size $\K$.}
We sweep $\K \in \{3,5,10,15,20\}$ at $N=400$ and local sampling temperature $1$, with \algnameSamp at $T{=}0.82$ and \algnameMAP at $T{=}50$ (\cref{fig:abl_k_size}). At every $\K$, \algnameMAP has the highest full-score log-determinant, with quality at or above \algnameKM.

\begin{figure}[t]
\centering
\includegraphics[width=.90\columnwidth]{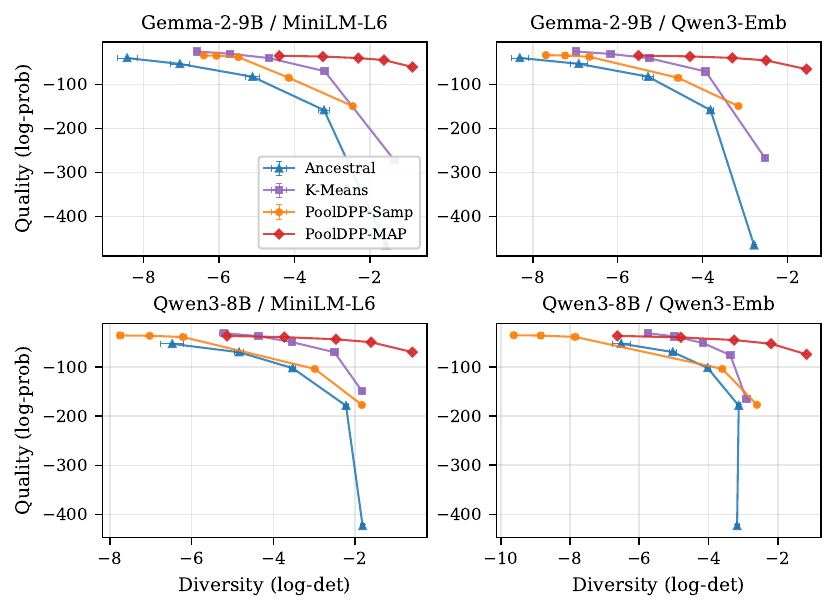}
\caption{\textbf{LM/embedding robustness} (FineWeb, $\K=10$, $N=400$). Pareto frontiers across two LMs and two embedders. The qualitative ordering is stable across all four configurations.}
\label{fig:abl_lm_emb}
\end{figure}

\paragraph{Embedder and language model.}
We replicate the FineWeb experiment with all four combinations of
\begin{align*}
\{\text{Gemma-2-9B-Instruct}, \text{Qwen3-8B}\}
\times
\{\text{MiniLM-L6}, \text{Qwen3-Embedding-0.6B}\}
\end{align*}
\Cref{fig:abl_lm_emb} shows the same ordering in all four configurations. \algnameMAP dominates \algnameKM at local sampling temperatures $1$, $1.2$, and $1.5$ and every ancestral point, and it reaches a higher log-determinant, while \algnameKM at keeps higher quality at a lower log-determinant. \algnameKM dominates most \algnameSamp settings. Thus, the DPP construction is not tied to a single LM or embedding geometry.

\begin{figure}[t]
\centering
\includegraphics[width=\textwidth]{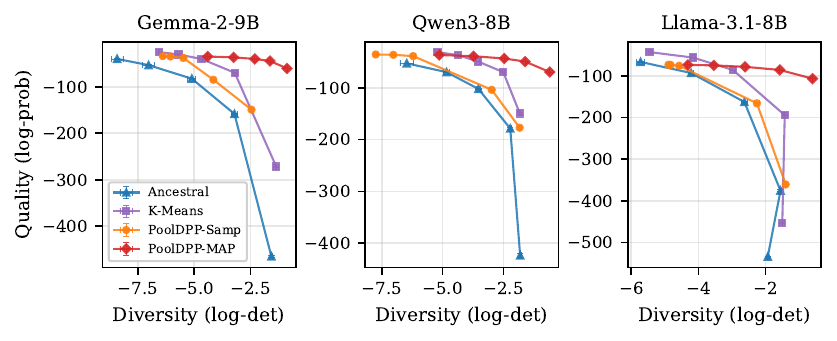}
\caption{\textbf{Three-model robustness} (FineWeb, MiniLM-L6, $\K=10$). Quality vs.\ diversity for Gemma-2-9B-Instruct (as in the main experiments), Qwen3-8B, and Llama-3.1-8B-Instruct; curves follow the same sweep protocol as \cref{fig:fineweb_pareto} (ancestral and \algnameKM sweep local sampling temperature; \algnameSamp and \algnameMAP draw the pool at local sampling temperature $1$ and sweep the global temperature $T$). \algnameMAP (red) dominates \algnameKM (purple) at higher local sampling temperatures.}
\label{fig:abl_multimodel}

\end{figure}

\paragraph{Generalization across language models.}
Beyond the 2$\times$2 LM/embedding ablation, we additionally validate the main FineWeb intrinsic experiment on \textbf{Llama-3.1-8B-Instruct}. \Cref{fig:abl_multimodel} reports the quality--diversity Pareto frontier on all three language models with a fixed embedding (MiniLM-L6, $\K=10$), following the same sweep protocol as \cref{fig:fineweb_pareto}: ancestral and \algnameKM sweep the local sampling temperature, while \algnameSamp and \algnameMAP draw the pool at local sampling temperature $1$ and sweep the global temperature $T$. On every language model, \algnameMAP dominates \algnameKM at higher local sampling temperatures, while \algnameKM keeps higher quality at a lower log-determinant. This indicates that the construction is not tied to a particular model family or post-training recipe.

\paragraph{Global temperature $T$.}
The Pareto curves in \cref{fig:fineweb_pareto} sweep $T$. Smaller $T$ sharpens the quality score toward high-likelihood strings, while larger $T$ flattens it and gives the geometric diversity term more influence. The best $T$ differs across tasks and between the two modes: \algnameMAP explicitly optimizes the determinant objective and typically does best with a flatter kernel than \algnameSamp, to avoid collapsing toward likelihood-dominated selections.

\paragraph{Weight diagnostics.}
The ESS of \algnameSamp's weights $w = \plm^{1/T}/\proposal$ measures how well the pool approximates the globally annealed target. Away from $T{=}1$ at local sampling temperature $1$, where $w \equiv 1$, the per-prompt ESS is below $\K$ for most prompts on FineWeb ($36$ to $49$ of $50$ at every plotted $T \ne 1$) and on GSM8K at local sampling temperature $1.2$ ($36$ to $50$ of $50$ at every plotted $T$ except $T{=}1$), and a $20$- to $25$-fold larger pool does not fix this at $T > 1$ (\cref{app:weights}). On Ambrosia ($\K{=}3$), no prompt is below $\K$. Here the pool is a poor approximation of the target, yet the quality and diversity metrics still vary smoothly with $T$, so $T$ remains a usable knob for the quality--diversity frontier. A high ESS can be achieved by changing the target: the ESS is high whenever the target stays close to the proposal, for example when the proposal is added into the target.

\paragraph{LLM-based selection.}
We compare against \algnameLLM, the pool-reranking baseline that replaces the DPP with a language model (\cref{sec:experiments}): we present $40$ pool items, drawn at local sampling temperature $1$and evenly spaced in the pool's ranking by log-probability, to Qwen3-8B and ask it to select $\K{=}10$ diverse, high-quality completions. Over $10$ pools, \algnameKM on the same $40$ items dominates the LLM on both axes: quality $-69.4$ and log-determinant $-3.58$, against $-71.5$ and $-6.16$. We omit \algnameLLM from \cref{fig:fineweb_pareto} for legibility.

\paragraph{Efficiency.}
Pool construction dominates runtime; selection is a small fraction of total wall-clock time. \Cref{tab:efficiency-fixed} reports per-stage wall-clock time at the main configuration ($N=400$, $\K=10$, Gemma-2-9B-Instruct on a single A100), and \cref{tab:efficiency-scaling} reports selection time as $N$ varies. \algnameMAP is the cheapest selection method in our implementation: the greedy update uses $\K$ sequential updates against the pool feature matrix or dual kernel, and its measured cost is substantially below that of $\K$-means across the tested pool sizes.

\begin{table}[h]
\centering
\small
\begin{minipage}[t]{0.46\linewidth}
\centering
\caption{\textbf{Wall-clock per stage} ($N=400$, $\K=10$, A100). Selection is small relative to LM sampling.}
\label{tab:efficiency-fixed}
\begin{tabular}{lr}
\toprule
Stage & Time \\
\midrule
LM sampling (vLLM, async)        & $\sim$20 s \\
Embedding (MiniLM-L6)            & $\sim$200 ms \\
\algnameKM selection             & 273 ms \\
\algnameSamp selection           & 139 ms \\
\algnameMAP selection            & 13 ms \\
\bottomrule
\end{tabular}
\end{minipage}\hfill
\begin{minipage}[t]{0.5\linewidth}
\centering
\setlength{\tabcolsep}{3pt}%
\caption{\textbf{Selection-time scaling with pool size $N$} ($\K=10$). \algnameMAP has the smallest measured selection cost across the tested pool sizes.}
\label{tab:efficiency-scaling}
\begin{tabular}{lrrr}
\toprule
$N$ & \algnameMAP & \algnameKM & \algnameSamp \\
\midrule
25  & 1 ms   & 152 ms & 8 ms \\
100 & 4 ms   & 120 ms & 138 ms \\
400 & 17 ms  & 336 ms & 154 ms \\
\bottomrule
\end{tabular}
\end{minipage}
\end{table}

\subsection{Weight Diagnostics}
\label{app:weights}

\Cref{tab:weights} reports, per task and plotted $T$, the effective sample size of the weights of \algnameSamp, $w = \plm^{1/T}/\proposal$, and, for \algnameMAP on Ambrosia, of the quality $\quality = \plm^{1/T}$ that takes the place of $w$. With self-normalized weights $\hat w_n = w(\yy^{(n)}) / \sum_{n'=1}^{N} w(\yy^{(n')})$, where a string drawn several times contributes one term per draw, $\mathrm{ESS} = 1/\sum_{n=1}^{N} \hat w_n^2$. At local sampling temperature $1$, $w \equiv 1$ at $T{=}1$, so $\mathrm{ESS} = N$. At the other $T$, ESS is below $\K$ for most prompts on FineWeb and GSM8K at the run pool size. The larger pool raises it at $T \le 1$ but not above: at $T{=}1.22$ on FineWeb and at $T{=}1.05$ and $1.22$ on GSM8K, all $50$ prompts stay below $\K$. On Ambrosia, only \algnameMAP at $T{=}1$ and $T{=}3$ has prompts below $\K$ at the run pool size ($18$ and $1$ of $100$).

\begin{table}[t]
\centering
\footnotesize
\setlength{\tabcolsep}{3pt}
\caption{\textbf{Weight diagnostics per task and temperature.} ESS$/N$ is a mean over prompts; ESS med.\ [IQR] is the median and interquartile range over prompts, and $\#{<}\K$ counts prompts with ESS below $\K$, at the run pool size $N$ and at a larger pool $N'$. Gemma-2-9B-Instruct throughout; $50$ prompts for FineWeb and GSM8K, $100$ for Ambrosia. Ambrosia at local sampling temperature $1.5$ is \algnameMAP. $N$: one pool of each task. $N'$: two independent pools of $1{,}250$ (GSM8K) or $4{,}000$ (FineWeb) draws per prompt, merged, and on Ambrosia the union of the $100$ pools of $N{=}200$.}

\label{tab:weights}
\begin{tabular*}{\textwidth}{@{\extracolsep{\fill}}lrrrrr}
\toprule
& & \multicolumn{2}{c}{run pool size $N$} & \multicolumn{2}{c}{larger pool $N^{\prime}$} \\
\cmidrule(lr){3-4}\cmidrule(l){5-6}
$T$ & ESS$/N$ & ESS med.\ [IQR] & $\#{<}\K$ & ESS med.\ [IQR] & $\#{<}\K$ \\
\midrule
\multicolumn{6}{@{}l}{\textbf{GSM8K}, $\K{=}10$, $N{=}100$, $N'{=}2{,}500$, local sampling temperature $1.2$, \algnameSamp} \\
\midrule
$T{=}0.43$ & 0.018 & 1.4 [1.0, 2.3] & 50 & 5.8 [2.9, 9.5] & 38 \\
$T{=}0.82$ & 0.078 & 6.7 [3.4, 9.8] & 38 & 99.4 [63.6, 203] & 0 \\
$T{=}1$ & 0.304 & 30.0 [21.5, 39.1] & 2 & 673 [494, 906] & 1 \\
$T{=}1.05$ & 0.113 & 1.4 [1.0, 11.9] & 36 & 1.0 [1.0, 1.0] & 50 \\
$T{=}1.22$ & 0.010 & 1.0 [1.0, 1.0] & 50 & 1.0 [1.0, 1.0] & 50 \\
\midrule
\multicolumn{6}{@{}l}{\textbf{FineWeb}, $\K{=}10$, $N{=}400$, $N'{=}8{,}000$, local sampling temperature $1$, \algnameSamp} \\
\midrule
$T{=}0.43$ & 0.007 & 1.1 [1.0, 1.5] & 48 & 4.8 [1.6, 11.7] & 34 \\
$T{=}0.67$ & 0.014 & 1.9 [1.0, 2.9] & 47 & 9.4 [3.7, 21.1] & 26 \\
$T{=}0.82$ & 0.034 & 3.8 [1.5, 11.2] & 36 & 32.2 [9.4, 87.4] & 13 \\
$T{=}1$ & 1.000 & 400 [400, 400] & 0 & 8000 [8000, 8000] & 0 \\
$T{=}1.22$ & 0.006 & 1.8 [1.1, 2.9] & 49 & 1.7 [1.2, 3.5] & 50 \\
\midrule
\multicolumn{6}{@{}l}{\textbf{Ambrosia}, $\K{=}3$, $N{=}200$, $N'{=}20{,}000$, local sampling temperature $1$, \algnameSamp} \\
\midrule
$T{=}0.43$ & 0.191 & 31.6 [20.0, 47.7] & 0 & 3079 [2078, 4686] & 0 \\
$T{=}0.54$ & 0.267 & 46.1 [35.0, 64.4] & 0 & 4505 [3501, 6346] & 0 \\
$T{=}0.67$ & 0.409 & 78.7 [62.7, 97.8] & 0 & 7728 [6470, 9461] & 0 \\
$T{=}0.82$ & 0.676 & 136 [122, 149] & 0 & 13454 [12377, 14777] & 0 \\
$T{=}1$ & 1.000 & 200 [200, 200] & 0 & 20000 [20000, 20000] & 0 \\
\midrule
\multicolumn{6}{@{}l}{\textbf{Ambrosia}, $\K{=}3$, $N{=}200$, $N'{=}20{,}000$, local sampling temperature $1.5$, \algnameMAP} \\
\midrule
$T{=}1$ & 0.048 & 5.4 [3.6, 11.4] & 18 & 481 [268, 1091] & 0 \\
$T{=}3$ & 0.131 & 20.9 [13.7, 33.1] & 1 & 1991 [1229, 3395] & 0 \\
$T{=}7.5$ & 0.277 & 51.4 [36.0, 70.9] & 0 & 5070 [3547, 6931] & 0 \\
$T{=}10$ & 0.327 & 63.9 [44.6, 83.2] & 0 & 6204 [4473, 8149] & 0 \\
\bottomrule
\end{tabular*}
\end{table}

\subsection{Intrinsic Evaluation with Held-Out and Embedder-Free Diversity Metrics}
\label{app:intrinsic-heldout}

Both axes of \cref{fig:fineweb_pareto} are terms of the objective that \algnameMAP maximizes, so its position there is partly a consequence of the objective. We re-score the FineWeb selections on one of the $10$ FineWeb pools with diversity metrics outside the kernel: the log-det and the mean pairwise cosine distance under a held-out embedder (Qwen3-Embedding-0.6B, whereas the kernel uses MiniLM-L6), and two lexical metrics that use no embedder, Distinct-$n$ (distinct whitespace-token $n$-grams divided by all $n$-grams in the set) and the mean pairwise Jaccard similarity of the strings' token sets (lower is more diverse). Quality is the mean $\log\plm$ of \cref{sec:intrinsic}. Across all $25$ settings, the Spearman rank correlation with the kernel log-det is $0.979$ for the held-out log-det, $0.988$ for the held-out cosine distance, $0.946$, $0.897$, and $0.878$ for Distinct-$1$, -$2$, and -$3$, and $-0.975$ for the Jaccard similarity (\cref{tab:intrinsic-heldout}).

A setting is on the Pareto frontier if no other setting is at least as good on quality and diversity and better on one of them. Under the kernel log-det and both held-out embedding metrics, the frontier is the five \algnameMAP settings and \algnameKM at local temperatures $0.6$ and $0.8$; under the four lexical metrics it also contains settings of the other methods (\cref{tab:intrinsic-heldout}). No setting of another method dominates any \algnameMAP setting under any of the seven metrics, so the high-diversity end of \cref{fig:fineweb_pareto} does not depend on the MiniLM-L6 geometry; the quality axis remains the LM log-probability that also enters the objective.

\begin{table}[h]
\centering
\footnotesize
\setlength{\tabcolsep}{3.5pt}
\caption{\textbf{FineWeb selections under non-kernel diversity metrics} (Gemma-2-9B-Instruct, MiniLM-L6 kernel, $\K{=}10$, $50$ prompts, one of the $10$ pools of \cref{fig:fineweb_pareto}, $N{=}400$). Quality: mean $\log\plm$ per string. \emph{Kernel}: log-det of the MiniLM-L6 Gram matrix, the axis of \cref{fig:fineweb_pareto}. \emph{Held-out}: log-det and mean pairwise cosine distance under Qwen3-Embedding-0.6B. D-1, D-2: Distinct-$1$, Distinct-$2$. Jaccard: mean pairwise Jaccard similarity of token sets (lower is more diverse). Means over selected sets: $10$ runs per prompt ($500$ sets) for every method except \algnameMAP (deterministic, $50$ sets). Rows are a subset of the $25$ settings (ancestral sampling, \algnameKM, and naive DPP at local sampling temperatures $\{0.6, 0.8, 1.0, 1.2, 1.5\}$; \algnameSamp and \algnameMAP at local temperature $1$ and the five values of $T$ of \cref{fig:fineweb_pareto}); $r_{\mathrm{s}}$ is computed over all $25$. Frontier: number of settings on the quality--diversity Pareto frontier, with the number of \algnameMAP settings among them in parentheses. No uncertainty estimates.}

\label{tab:intrinsic-heldout}
\begin{tabular}{llrrrrrrr}
\toprule
& & & & \multicolumn{2}{c}{Held-out} & & & \\
\cmidrule(lr){5-6}
Method & Setting & Quality & Kernel & log-det & cos.\ dist. & D-1 & D-2 & Jaccard \\
\midrule
Ancestral    & local $0.6$ & $-39.6$ & $-8.44$ & $-8.30$ & $0.445$ & $0.521$ & $0.831$ & $0.131$ \\
Ancestral    & local $1.0$ & $-82.2$ & $-5.11$ & $-5.28$ & $0.538$ & $0.629$ & $0.933$ & $0.087$ \\
\algnameKM   & local $0.6$ & $-24.8$ & $-6.58$ & $-7.42$ & $0.487$ & $0.510$ & $0.792$ & $0.128$ \\
\algnameKM   & local $0.8$ & $-30.0$ & $-5.70$ & $-6.40$ & $0.518$ & $0.558$ & $0.851$ & $0.105$ \\
\algnameKM   & local $1.0$ & $-39.8$ & $-4.67$ & $-5.53$ & $0.551$ & $0.609$ & $0.895$ & $0.083$ \\
\algnameSamp & $T{=}0.82$ & $-37.5$ & $-5.49$ & $-6.52$ & $0.514$ & $0.604$ & $0.878$ & $0.089$ \\
\algnameSamp & $T{=}1$ & $-84.5$ & $-4.15$ & $-4.70$ & $0.569$ & $0.638$ & $0.940$ & $0.082$ \\
\algnameMAP  & $T{=}500$ & $-60.1$ & $-0.87$ & $-2.44$ & $0.711$ & $0.704$ & $0.961$ & $0.048$ \\
\algnameMAP  & $T{=}100$ & $-44.2$ & $-1.63$ & $-3.24$ & $0.662$ & $0.682$ & $0.943$ & $0.054$ \\
\algnameMAP  & $T{=}50$ & $-39.7$ & $-2.30$ & $-3.83$ & $0.632$ & $0.670$ & $0.930$ & $0.059$ \\
\algnameMAP  & $T{=}25$ & $-36.3$ & $-3.24$ & $-4.71$ & $0.589$ & $0.645$ & $0.910$ & $0.068$ \\
\algnameMAP  & $T{=}10$ & $-34.4$ & $-4.40$ & $-5.76$ & $0.547$ & $0.619$ & $0.885$ & $0.080$ \\
\midrule
\multicolumn{3}{l}{$r_{\mathrm{s}}$ with kernel log-det ($25$ settings)} & $1$ & $0.979$ & $0.988$ & $0.946$ & $0.897$ & $-0.975$ \\
\multicolumn{3}{l}{Frontier size (\algnameMAP settings)} & $7$ ($5$) & $7$ ($5$) & $7$ ($5$) & $14$ ($5$) & $13$ ($5$) & $12$ ($5$) \\
\bottomrule
\end{tabular}
\end{table}

\subsection{Dual versus Primal Cost}
\label{app:cost}

\Cref{tab:dual-primal} measures where the dual kernel $\CC$ of \cref{sec:dpp} becomes faster than the primal kernel in our implementation. We time exact $\K$-DPP sampling with $\K{=}10$ on synthetic pools of $N$ random unit-norm embeddings in $\R^D$, on one CPU thread. The dual time covers forming $\CC$, its eigendecomposition, and one draw of \cref{alg:dual-kdpp}; the primal time covers forming the $N \times N$ kernel and its eigendecomposition only, so it is a lower bound on the cost of a primal sampler.

In every reported experiment, \algnameSamp uses the dual sampler of \cref{alg:dual-kdpp}, and a pool with $\Zhat = 0$ is redrawn (\cref{fn:wlog}), so every prompt contributes. \algnameMAP runs the greedy algorithm (\cref{alg:greedy-map}) on the primal kernel over the unique pool strings, whose number is at most $N$. Its measured cost, which includes forming that kernel, is $13.7$~ms at $D{=}384$, $N{=}400$ and grows to $1.4$~s ($D{=}384$) and $2.5$~s ($D{=}1024$) at $N{=}10^4$.

The dual therefore pays off for pools of a few thousand strings or more, such as the $N{=}2{,}500$ pools of the pool-size experiment (\cref{app:zfree-eval}). At the pool sizes of the main experiments, either representation costs at most $0.2$~s per prompt, which is small next to LM sampling (\cref{tab:efficiency-fixed}).

\begin{table}[h]
\centering
\small
\caption{\textbf{Wall-clock time for exact $\K$-DPP sampling} ($\K{=}10$, milliseconds, one CPU thread). Synthetic pools: $N$ embeddings, each a Gaussian vector in $\R^D$ with a shared mean direction, normalized to unit norm, and quality weights from synthetic log-probabilities. \emph{Dual}: form $\CC$, eigendecompose it, and draw one sample with \cref{alg:dual-kdpp}. \emph{Primal eig.}: form the $N \times N$ kernel and eigendecompose it (the sampling phase is not timed). \emph{Primal MAP}: form the $N \times N$ kernel and run the greedy algorithm (\cref{alg:greedy-map}) as implemented in our code. Median of $5$ repetitions for $N \le 800$ and a single repetition for $N \ge 2{,}000$; no uncertainty estimates. Last column: primal eig.\ divided by dual; values above $1$ favor the dual.}
\label{tab:dual-primal}
\begin{tabular}{rrrrrr}
\toprule
$D$ & $N$ & Dual & Primal eig. & Primal MAP & Primal eig.\ / dual \\
\midrule
$384$  & $100$    & $20.1$  & $0.80$        & $3.2$       & $0.040$ \\
$384$  & $200$    & $21.4$  & $3.07$        & $6.6$       & $0.14$ \\
$384$  & $400$    & $26.5$  & $15.5$        & $13.7$      & $0.59$ \\
$384$  & $800$    & $28.2$  & $67.9$        & $29.8$      & $2.4$ \\
$384$  & $2{,}000$  & $35.2$  & $790$         & $108$       & $22$ \\
$384$  & $10{,}000$ & $145$   & $115{,}044$   & $1{,}411$   & $791$ \\
\midrule
$1024$ & $100$    & $184$   & $1.01$        & $3.4$       & $0.0055$ \\
$1024$ & $200$    & $191$   & $3.73$        & $7.2$       & $0.020$ \\
$1024$ & $400$    & $200$   & $16.6$        & $15.8$      & $0.083$ \\
$1024$ & $800$    & $233$   & $86.5$        & $37.7$      & $0.37$ \\
$1024$ & $2{,}000$  & $294$   & $873$         & $141$       & $3.0$ \\
$1024$ & $10{,}000$ & $559$   & $115{,}550$   & $2{,}514$   & $207$ \\
\bottomrule
\end{tabular}
\end{table}

\clearpage
\section{GSM8K Extended Results}
\label{app:gsm8k_table}

\begin{table}[h]
\centering
\small
\caption{\textbf{GSM8K per-chain accuracy vs.\ pass@$\K$} with $\K{=}10$. All rows use 100 pools at local sampling temperature $1.2$ with the plain Qwen3-Embedding-0.6B kernel (no task instruction; \cref{app:embedder}); the two PoolDPP rows are at each mode's highest pass@$\K$ in \cref{fig:gsm8k_pareto}. Values are means over 50 questions and 100 pools.}
\label{tab:gsm8k_meank_passk}
\begin{tabular}{lcc}
\toprule
Method & Per-chain acc. & Pass@$\K$ \\
\midrule
Ancestral            & 0.682 & 0.925 \\
\algnameKM           & 0.560 & 0.905 \\
\algnameSamp ($T{=}1.05$) & 0.565 & 0.915 \\
\algnameMAP ($T{=}500$)  & 0.502 & 0.915 \\
\bottomrule
\end{tabular}
\end{table}

\subsection{Embedding Geometry and Solution Diversity on GSM8K}
\label{app:embedder}

\paragraph{Embedder probe.}
We test whether an embedder separates correctness-critical changes from surface changes with paired edits whose labels are verified: on SQL by executing both queries, on GSM8K by the final answer and the stated arithmetic.

Each probe item is a pool string $\yy$ (Gemma-2-9B-Instruct, local sampling temperature $1$, one pool) with one critical edit $\yy_{\mathrm{c}}$ and one surface edit $\yy_{\mathrm{s}}$. On GSM8K, the critical edit flips the final answer token, or replaces the operator of the last correct arithmetic step (e.g., $+ \leftrightarrow -$) and shifts the stated final answer accordingly. The surface edit renames list markers, substitutes connectives (``So'' $\to$ ``Therefore''), and normalizes whitespace, and it is kept only if the final answer and the stated arithmetic are unchanged. On Ambrosia, the critical edit changes one predicate (e.g., \texttt{AND}~$\leftrightarrow$~\texttt{OR}, \texttt{=}~$\to$~\texttt{!=}), and the surface edit lowercases keywords and normalizes whitespace. An item is kept only if the critical edit changes the query's execution result and the surface edit does not.

With $d(\yy, \yy') = 1 - \embed(\yy)^\top \embed(\yy')$, the separation AUC (area under the ROC curve) is $\Pr[d(\yy, \yy_{\mathrm{c}}) > d(\yy', \yy'_{\mathrm{s}})]$ for independently drawn items $\yy, \yy'$, with ties counted as $1/2$. A value of $0.5$ means that the embedder cannot tell the two kinds of edit apart.

The instructed variant embeds the unchanged string in the Qwen3-Embedding query format with a one-line instruction. For GSM8K it reads \emph{``Given a step-by-step solution to a math word problem, represent it by its final numeric answer and the sequence of arithmetic operations it performs, ignoring wording, phrasing, and formatting.''}; for SQL, it asks to represent the query by the result set it returns.

Under the baseline GSM8K kernel (Qwen3-Embedding-0.6B), surface edits move the embedding further than correctness-critical edits (\cref{tab:embedder-probe}): the AUC is $0.051$ for answer flips and $0.353$ for operator flips. MiniLM-L6 has lower AUCs on both GSM8K edit types. On SQL, MiniLM-L6 reaches AUC $1.000$, but its distance for the SQL surface edits is zero up to floating-point error, consistent with its uncased tokenizer, so this value follows from the choice of surface edits. The instruction raises the Qwen3-Embedding-0.6B AUC to $0.151$ (answer flips), $0.842$ (operator flips), and $0.892$ (SQL). A one-hot encoding of the extracted final answer separates both GSM8K edit types by construction.

\begin{table}[h]
\centering
\small
\caption{\textbf{Embedder probe.} Separation AUC $= \Pr[d_{\mathrm{critical}} > d_{\mathrm{surface}}]$, where $d$ is the cosine distance between an edited string and its original; $0.5$ means the embedder does not distinguish the two edit types, and values below $0.5$ mean surface edits move the embedding further. Probe strings come from one Gemma-2-9B-Instruct pool at local sampling temperature $1$ ($n = 582$, $549$, and $468$ items per column); all embedders score the same items. Brackets: 95\% percentile bootstrap CIs over probe items ($2{,}000$ resamples; resampling ignores clustering of items by question).}
\label{tab:embedder-probe}
\begin{tabular}{lccc}
\toprule
Embedder & GSM8K answer flip & GSM8K operator flip & SQL semantic edit \\
\midrule
MiniLM-L6 (Ambrosia kernel)         & $0.002$ {\scriptsize $[0.001, 0.004]$} & $0.111$ {\scriptsize $[0.093, 0.131]$} & $1.000$ {\scriptsize $[1.000, 1.000]$} \\
Qwen3-Emb.-0.6B (GSM8K kernel)      & $0.051$ {\scriptsize $[0.040, 0.063]$} & $0.353$ {\scriptsize $[0.321, 0.384]$} & $0.740$ {\scriptsize $[0.708, 0.770]$} \\
Qwen3-Emb.-0.6B, instructed         & $0.151$ {\scriptsize $[0.130, 0.175]$} & $0.842$ {\scriptsize $[0.818, 0.866]$} & $0.892$ {\scriptsize $[0.868, 0.917]$} \\
\end{tabular}
\end{table}

\begin{figure}[t]
\centering
\includegraphics[width=0.7\columnwidth]{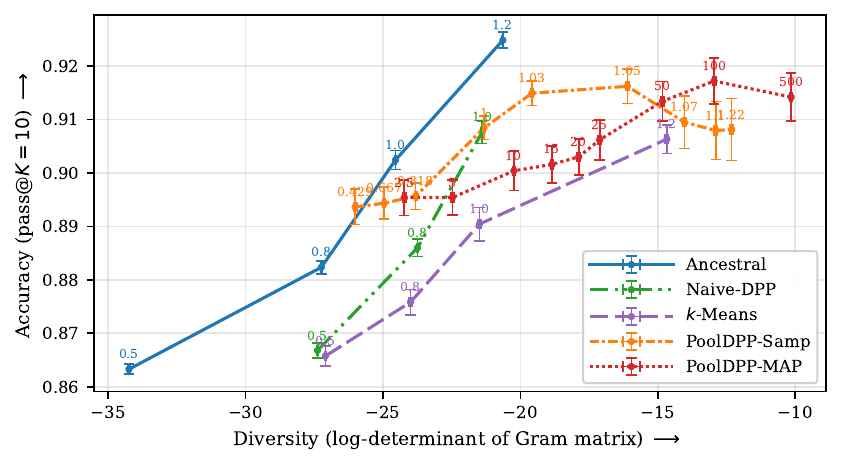}
\caption{\textbf{Math reasoning (GSM8K) with the instructed kernel.} Diversity (log-det Gram matrix) vs.\ pass@$\K{=}10$, on the $100$ pools of \cref{tab:gsm8k-kernels}. Ancestral, naive DPP, and $\K$-means sweep the local sampling temperature. \algnameSamp and \algnameMAP hold local sampling temperature $1.2$ and sweep the global temperature $T$. Error bars are $95\%$ $t$-intervals over pools. The instruction also defines the diversity axis, so the horizontal axis is not comparable with \cref{fig:gsm8k_pareto}.}
\label{fig:gsm8k_instructed}
\end{figure}

\paragraph{Instructed kernel downstream.}
\Cref{tab:gsm8k-kernels} runs the same selection on the same $100$ pools (local sampling temperature $1.2$, $\K{=}10$, $50$ questions) under the plain and the instructed kernel; \cref{fig:gsm8k_instructed} shows the frontier under the instructed kernel. The instruction raises the number of distinct final answers that \algnameMAP selects ($1.75 \to 1.94$ per set at $T{=}500$, $1.65 \to 1.84$ at $T{=}50$), but does not raise pass@$10$: no \algname setting under either kernel exceeds ancestral sampling at the matched local temperature, and the differences at $T{=}10$ are negative (question-paired $p_{\text{q}} = 0.016$, \cref{app:significance}).

Appending a one-hot encoding of each chain's extracted final answer to the embedding (weight $0.99$) raised \algnameMAP at $T{=}25$ to a pass@$10$ of $0.975$, against $0.925$ for ancestral sampling and a pool ceiling of $0.987$ (question-paired $p = 0.0002$). This kernel encodes the evaluation's answer key, so it is a domain-specific reference, not a general method.

\begin{table}[t]
\centering
\footnotesize
\setlength{\tabcolsep}{3.5pt}
\caption{\textbf{GSM8K under the plain and the instructed kernel} (Gemma-2-9B-Instruct, local sampling temperature $1.2$, $\K{=}10$, $N{=}100$, $100$ pools, $50$ questions). Pass@10 $\pm$ a 95\% $t$-interval over pools; per-chain accuracy as in \cref{sec:gsm8k}; \emph{answers} is the number of distinct canonicalized final answers per set. $\Delta$: pass@10 minus ancestral, paired over questions (positive favors the row), with the two-sided sign-flip $p$-value ($10^5$ flips). Log-det is measured in each row's own kernel and is not comparable across the two blocks.}

\label{tab:gsm8k-kernels}
\begin{tabular}{lcccccc}
\toprule
Method & pass@10 & per-chain acc. & Answers & log-det & $\Delta$ pass@10 & $p_{\mathrm{q}}$ \\
\midrule
Ancestral & $0.925 \pm 0.002$ & $0.682$ & $1.54$ & $-20.0$ & & \\
\midrule
\multicolumn{7}{l}{\emph{Plain kernel} (Qwen3-Embedding-0.6B)} \\
\algnameSamp, $T{=}1.05$ & $0.915 \pm 0.003$ & $0.565$ & $1.70$ & $-15.8$ & $-0.0100$ & 0.12 \\
\algnameSamp, $T{=}1$ & $0.908 \pm 0.002$ & $0.701$ & $1.58$ & $-20.6$ & $-0.0169$ & 0.03 \\
\algnameMAP, $T{=}500$ & $0.915 \pm 0.005$ & $0.502$ & $1.75$ & $-9.6$ & $-0.0096$ & 0.028 \\
\algnameMAP, $T{=}50$ & $0.912 \pm 0.004$ & $0.626$ & $1.65$ & $-13.9$ &  &  \\
\algnameMAP, $T{=}10$ & $0.899 \pm 0.003$ & $0.697$ & $1.47$ & $-19.2$ & $-0.0258$ & 0.016 \\
\midrule
\multicolumn{7}{l}{\emph{Instructed kernel} (same embedder, task instruction)} \\
\algnameSamp, $T{=}1.05$ & $0.916 \pm 0.003$ & $0.560$ & $1.73$ & $-16.1$ & $-0.0086$ & 0.19 \\
\algnameSamp, $T{=}1$ & $0.908 \pm 0.002$ & $0.696$ & $1.62$ & $-21.3$ & $-0.0164$ & 0.029 \\
\algnameMAP, $T{=}500$ & $0.914 \pm 0.005$ & $0.459$ & $1.94$ & $-10.2$ & $-0.0106$ & 0.0029 \\
\algnameMAP, $T{=}50$ & $0.913 \pm 0.004$ & $0.597$ & $1.84$ & $-14.8$ &  &  \\
\algnameMAP, $T{=}10$ & $0.900 \pm 0.004$ & $0.690$ & $1.54$ & $-20.2$ & $-0.0244$ & 0.016 \\
\bottomrule
\end{tabular}
\end{table}

\paragraph{Role of the determinant on GSM8K.}
To test whether the selected chains differ in reasoning or only in wording, we split each selected set into its $45$ pairs and label every pair in which both chains state a final answer in the \texttt{\#\#\#\#} format. A pair is \emph{different answer} if the canonicalized answers differ (decided by the parser). Every same-answer pair is labeled \emph{same answer, different derivation} or \emph{surface only} by Gemini 3.6 Flash (temperature $0$), which is asked whether the two chains use different intermediate quantities or strategy, or the same underlying arithmetic.

On a stratified validation sample of $300$ pairs, the judge agreed with the parser on $100$ of $100$ different-answer pairs. For same-answer pairs, it agreed with a deterministic rule (different multiset of stated arithmetic facts) on $90$ of $100$ pairs that the rule calls surface only, but on only $13$ of $100$ that the rule calls different derivations (it labeled the other $87$ surface only). We therefore report the judge's labels for same-answer pairs.

The selections use the instructed kernel on one pool. \Cref{tab:gsm8k-pairs} shows that the different-answer share rises from $0.120$ (ancestral) to $0.307$ (\algnameMAP), and the surface-only share falls from $0.786$ to $0.579$ ($21$ points). The same-answer, different-derivation share is $0.09$--$0.12$ for every method: the determinant adds chains with different final answers, and it does not add more distinct derivations of the same answer.

\begin{table}[h]
\centering
\small
\caption{\textbf{Pair decomposition of selected GSM8K sets} (instructed Qwen3-Embedding-0.6B kernel, Gemma-2-9B-Instruct, $\K{=}10$, one pool, one selection run, $50$ questions). Shares of the labeled pairs; each row sums to $1$. \emph{Pairs} counts pairs in which both chains have a \texttt{\#\#\#\#} final answer, out of at most $50 \times 45 = 2{,}250$; the excluded pairs contain a chain without such an answer. \emph{Agreement}: fraction of same-answer pairs on which the judge's label equals that of the deterministic arithmetic-multiset rule. Ancestral and \algnameKM use local sampling temperature $1$; \algnameSamp and \algnameMAP use local sampling temperature $1.2$ and $T{=}500$, so the rows are not at matched local temperature. No uncertainty estimates.}

\label{tab:gsm8k-pairs}
\begin{tabular}{lccccc}
\toprule
Method & \makecell{Different\\answer} & \makecell{Same answer,\\different derivation} & \makecell{Surface\\only} & Pairs & Agreement \\
\midrule
Ancestral                          & $0.120$ & $0.094$ & $0.786$ & $1{,}475$ & $0.59$ \\
\algnameKM                         & $0.157$ & $0.115$ & $0.728$ & $1{,}703$ & $0.60$ \\
\algnameSamp, $T{=}500$   & $0.245$ & $0.116$ & $0.639$ & $812$     & $0.60$ \\
\algnameMAP, $T{=}500$    & $0.307$ & $0.114$ & $0.579$ & $466$     & $0.50$ \\
\bottomrule
\end{tabular}
\end{table}

\subsection{Saturation Check: MATH-500}
\label{app:math500}

On GSM8K, ancestral pass@$10$ is $94\%$ of the pool ceiling, the fraction of pools that contain a correct chain ($0.925$ vs.\ $0.987$ at local sampling temperature $1.2$), which leaves a selector little to gain. MATH-500 \citep{hendrycks2021math,lightman2023lets} has more headroom. We ran the same pipeline on $100$ of its problems ($20$ per difficulty level), with Gemma-2-9B-Instruct, $10$ pools of $N{=}100$ at each local sampling temperature $0.5$, $0.8$, and $1.2$, Qwen3-Embedding-0.6B with the GSM8K instruction of \cref{app:embedder} reworded to ``math problem'', ``final answer'', and ``mathematical operations'', and answers checked with the \href{https://github.com/huggingface/math-verify}{\code{math-verify}} package. At local sampling temperature $1.2$, ancestral pass@$10$ is $0.735$, $83\%$ of the pool ceiling of $0.889$, and a pool holds $21.5$ distinct answers on average.

\begin{figure}[ht]
\centering
\includegraphics[width=0.75\textwidth]{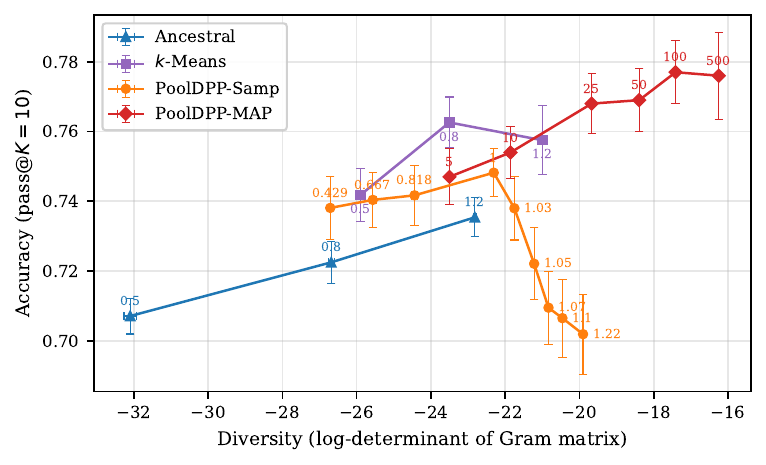}
\caption{\textbf{MATH-500 saturation check} (Gemma-2-9B-Instruct, $100$ problems, $\K{=}10$, $10$ pools of $N{=}100$, instructed Qwen3-Embedding-0.6B kernel). Diversity (log-det of the Gram matrix) vs.\ pass@$10$, with $95\%$ bootstrap intervals over pools. Ancestral sampling and \algnameKM vary the local sampling temperature $T'$. \algname varies the global temperature $T$ on the pools drawn at $T'{=}1.2$. Point labels give the varied temperature.}
\label{fig:math500}
\end{figure}

\algnameMAP reaches the highest pass@$10$ (\cref{fig:math500}): $0.777$ at $T{=}100$, $4.2$ points above ancestral sampling at local sampling temperature $1.2$ (question-paired $p_{\text{q}} < 0.001$, \cref{app:significance}), at log-det $-17.4$ against $-22.8$. The gain comes from levels 3 to 5 ($+8.6$, $+5.2$, and $+5.5$ points), against $+0.5$ and $+1.1$ points at levels 1 and 2. \algnameKM reaches $0.763$ at local sampling temperature $0.8$ (log-det $-23.5$), and the test does not resolve its $1.4$-point gap to \algnameMAP ($p_{\text{q}}{=}0.12$). \algnameSamp improves on ancestral sampling only at $T{=}1$ ($+1.3$ points, $p_{\text{q}}{=}0.02$). Above $T{=}1.03$ its pass@$10$ falls below that of ancestral sampling, down to $0.702$ at $T{=}1.22$ ($-3.4$ points, $p_{\text{q}}{=}0.01$, log-det $-19.9$).

\clearpage
\section{Beam-Search Baselines}
\label{app:beam_baselines}

Diverse beam search \citep{vijayakumar2018diversebs} and determinantal beam search \citep{meister-etal-2021-determinantal} are both designed to fix vanilla beam search's tendency to return near-identical hypotheses. Both succeed at that narrow goal, but neither matches sampling-based methods on the quality--diversity trade-off: DivBS has lower semantic and token diversity, and DetBS reaches sampling-level diversity only by collapsing quality. \Cref{tab:beam_baselines} reports both alongside ancestral sampling and representative \algname{} operating points; the sampling rows are the lowest-temperature points we ran.

\begin{table}[h]
\centering
\small
\setlength{\tabcolsep}{4pt}
\caption{\textbf{Beam-search baselines vs.\ \algname{} (intrinsic, FineWeb).} $G$: number of beam groups of DivBS. Quality is the mean per-sequence $\log\plm(\yy\mid\xx)$; \emph{semantic diversity} is $\log\det(\Embs^\top\Embs + 10^{-6} I)$, where $\Embs$ collects the $\K{=}10$ unit-norm sentence embeddings as columns; \emph{token diversity} is Distinct-2, the fraction of unique whitespace bigrams across the $\K$ outputs. Means over $50$ prompts (the sampling rows: one pool, $500$ sets for ancestral sampling and \algnameSamp, $50$ for \algnameMAP), Gemma-2-9B-Instruct with MiniLM-L6 embeddings; higher is better on all three axes.}
\label{tab:beam_baselines}
\begin{tabular}{lccc}
\toprule
\textbf{Method} & \textbf{Quality}~$\uparrow$ & \makecell{\textbf{Sem.\ div.}\\$\log\det \Embs^\top\Embs \uparrow$} & \makecell{\textbf{Tok.\ div.}\\\textbf{Distinct-2} $\uparrow$} \\
\midrule
\hyperref[item:divbs]{DivBS} ($G{=}10$) & $-15.0$ & $-15.1$ & $0.58$ \\
\hyperref[item:detbs]{DetBS} ($w_{\mathrm{DetBS}}{=}1.0$) & $-161.8$ & $\phantom{0}-5.2$ & $0.87$ \\
\midrule
Ancestral ($T'{=}0.6$) & $\phantom{0}-39.6$ & $\phantom{0}-8.4$ & $0.83$ \\
\algnameSamp{} ($T{=}0.43$) & $\phantom{0}-33.6$ & $\phantom{0}-6.4$ & $0.85$ \\
\algnameMAP{} ($T{=}10$) & $\phantom{0}-34.4$ & $\phantom{0}-4.4$ & $0.89$ \\
\bottomrule
\end{tabular}
\end{table}

DivBS keeps quality high ($-15.0$, comparable to greedy beam search) but has the lowest semantic and token diversity of all rows. DetBS reaches sampling-level diversity only by collapsing quality to $-161.8$, against $-33.6$ to $-39.6$ for the sampling rows. \algnameMAP{} at $T{=}10$ matches DetBS on both diversity metrics with quality $-34.4$ instead of $-161.8$, and it has higher diversity than DivBS at lower quality ($-34.4$ against $-15.0$). \algnameSamp{} at $T{=}0.43$ likewise gives up $19$ nats of quality relative to DivBS for higher diversity.

\paragraph{Configurations.}
For DivBS, we use \code{num_beams=10}, \code{num_beam_groups=10}, \code{diversity_penalty=1.0}, and \code{do_sample=False}, via the \href{https://huggingface.co/transformers-community/group-beam-search}{\code{transformers-community/group-beam-search}} HuggingFace \code{custom_generate} module. For DetBS, we use a port of \citet{meister-etal-2021-determinantal} with the string-subsequence kernel of subsequence length $n{=}2$ and gap decay $\lambda_{\mathrm{DetBS}}{=}0.5$, \code{beam_size=10}, \code{sub_beam_size=10}, and decoding temperature $T'{=}1.0$. The interaction weight $w_{\mathrm{DetBS}}$ scales the kernel against the candidate scores, and $w_{\mathrm{DetBS}}{=}0$ is beam search. The DetBS rows of \cref{tab:baselines} use a cosine Gram matrix over hashed $n$-gram counts instead (\cref{app:baselines}).

We swept $w_{\mathrm{DetBS}} \in \{0.05, 0.1, 0.3, 0.5, 1.0, 2.0\}$ and $T' \in \{0.7, 1.0, 1.3\}$ on the first $8$ prompts. The resulting $T' \times w_{\mathrm{DetBS}}$ Pareto frontier sits strictly below ancestral sampling at every diversity level. The $w_{\mathrm{DetBS}}{=}1.0$ row, run on all $50$ prompts, is the operating point on that frontier closest to the sampling rows on diversity.

The code, the sweep, and the outputs are in \texttt{baselines/detbs\_ssk} and \texttt{analysis/beam\_baselines\_intrinsic.json} of the repository.

\paragraph{Why DetBS falls short.}
DetBS's string-subsequence kernel measures bigram overlap, not semantic similarity, so it promotes a different kind of diversity than the embedding-based metric rewards.

\clearpage
\section{Set-Level Baselines, Significance Tests, and Pool Coverage}
\label{app:rebuttal-evidence}

\subsection{Set-Level Baselines}
\label{app:baselines}

We compare \algname with five set-level baselines. MMR and MBR rerank the same pools, embeddings, and $\K$ as \algname; SBS, CPSBS, and DetBS are decoding-time methods that produce a size-$\K$ set directly from the language model.
\begin{itemize}[nosep,leftmargin=*]
\item \textbf{MMR} \citep{carbonell1998mmr} starts from the highest-likelihood pool item and greedily adds the item maximizing $\lambda_{\mathrm{MMR}}\ell(\yy) - (1-\lambda_{\mathrm{MMR}})\max_{\yy' \in \set}\cos(\embedy, \embed(\yy'))$, where $\ell$ is $\log\plm$ min--max normalized within the pool. We sweep $\lambda_{\mathrm{MMR}} \in \{0, 0.1, 0.3, 0.5, 0.7, 0.9, 1\}$; $\lambda_{\mathrm{MMR}}{=}1$ is quality top-$\K$.
\item \textbf{MBR} \citep{kumar-byrne-2004-minimum} with a cosine-similarity utility returns the $\K$ pool items with the highest mean cosine similarity to the pool. Its only setting is the pool's local sampling temperature.
\item \textbf{Stochastic beam search (SBS)} \citep{kool2019stochasticbeamsthemgumbeltopk} draws $\K$ distinct sequences without replacement from the locally annealed model with the Gumbel-top-$\K$ trick. We sweep the local sampling temperature $T'$.
\item \textbf{Conditional Poisson SBS (CPSBS)} \citep{meister-etal-2021-conditional} replaces the per-step top-$\K$ of beam search with a conditional Poisson sample of size $\K$ over the candidate extensions, with inclusion probabilities set by the annealed prefix probabilities. We sweep $T'$ and draw each per-step sample from the highest-weight candidates, which changes the per-step distribution by at most $10^{-7}$ in total variation.
\item \textbf{Determinantal beam search (DetBS)} \citep{meister-etal-2021-determinantal} replaces the per-step top-$\K$ with a greedy $\K$-DPP MAP step on the candidate scores plus $w_{\mathrm{DetBS}}$ times a similarity Gram matrix; $w_{\mathrm{DetBS}}{=}0$ is beam search. We use a cosine Gram matrix over hashed unigram and bigram counts, the gapless limit of the string-subsequence kernel, whose cost is quadratic in sequence length. We sweep $w_{\mathrm{DetBS}}$.
\end{itemize}
SBS and CPSBS use $10$ independent runs per prompt and temperature; DetBS is deterministic. Our SBS, CPSBS, and DetBS implementations reproduce exact enumeration on small enumerable models, and the released code includes these tests.

\Cref{tab:baselines} reports each method at its best setting over its own sweep: the highest pass@10 on GSM8K (ties broken by per-chain accuracy), the highest recall on Ambrosia, and on FineWeb the most diverse setting whose quality is at least that of \algnameKM at local sampling temperature $1$ ($-40.3$ nats). MBR has a single FineWeb point, below this quality, and is shown there.

On Ambrosia, \algnameMAP has the highest recall ($0.251$), followed by MMR ($0.246$), \algnameSamp ($0.244$), and \algnameKM ($0.240$). The best decoding-time recall is SBS at $0.235$, against $0.233$ for ancestral sampling; SBS, CPSBS, DetBS, and MBR reach higher precision ($0.392$--$0.417$ against $0.322$ for \algnameMAP).

On GSM8K, SBS at $T'{=}1.2$ has the highest pass@10 ($0.938 \pm 0.009$) and $2.91$ distinct final answers per set, against $0.912$--$0.915$ and $1.68$--$1.75$ for MMR and both \algname modes; CPSBS and DetBS reach at most $0.820$. MBR has the highest per-chain accuracy ($0.773$) and, among the reranking methods, the fewest distinct answers ($1.26$) and the lowest log-determinant ($-29.6$), because it selects the chains that agree with the pool consensus.

On FineWeb, among settings above the quality threshold, \algnameMAP has the highest log-determinant ($-2.36$) and MMR the next ($-3.75$, at $2.3$ nats higher quality), so neither dominates the other; SBS at $T'{=}0.6$ matches ancestral sampling at the same temperature in quality ($-39.5$ vs.\ $-39.4$), and CPSBS and DetBS stay below $-25$.

\begin{table}[t]
\centering
\footnotesize
\setlength{\tabcolsep}{4pt}
\caption{\textbf{Set-level baselines}, each at its best setting over its own sweep (selection rule in \cref{app:baselines}). Gemma-2-9B-Instruct throughout. \emph{GSM8K}: $\K{=}10$, $50$ questions; pool-based methods use $100$ pools of $N{=}100$ at local sampling temperature $1.2$; answers/set counts distinct numerically canonicalized final answers; log-det uses the Qwen3-Embedding-0.6B Gram matrix. \emph{Ambrosia}: $\K{=}3$, $100$ questions, $100$ pools of $N{=}200$; precision requires the whole execution result to match a golden result. \emph{FineWeb}: $\K{=}10$, $50$ prompts, $10$ pools of $N{=}400$ ( the pools of \cref{fig:fineweb_pareto}); quality is mean $\log\plm(\yy \mid \xx)$ in nats and log-det uses MiniLM-L6. $T'$ is the local sampling temperature of the pool or, for SBS and CPSBS, of the decoder, and $T$ the global temperature; $\lambda_{\mathrm{MMR}}$ is the MMR trade-off and $w_{\mathrm{DetBS}}$ the DetBS interaction weight. $\pm$ is a 95\% confidence interval over $10$ independent decoding runs (SBS, CPSBS) or, for the pool-based rows, a 95\% $t$-interval of the per-pool means over the $100$ (GSM8K, Ambrosia), or $10$ (FineWeb) pools.}
\label{tab:baselines}
\begin{tabular}{@{}llcccc@{}}
\toprule
Method & Setting & & & & \\
\midrule
\multicolumn{6}{@{}l}{\emph{GSM8K}} \\
& & pass@10 & per-chain acc. & answers/set & log-det \\
\cmidrule(l){3-6}
Ancestral & $T'{=}1.2$ & 0.925 {\scriptsize $\pm 0.002$} & 0.682 {\scriptsize $\pm 0.001$} & 1.54 {\scriptsize $\pm 0.01$} & $-20.0$ {\scriptsize $\pm 0.0$} \\
\algnameKM & $T'{=}1.2$ & 0.905 {\scriptsize $\pm 0.003$} & 0.560 {\scriptsize $\pm 0.003$} & 1.52 {\scriptsize $\pm 0.01$} & $-13.8$ {\scriptsize $\pm 0.1$} \\
MMR & $\lambda_{\mathrm{MMR}}{=}0.3$ & 0.912 {\scriptsize $\pm 0.005$} & 0.506 {\scriptsize $\pm 0.004$} & 1.68 {\scriptsize $\pm 0.02$} & $-10.4$ {\scriptsize $\pm 0.1$} \\
MBR & -- & 0.913 {\scriptsize $\pm 0.003$} & 0.773 {\scriptsize $\pm 0.002$} & 1.26 {\scriptsize $\pm 0.01$} & $-29.6$ {\scriptsize $\pm 0.0$} \\
SBS & $T'{=}1.2$ & 0.938 {\scriptsize $\pm 0.009$} & 0.670 {\scriptsize $\pm 0.009$} & 2.91 {\scriptsize $\pm 0.09$} & $-19.9$ {\scriptsize $\pm 0.1$} \\
CPSBS & $T'{=}0.5$ & 0.816 {\scriptsize $\pm 0.011$} & 0.732 {\scriptsize $\pm 0.007$} & 1.28 {\scriptsize $\pm 0.03$} & $-48.8$ {\scriptsize $\pm 0.2$} \\
DetBS & $w_{\mathrm{DetBS}}{=}0.06$ & 0.820 & 0.708 & 1.30 & $-47.7$ \\
\algnameSamp & $T{=}1.05$ & 0.915 {\scriptsize $\pm 0.003$} & 0.565 {\scriptsize $\pm 0.003$} & 1.70 {\scriptsize $\pm 0.01$} & $-15.8$ {\scriptsize $\pm 0.1$} \\
\algnameMAP & $T{=}500$ & 0.915 {\scriptsize $\pm 0.005$} & 0.502 {\scriptsize $\pm 0.004$} & 1.75 {\scriptsize $\pm 0.02$} & $-9.6$ {\scriptsize $\pm 0.1$} \\
\midrule
\multicolumn{6}{@{}l}{\emph{Ambrosia}} \\
& & recall & precision & & \\
\cmidrule(l){3-4}
Ancestral & $T'{=}1.0$ & 0.233 {\scriptsize $\pm 0.001$} & 0.394 {\scriptsize $\pm 0.001$} &  &  \\
\algnameKM & $T'{=}1.0$ & 0.240 {\scriptsize $\pm 0.001$} & 0.364 {\scriptsize $\pm 0.001$} &  &  \\
MMR & $\lambda_{\mathrm{MMR}}{=}0.3$, $T'{=}1.0$ & 0.246 {\scriptsize $\pm 0.002$} & 0.331 {\scriptsize $\pm 0.002$} &  &  \\
MBR & $T'{=}1.2$ & 0.191 {\scriptsize $\pm 0.001$} & 0.395 {\scriptsize $\pm 0.002$} &  &  \\
SBS & $T'{=}1.0$ & 0.235 {\scriptsize $\pm 0.007$} & 0.392 {\scriptsize $\pm 0.008$} & & \\
CPSBS & $T'{=}0.6$ & 0.207 {\scriptsize $\pm 0.004$} & 0.416 {\scriptsize $\pm 0.009$} & & \\
DetBS & $w_{\mathrm{DetBS}}{=}0.4$ & 0.227 & 0.417 & & \\
\algnameSamp & $T{=}0.82$, $T'{=}1.0$ & 0.244 {\scriptsize $\pm 0.001$} & 0.375 {\scriptsize $\pm 0.001$} &  &  \\
\algnameMAP & $T{=}10$, $T'{=}1.5$ & 0.251 {\scriptsize $\pm 0.002$} & 0.322 {\scriptsize $\pm 0.002$} &  &  \\
\midrule
\multicolumn{6}{@{}l}{\emph{FineWeb}} \\
& & quality & log-det & & \\
\cmidrule(l){3-4}
Ancestral & $T'{=}0.6$ & $-39.4$ {\scriptsize $\pm 0.1$} & $-8.45$ {\scriptsize $\pm 0.13$} & & \\
\algnameKM & $T'{=}1.0$ & $-40.3$ {\scriptsize $\pm 0.2$} & $-4.70$ {\scriptsize $\pm 0.04$} & & \\
MMR & $\lambda_{\mathrm{MMR}}{=}0.7$, $T'{=}1.0$ & $-37.7$ {\scriptsize $\pm 0.3$} & $-3.75$ {\scriptsize $\pm 0.28$} & & \\
MBR & $T'{=}1.0$ & $-68.7$ {\scriptsize $\pm 0.4$} & $-11.91$ {\scriptsize $\pm 0.08$} & & \\
SBS & $T'{=}0.6$ & $-39.5$ {\scriptsize $\pm 0.4$} & $-7.96$ {\scriptsize $\pm 0.17$} & & \\
CPSBS & $T'{=}0.6$ & $-26.2$ {\scriptsize $\pm 0.4$} & $-41.46$ {\scriptsize $\pm 0.56$} & & \\
DetBS & $w_{\mathrm{DetBS}}{=}0.1$ & $-38.4$ & $-25.63$ & & \\
\algnameSamp & $T{=}0.82$, $T'{=}1.0$ & $-37.7$ {\scriptsize $\pm 0.3$} & $-5.71$ {\scriptsize $\pm 0.07$} &  &  \\
\algnameMAP & $T{=}50$, $T'{=}1.0$ & $-40.0$ {\scriptsize $\pm 0.4$} & $-2.36$ {\scriptsize $\pm 0.04$} &  &  \\
\bottomrule
\end{tabular}
\end{table}

\clearpage
\subsection{Paired Significance Tests}
\label{app:significance}

Every method is evaluated on the same pools and questions, so we test each comparison with a paired sign-flip permutation test instead of comparing the marginal confidence intervals of \cref{fig:ambrosia_pareto,fig:gsm8k_pareto}. For methods $A$ and $B$, we compute one paired difference $A - B$ per unit and its mean over units. The null distribution flips the sign of each unit's difference independently with probability $1/2$ ($100{,}000$ random sign vectors), and $p$ is the two-sided fraction of draws whose absolute mean is at least the observed one, so $p$ cannot resolve below $10^{-5}$. Positive differences favor $A$, the \algname variant.

We use two pairing units. Question-paired tests average each method over pools and runs, giving one value per question ($n{=}50$ on GSM8K, $n{=}100$ on Ambrosia), and test whether the difference holds over questions. Pool-paired tests average over questions, giving one value per pool ($n{=}100$), and test reproducibility over pool draws for a fixed question set. We treat the question-paired test as primary. The reported interval is a $95\%$ percentile bootstrap over questions ($10{,}000$ resamples) of the mean paired difference, which is the same under both pairings.

\Cref{tab:significance} lists every comparison. On GSM8K, with either kernel, no tested \algname setting improves pass@10 over ancestral sampling at the matched local sampling temperature $1.2$, and \algnameMAP at $T{=}500$ does not significantly exceed \algnameKM under the question-paired test.

On Ambrosia, \algnameSamp at $T{=}0.82$ and \algnameMAP at $T{=}10$ have higher recall and lower precision than ancestral sampling at its highest-recall local temperature (recall $p_{\text{q}} = 6{\times}10^{-5}$ and $0.015$). Against \algnameKM at local temperature $1$, the question-paired test does not resolve the recall difference of either mode, while the pool-paired test resolves both.

\begin{table}[t]
\centering
\footnotesize
\setlength{\tabcolsep}{4pt}
\caption{\textbf{Paired differences}, first method minus second; positive values favor the first. Set metric: pass@10 (GSM8K) or recall (Ambrosia). Item metric: per-chain accuracy (GSM8K) or precision (Ambrosia). CI: $95\%$ bootstrap interval over questions of the set-metric difference. $p_{\text{q}}$ and $p_{\text{pool}}$: two-sided sign-flip permutation tests with $10^{5}$ random sign vectors, paired over questions and over pools; $<10^{-5}$ marks the resolution limit. \emph{GSM8K}: Gemma-2-9B-Instruct, $\K{=}10$, $50$ questions, $100$ pools of $N{=}100$; every method draws its pool at local sampling temperature $T'{=}1.2$. \emph{Ambrosia}: $\K{=}3$, $100$ questions, $100$ pools of $N{=}200$. }

\label{tab:significance}
\begin{tabular}{@{}lrcllrl@{}}
\toprule
& \multicolumn{4}{c}{set metric} & \multicolumn{2}{c}{item metric} \\
\cmidrule(lr){2-5}\cmidrule(l){6-7}
Comparison & $\Delta$ & 95\% CI & $p_{\text{q}}$ & $p_{\text{pool}}$ & $\Delta$ & $p_{\text{q}}$ \\
\midrule
\multicolumn{7}{@{}l}{\emph{GSM8K, plain Qwen3-Embedding-0.6B kernel}} \\
\algnameSamp $T{=}1.05$ vs.\ ancestral & $-0.0100$ & $[-0.0221, +0.0018]$ & 0.12 & $<10^{-5}$ & $-0.116$ & $<10^{-5}$ \\
\algnameSamp $T{=}1$ vs.\ ancestral & $-0.0169$ & $[-0.0331, -0.0039]$ & 0.03 & $<10^{-5}$ & $+0.019$ & 0.0025 \\
\algnameMAP $T{=}500$ vs.\ ancestral & $-0.0096$ & $[-0.0185, -0.0017]$ & 0.028 & $6{\times}10^{-4}$ & $-0.180$ & $<10^{-5}$ \\
\algnameMAP $T{=}500$ vs.\ \algnameKM & $+0.0099$ & $[-0.0028, +0.0260]$ & 0.22 & $7{\times}10^{-5}$ & $-0.058$ & $<10^{-5}$ \\
\algnameMAP $T{=}10$ vs.\ ancestral & $-0.0258$ & $[-0.0487, -0.0071]$ & 0.016 & $<10^{-5}$ & $+0.015$ & 0.11 \\
\algnameMAP $T{=}10$ vs.\ \algnameKM & $-0.0063$ & $[-0.0145, +0.0004]$ & 0.12 & $5{\times}10^{-4}$ & $+0.137$ & $<10^{-5}$ \\
\midrule
\multicolumn{7}{@{}l}{\emph{GSM8K, instructed Qwen3-Embedding-0.6B kernel}} \\
\algnameSamp $T{=}1.05$ vs.\ ancestral & $-0.0086$ & $[-0.0211, +0.0033]$ & 0.19 & $<10^{-5}$ & $-0.122$ & $<10^{-5}$ \\
\algnameSamp $T{=}1$ vs.\ ancestral & $-0.0164$ & $[-0.0325, -0.0035]$ & 0.029 & $<10^{-5}$ & $+0.014$ & 0.015 \\
\algnameMAP $T{=}500$ vs.\ ancestral & $-0.0106$ & $[-0.0188, -0.0039]$ & 0.0029 & $1{\times}10^{-5}$ & $-0.223$ & $<10^{-5}$ \\
\algnameMAP $T{=}500$ vs.\ \algnameKM & $+0.0079$ & $[-0.0043, +0.0221]$ & 0.27 & $7{\times}10^{-4}$ & $-0.083$ & $<10^{-5}$ \\
\algnameMAP $T{=}10$ vs.\ ancestral & $-0.0244$ & $[-0.0466, -0.0064]$ & 0.016 & $<10^{-5}$ & $+0.008$ & 0.35 \\
\algnameMAP $T{=}10$ vs.\ \algnameKM & $-0.0059$ & $[-0.0161, +0.0026]$ & 0.25 & 0.0019 & $+0.149$ & $<10^{-5}$ \\
\midrule
\multicolumn{7}{@{}l}{\emph{Ambrosia}} \\
\algnameSamp $T{=}0.82$ vs.\ ancestral & $+0.0113$ & $[+0.0058, +0.0175]$ & $6{\times}10^{-5}$ & $<10^{-5}$ & $-0.019$ & 0.0012 \\
\algnameSamp $T{=}0.82$ vs.\ \algnameKM & $+0.0048$ & $[-0.0088, +0.0206]$ & 0.54 & $<10^{-5}$ & $+0.011$ & 0.21 \\
\algnameSamp $T{=}0.67$ vs.\ ancestral & $+0.0082$ & $[+0.0016, +0.0150]$ & 0.019 & $<10^{-5}$ & $-0.009$ & 0.14 \\
\algnameMAP $T{=}10$ vs.\ ancestral & $+0.0176$ & $[+0.0037, +0.0320]$ & 0.015 & $<10^{-5}$ & $-0.072$ & $1{\times}10^{-5}$ \\
\algnameMAP $T{=}10$ vs.\ \algnameKM & $+0.0111$ & $[-0.0020, +0.0255]$ & 0.12 & $<10^{-5}$ & $-0.042$ & $3{\times}10^{-4}$ \\
\bottomrule
\end{tabular}
\end{table}

\clearpage
\subsection{Ambrosia Pool Ceiling}
\label{app:pool-ceiling}

Every pool-based method selects from the pool, so its recall is bounded by the fraction of gold interpretations that the pool contains. We call this fraction the \emph{pool coverage} of a (question, pool) cell: a pool string covers a gold interpretation if executing it on the question's database returns exactly the gold query's result, the strict rule of \cref{sec:ambrosia}. The $100$ questions have $2$ ($56$ questions) or $3$ ($44$ questions) gold interpretations.

\Cref{tab:pool-ceiling}a gives the coverage by local sampling temperature. It is highest at $1.3$ ($0.418$), against $0.384$ at local temperature $1$, and it still increases from $N{=}100$ to $N{=}200$ at every local temperature, so it has not saturated at the pool size we use. Relative to this ceiling (\cref{tab:pool-ceiling}b), all four methods recover between $0.508$ (ancestral sampling) and $0.528$ (\algnameSamp) of their pools' coverage at their settings in \cref{tab:baselines}. Both baselines recover the largest fraction at the coldest pool, where coverage is lowest.

\begin{table}[h]
\centering
\small
\setlength{\tabcolsep}{5pt}
\caption{\textbf{Ambrosia pool ceiling} (Gemma-2-9B-Instruct, MiniLM-L6, $\K{=}3$, $100$ pools, $100$ questions). (a) Pool coverage: fraction of a question's gold interpretations matched by execution by at least one pool string, averaged over questions and pools; for $N < 200$, averaged over $10$ random subsamples of the pool. \emph{None} and \emph{All}: fraction of (question, pool) cells with coverage $0$ and $1$ at $N{=}200$. 95\% bootstrap CI half-widths over pools are at most $0.0023$ for every coverage entry. (b) Each method at its setting in \cref{tab:baselines}. Recall: fraction of gold interpretations matched by the $\K$ selected queries. Fraction of coverage: recall divided by the question's mean coverage over the $100$ pools at the same local temperature, averaged over questions with nonzero coverage, then over runs and pools. $\pm$: half-width of the 95\% bootstrap CI over the $100$ pools ($10^4$ resamples).}

\label{tab:pool-ceiling}
\textbf{(a)}\par\smallskip
\begin{tabular}{lccccc}
\toprule
Local temperature & $N{=}25$ & $N{=}100$ & $N{=}200$ & None & All \\
\midrule
$0.6$  & $0.277$ & $0.305$ & $0.319$ & $0.348$ & $0.020$ \\
$0.7$  & $0.286$ & $0.318$ & $0.334$ & $0.330$ & $0.030$ \\
$1.0$  & $0.311$ & $0.356$ & $0.384$ & $0.266$ & $0.065$ \\
$1.2$  & $0.325$ & $0.380$ & $0.413$ & $0.234$ & $0.083$ \\
$1.25$ & $0.326$ & $0.384$ & $0.416$ & $0.230$ & $0.084$ \\
$1.3$  & $0.326$ & $0.385$ & $0.418$ & $0.232$ & $0.086$ \\
$1.5$  & $0.300$ & $0.361$ & $0.391$ & $0.267$ & $0.063$ \\
\bottomrule
\end{tabular}

\medskip
\textbf{(b)}\par\smallskip
\begin{tabular}{lcccc}
\toprule
Method & $T'$ & Recall & Pool coverage & Fraction of coverage \\
\midrule
Ancestral & $1.0$ & $0.233 \pm 0.001$ & $0.384$ & $0.508 \pm 0.003$ \\
\algnameKM & $1.0$ & $0.240 \pm 0.001$ & $0.384$ & $0.521 \pm 0.002$ \\
\algnameSamp, $T{=}0.82$ & $1.0$ & $0.244 \pm 0.001$ & $0.384$ & $0.528 \pm 0.002$ \\
\algnameMAP, $T{=}10$ & $1.5$ & $0.251 \pm 0.002$ & $0.391$ & $0.516 \pm 0.004$ \\
\bottomrule
\end{tabular}
\end{table}

\clearpage
\subsection{Sequence Length and the Quality Term}
\label{app:length}

The quality term uses the raw sequence probability $\plm(\yy)$ without length normalization, so $\log\plm$ decreases with length: over the unique strings of one pool, the Pearson correlation between character length and $\log\plm$ is negative on every task and close to $-1$ on Ambrosia at local sampling temperature $1.5$ (\cref{tab:length}). Within a prompt, the coupling comes from a tail of long, low-quality strings: on Ambrosia at local temperature $1.5$, the longest fifth of each prompt's pool averages $2{,}742$ characters and $0.5\%$ of it executes, against $71\%$ of the shortest fifth.

Restricting each prompt's pool to its shorter half removes most of the coupling (the Spearman correlation $r_{\mathrm{s}}$ in \cref{tab:length}), and $\log\plm$ still ranks strings by quality there: the AUC changes little at local temperature $1$ and falls from $0.856$ to $0.747$ for golden Ambrosia queries at $1.5$, because part of the full-pool separation is the separation of the long, non-executing tail.

A length-only selection does not reproduce the FineWeb frontier. On pools drawn at local sampling temperature $1$, splitting each prompt's pool into $\K{=}10$ equal-width length bins and keeping the most probable string in each nonempty bin reaches quality $-41.8$ and log-det $-4.59$ (one pool), and \algnameMAP at $T{=}50$, on the same pool, is better on both axes, with quality $-39.7$ and log-det $-2.30$.

\begin{table}[h]
\centering
\footnotesize
\setlength{\tabcolsep}{3pt}
\caption{\textbf{Length and log-probability} (Gemma-2-9B-Instruct). $r$: Pearson correlation between character length and $\log\plm$ over all unique strings of one pool, pooled across prompts ($n = 4{,}999$ GSM8K strings; $15{,}045$ and $19{,}150$ Ambrosia strings at local temperatures $1.0$ and $1.5$). Remaining columns: statistics over the unique strings of each prompt's pool, averaged over prompts and the $100$ pools. \emph{Quality by length quintile}: fraction of strings with the quality label in each within-prompt length quintile, shortest (Q1) to longest (Q5); \emph{Chars}: mean characters in Q1 and Q5; $r_{\mathrm{s}}$: Spearman correlation between $\log\plm$ and length; AUC: $\Pr[\log\plm(\yy^+) > \log\plm(\yy^-)]$ for a string $\yy^+$ with the label and $\yy^-$ without it. Arrows: full pool $\to$ shorter half (length at most the prompt's median). $r$ and $r_{\mathrm{s}}$ do not depend on the label and are shown once per  task and local temperature. Single computation; no uncertainty estimates.}
\label{tab:length}
\begin{tabular}{llccccc}
\toprule
\makecell[lb]{Task,\\local temp.} & Label & $r$ & \makecell[cb]{Quality by length quintile\\(Q1 $\to$ Q5)} & \makecell[cb]{Chars\\Q1 / Q5} & $r_{\mathrm{s}}$ & AUC \\
\midrule
GSM8K, $1.2$     & correct  & $-0.730$ & $0.64,\ 0.72,\ 0.71,\ 0.71,\ 0.63$  & $218$ / $644$  & $-0.34 \to -0.02$ & $0.670 \to 0.663$ \\
Ambrosia, $1.0$  & executes & $-0.353$ & $0.91,\ 0.92,\ 0.91,\ 0.91,\ 0.87$  & $151$ / $248$  & $-0.25 \to -0.06$ & $0.801 \to 0.786$ \\
                 & golden   &          & $0.31,\ 0.39,\ 0.43,\ 0.42,\ 0.38$  &                &                   & $0.584 \to 0.594$ \\
Ambrosia, $1.5$  & executes & $-0.994$ & $0.71,\ 0.67,\ 0.48,\ 0.18,\ 0.005$ & $155$ / $2{,}742$ & $-0.82 \to -0.26$ & $0.966 \to 0.910$ \\
                 & golden   &          & $0.26,\ 0.32,\ 0.26,\ 0.10,\ 0.003$ &                &                   & $0.856 \to 0.747$ \\
\bottomrule
\end{tabular}
\end{table}

\paragraph{Generation length.}
\label{app:length-cap}
Our pools are drawn with a cap on generation length: at most $50$ new tokens on FineWeb, $800$ on Ambrosia, and $4{,}000$ tokens of prompt and output together on GSM8K. A string that reaches the cap without generating $\eos$ is kept and scored by its prefix probability. This is exactly \algname for the \defn{capped language model} $\plm_L$, which agrees with $\plm$ on strings that end within the cap and assigns each capped string its prefix probability. $\plm_L$ is a language model, so \cref{sec:analysis} applies to it as stated; the target is the $\K$-DPP with quality $\plm_L^{1/T}$.

\clearpage
\section{Qualitative Examples}
\label{app:qualitative}

We provide sample outputs comparing ancestral sampling and \algname. These are representative examples; aggregate statistics are in the main text.

\subsection{Open-Ended Generation (FineWeb)}

\paragraph{Example 1: Rolling Stones.}
Prompt: \emph{``The Rolling Stones in Concert has since been released solely by ABKCO Records. The band would remain incensed with Klein for decades for that act. Klein died in 2009.''}

\textbf{Ancestral sampling} produces continuations focused on the Klein controversy:
\begin{itemize}[nosep,leftmargin=*]
\item ``Following Klein's death, the Stones published their memoir `Life' in 2010...''
\item ```There was a lot of bad blood,' Scorsese told TheWrap...''
\item ``The Rolling Stones have a tumultuous relationship with their music catalog...''
\end{itemize}

\textbf{\algname} covers distinct topics:
\begin{itemize}[nosep,leftmargin=*]
\item ``The Rolling Stones have played many legendary concerts at Madison Square Garden...'' \emph{(concert venues)}
\item ``The Rolling Stones haven't sought to reclaim the rights to the Monterey Pop concert footage...'' \emph{(specific concert)}
\item ``The live album `Get Yer Ya-Ya's Out!' also remains controversial...'' \emph{(specific album)}
\item ``Early in their career, the band signed with Decca Records...'' \emph{(different era)}
\end{itemize}

\paragraph{Example 2: Creative writing.}
Prompt: \emph{``It is a field that is both fascinating, misunderstood, and in some cases resented. Arguably the best division in the Foundation.''}

\textbf{Ancestral sampling} shows variance in quality, including off-topic drift:
\begin{itemize}[nosep,leftmargin=*]
\item ``Xenosanthropologists dedicate their lives to the study of alien life...''
\item ``The Division of Genetics holds the secrets to life itself...''
\item ``If you mean a good deal on a product, you would tell someone that you got a steal. Knives, particularly Damascus steel knives...'' \emph{(off-topic drift)}
\end{itemize}

\textbf{\algname} produces diverse yet coherent continuations, each picking a distinct ``division'':
\begin{itemize}[nosep,leftmargin=*]
\item ``Black Ops---the shadows, the rumors, the whispered commands...'' \emph{(covert operations)}
\item ``I speak of course of the Screws... tasked with maintaining order...'' \emph{(social control)}
\item ``I'm talking about Ethics. What good is power without a moral compass?'' \emph{(philosophy)}
\item ``The Genesis Project is about ensuring survival for millions of years...'' \emph{(long-term planning)}
\end{itemize}

\algname produces structured diversity (meaningful variations) rather than noisy variance (random drift including failure modes like the knife tangent).

\clearpage
\subsection{SQL Generation (Ambrosia)}

\paragraph{Example 1: Airport terminals.}
Query: \emph{``Which terminals are in airports?''}

This query is ambiguous between universal quantification (terminals in \emph{all} airports) and existential quantification (terminals in \emph{any} airport).

\textbf{Ancestral sampling} produces only the universal interpretation:
\begin{itemize}[nosep,leftmargin=*]
\item \texttt{SELECT terminal\_id FROM Airports\_Terminals GROUP BY terminal\_id HAVING COUNT(DISTINCT airport\_id) = (SELECT COUNT(*) FROM Airports);}
\item \texttt{SELECT DISTINCT t.id FROM Terminals t JOIN Airports\_Terminals at ON t.id = at.terminal\_id GROUP BY t.id HAVING COUNT(DISTINCT at.airport\_id) = (SELECT COUNT(*) FROM Airports);}
\item (All 3 samples use variants of the \texttt{GROUP BY ... HAVING COUNT} pattern)
\end{itemize}

\textbf{\algname} captures both interpretations:
\begin{itemize}[nosep,leftmargin=*]
\item \texttt{SELECT terminal\_id FROM Airports\_Terminals GROUP BY terminal\_id HAVING COUNT(DISTINCT airport\_id) = (SELECT COUNT(*) FROM Airports);} \emph{(universal: in all airports)}
\item \texttt{SELECT DISTINCT terminal\_id FROM Airports\_Terminals;} \emph{(existential: in any airport)}
\end{itemize}

\paragraph{Example 2: Fashion designs.}
For the query \textit{``Which terminals are in airports?''}, ancestral sampling produces three variants of the universal interpretation (terminals in \emph{all} airports):
{\small\texttt{... GROUP BY terminal\_id HAVING COUNT(DISTINCT airport\_id) = (SELECT COUNT(*) FROM Airports)}}.
\algname captures both the universal and the existential reading (terminals in \emph{any} airport): {\small\texttt{SELECT DISTINCT terminal\_id FROM Airports\_Terminals}}.

Query: \emph{``What are the fashion designs?''}

\textbf{Ancestral sampling} repeats the same interpretation:
\begin{itemize}[nosep,leftmargin=*]
\item \texttt{SELECT design\_name FROM Fashion\_Design;} \emph{(just names)}
\item \texttt{SELECT design\_name, color\_name, fabric\_type FROM Fashion\_Design JOIN Color\_Palette ON ... JOIN Fabric\_Choice ON ...;} \emph{(enriched)}
\item \texttt{SELECT design\_name FROM Fashion\_Design;} \emph{(duplicate)}
\end{itemize}

\textbf{\algname} covers three distinct interpretations:
\begin{itemize}[nosep,leftmargin=*]
\item \texttt{SELECT design\_name FROM Fashion\_Design;} \emph{(projection: just names)}
\item \texttt{SELECT * FROM Fashion\_Design;} \emph{(raw table dump)}
\item \texttt{SELECT design\_name, color\_name, fabric\_type FROM Fashion\_Design fd JOIN Color\_Palette cp ON ... JOIN Fabric\_Choice fc ON ...;} \emph{(human-interpretable descriptions via joins)}
\end{itemize}

These are three valid but conceptually different answers to ``what are the designs?''; \algname finds all three, while ancestral sampling produces a duplicate.

\clearpage
\subsection{Math Reasoning (GSM8K)}
\label{app:qualitative:math-reasoning}

\paragraph{Example 1: Freelance blogger.}
Prompt: \emph{``Meredith is a freelance blogger who writes about health topics... A blog article takes an average of 4 hours to research and write about. Last week, she wrote 5 articles on Monday and 2/5 times more articles on Tuesday than on Monday. On Wednesday, she wrote twice the number of articles she wrote on Tuesday. Calculate the total number of hours she spent writing articles in the three days.''}

The phrase ``2/5 times more'' is linguistically ambiguous.

\textbf{Ancestral sampling} predominantly produces one interpretation:
\begin{itemize}[nosep,leftmargin=*]
\item ``Tuesday: 5 + (2/5)$\times$5 = 7 articles $\to$ Wednesday: 14 $\to$ Total: 104 hours''
\item (7 of 10 samples follow this pattern with minor formatting differences)
\end{itemize}

\textbf{\algname} finds both interpretations:
\begin{itemize}[nosep,leftmargin=*]
\item ``Tuesday: 5 + 2 = 7 articles $\to$ 104 hours'' \emph{(additive: ``2/5 more'' = add 2/5 of Monday)}
\item ``Tuesday: (2/5)$\times$5 = 2 articles $\to$ 44 hours'' \emph{(multiplicative: ``2/5 times'' = 2/5 as many)}
\end{itemize}

Even when the model's mode favors one interpretation, the DPP enhances coverage of both readings, which is valuable for pass@$\K$.

\paragraph{Example 2: Dog walking.}
Prompt: \emph{``John takes care of 10 dogs. Each dog takes .5 hours a day to walk and take care of their business. How many hours a week does he spend taking care of dogs?''}

Both methods arrive at the correct answer (35 hours), but \algname produces a much wider range of formatting and presentation styles.

\textbf{Ancestral sampling} produces uniform structure:
\begin{itemize}[nosep,leftmargin=*]
\item All 10 samples follow the same two-step pattern with minor wording variations
\item Consistent formatting: numbered steps, same answer format
\end{itemize}

\textbf{\algname} produces format outliers alongside standard solutions:
\begin{itemize}[nosep,leftmargin=*]
\item Standard prose explanation
\item Bullet points with inline math: ``$10 \times 0.5 = 5$ hours/day''
\item Heavy \LaTeX{} display math formatting
\item Russian-language output (Cyrillic ``Answer: 35.0'')
\item Placeholder template with no actual computation
\item Non-standard answer formats: \texttt{\$\$35\$\$}, \texttt{***35***}
\end{itemize}

Some outliers (placeholder templates, wrong language) are failure modes, but this diversity can aid pass@$\K$ when downstream systems parse outputs flexibly. It illustrates how the DPP explores the full space of model behaviors rather than collapsing to the mode.

\clearpage
\section{Notation and Terminology}
\label{app:notation}
\paragraph{Conventions.}
Strings are bold italic ($\yy$, $\boldsymbol{Y}$); vectors and matrices are bold upright ($\embed(\yy)$, $\Embs_\set$, $\LL$, $\mathbf{I}$); a set of strings is $\set$. Where a random string or set must be told apart from a fixed one of the same name, the random one is capital and bold italic ($\boldsymbol{Y}_j$ vs.\ $\yy_j$, $\boldsymbol{S}$ vs.\ $\set$), which also keeps it apart from matrices; elsewhere, the law under which a quantity is drawn ($\yy^{(n)} \sim \proposal$, $\omega \sim \mu$) says what is random. Each symbol names one object; a baseline's hyperparameters carry the baseline's name ($\lambda_{\mathrm{MMR}}$, $w_{\mathrm{DetBS}}$). Each accent has one meaning: a hat marks a quantity computed from a pool or by Monte Carlo ($\Bapprox$, $\Capprox$, $\Lapprox$, $\Zhat$, $\pihat$, $\hat w_n$, $\widehat V$); a bar marks a law averaged over the pool ($\pibar$); a tilde marks the naive sampler of \cref{rem:rao-blackwell} and its estimate ($\widetilde{f}$). Expectations, variances, and probabilities carry their random variable or law as a subscript ($\E_\omega$, $\Var_{\proposal^{\otimes \overlap}}$, $\Pr_{\Lapprox}$), and sums carry their domain; generic probability facts stated for arbitrary random variables are the exception. Index letters are fixed: $t$ is a token position, $n$ and $n'$ are pool positions, $j$ indexes the members of a $\K$-tuple, $d$ and $d'$ index eigenvectors or coordinates of $\R^D$, and $m$ indexes repeated sets. $\plm$ is the language model and $\proposal$ the proposal. $T$ is the global temperature in $\quality = \plm^{1/T}$, and $T'$ is the local sampling temperature, at which a pool is drawn.

\paragraph{Notation.}\leavevmode\par
\begingroup
\small
\begin{longtable}{@{}lp{0.52\linewidth}l@{}}
\toprule
symbol & meaning & defined \\
\midrule
\endhead
\multicolumn{3}{@{}l}{\emph{Strings and language models}} \\*
$\alphabet$, $\strings$, $\eos$ & alphabet, strings over it, end-of-string marker & \cref{sec:background} \\
$\yy$, $\yyPast$, $\abs{\yy}$ & string, its prefix before position $t$, its length & \cref{sec:background} \\
$\plm$, $\plmPrefix$ & language model, prefix probability & \cref{sec:background} \\
$\xx$, $\plm(\yy \mid \xx)$ & prompt, language model conditioned on it & \cref{sec:background} \\
$\localTmodel$ & locally annealed language model at temperature $T'$ & \cref{eq:local-temp-string} \\
$\globalTmodel$ & globally annealed language model at temperature $T$ & \cref{eq:global-temp} \\
$T$ & global temperature, $\quality = \plm^{1/T}$ & \cref{eq:quality-temp} \\
$T'$ & local sampling temperature, at which a pool is drawn & \cref{eq:local-temp} \\
\midrule
\multicolumn{3}{@{}l}{\emph{Determinantal point processes}} \\*
$\universe$ & base set & \cref{sec:dpp} \\
$\quality$ & quality & \cref{eq:quality-diversity} \\
$\embed$, $D$ & unit-norm embedding, its dimension & \cref{sec:dpp} \\
$\LL$, $\LL_\set$ & kernel, its submatrix indexed by $\set$ & \cref{eq:quality-diversity} \\
$\Embs_\set$ & embeddings of $\set$ as columns & \cref{sec:dpp} \\
$\set$, $\K$ & set, its size & \cref{eq:k-dpp} \\
$\pDPP$, $\normalizer$ & target $\K$-DPP, its normalizer & \cref{eq:k-dpp} \\
$\BB$, $\CC$ & feature matrix, dual kernel & \cref{eq:dual-kernel} \\
$e_k$ & elementary symmetric polynomial & \cref{app:dual-sampler}, footnote \labelcref{esp} \\
\midrule
\multicolumn{3}{@{}l}{\emph{Pool-based inference}} \\*
$\proposal$ & proposal & \cref{sec:algorithm} \\
$\quality \ll \proposal$ & $\proposal(\yy) = 0 \implies \quality(\yy) = 0$ (absolute continuity) & footnote \labelcref{fn:ac} \\
$N$, $\U$, $\yy^{(n)}$ & pool size, pool, its $n$th string & \cref{sec:algorithm} \\
$w$ & importance weight $\quality/\proposal$ & \cref{eq:B-matrix} \\
$\Bapprox$, $\Capprox$ & pool feature matrix, pool dual kernel & \cref{eq:B-matrix,eq:c_hat} \\
$\Lapprox$ & pool kernel $\Bapprox^\top\Bapprox$ & \cref{eq:c_hat} \\
$[N]$, $I$ & pool indices, a set of pool indices & \cref{eq:PoolDPP-fhat} \\
$\Pr_{\Lapprox}$ & law of the index set drawn by \cref{alg:dual-kdpp} & \cref{app:proof-pushforward} \\
$\set^\star$, $\set^\dagger$ & best set in the pool, greedy set & \cref{sec:algorithm} \\
\midrule
\multicolumn{3}{@{}l}{\emph{Analysis}} \\*
$f$, $\mathcal{D}$ & unnormalized target, its domain & \cref{sec:analysis} \\
$\mu$, $\omega$ & law of the sketch's randomness, its draw (the pool) & \cref{sec:analysis} \\
$\fhatom$, $\Zhat$, $\pihat$ & sketched weight, normalizer, distribution & \cref{sec:analysis} \\
$\pibar$ & marginal output law & \cref{eq:marginal-output-law} \\
$\gamma$ & $\Pr_\omega[\Zhat > 0]$ & \cref{app:redraw} \\
$\alphaMK$, $\falling{N}{\K}$ & collision constant, falling factorial & \cref{eq:PoolDPP-fhat} \\
$\incl{\overlap}$, $\overlap$ & inclusion ratio, overlap & \cref{eq:incl-def} \\
$a_\overlap$ & probability that two index sets overlap in $\overlap$ positions & \cref{eq:zfree} \\
$\rho$ & second moment of $\incl{1}$ & \cref{eq:incl-def} \\
$\theta$ & closed-form bound & \cref{thm:zfree} \\
$C_0$ & constant, set by the spectrum of $\CC$, with $\incl{1} \le C_0 w$ & \cref{sec:analysis} \\
$\pi_1$ & law of a uniformly chosen member of a target set & \cref{sec:analysis} \\
$\chi^2$, $\KL$ & $\chi^2$ and Kullback--Leibler divergences & \cref{thm:var-kl-mixture} \\
$\Ptilde$ & joint law of the pool and the output set & \cref{app:proof-var-kl-mixture} \\
$\widetilde{f}$ & weight estimate of the naive sampler & \cref{rem:rao-blackwell} \\
$\widehat V$ & Monte Carlo estimate of $\Var_\omega(\Zhat)/\normalizer^2$ & \cref{app:zfree-eval} \\
\midrule
\multicolumn{3}{@{}l}{\emph{Experiments}} \\*
$M$ & number of sets per prompt & \cref{eq:intrinsic-metrics} \\
$N'$ & larger pool size & \cref{tab:weights} \\
$\hat w_n$ & self-normalized weight of the $n$th draw & \cref{app:weights} \\
$\ell$ & normalized log-probability in MMR & \cref{app:baselines} \\
\bottomrule
\end{longtable}
\endgroup

\paragraph{Terminology.}\leavevmode\par
\begingroup
\small
\begin{longtable}{@{}lp{0.44\linewidth}l@{}}
\toprule
term & meaning & defined \\
\midrule
\endhead
ancestral sampling & sampling one token at a time from the conditionals & \cref{sec:background} \\
local, global annealing & tempering each conditional, or the whole string & \cref{eq:local-temp,eq:global-temp} \\
local sampling temperature & $T'$, the temperature at which a pool is drawn & \cref{eq:local-temp} \\
global temperature & $T$ in $\quality = \plm^{1/T}$ & \cref{sec:dpp} \\
embedding, embedder & unit-norm vector $\embedy$ of a string, the model that computes it & \cref{sec:dpp} \\
quality-scaled embedding & $\sqrt{\quality(\yy)} \embedy$, a column of $\BB$ & \cref{eq:quality-diversity} \\
DPP, $\K$-DPP & distribution over sets $\propto \det(\LL_\set)$, of size $\K$ & \cref{sec:dpp} \\
dual kernel & $D \times D$ kernel with the same nonzero spectrum & \cref{eq:dual-kernel} \\
pool, proposal & $N$ i.i.d.\ strings drawn from $\proposal$ & \cref{sec:algorithm} \\
pool $\K$-DPP & $\K$-DPP over the pool indices with kernel $\Lapprox$ & \cref{eq:c_hat} \\
importance weight & $w = \quality/\proposal$ & \cref{eq:B-matrix} \\
\algname & $\K$-DPP inference on the importance-weighted pool & \cref{sec:algorithm} \\
MAP & maximum a posteriori: the most probable set & \cref{sec:algorithm} \\
\algnameSamp, \algnameMAP & exact sample, greedy MAP set from the pool $\K$-DPP & \cref{sec:algorithm} \\
sketch, sketched distribution & random weight function and its normalization & \cref{sec:analysis} \\
marginal output law & law of one run's output set & \cref{eq:marginal-output-law} \\
well-behaved & unbiased weights with finite second moment & \cref{sec:analysis} \\
fallback & output rule when $\Zhat = 0$ & \cref{app:redraw} \\
inclusion ratio & target tuple law over proposal tuple law & \cref{eq:incl-def} \\
relative variance & $\Var_\omega(\Zhat)/\normalizer^2$ & \cref{sec:analysis} \\
$U$-statistic & average of a symmetric function over subsets & \cref{sec:analysis} \\
pool-built instance & base set of pooled strings, proposal their frequencies & \cref{app:zfree-eval} \\
quality, diversity (intrinsic) & mean log-probability, log-det of the Gram matrix & \cref{eq:intrinsic-metrics} \\
recall, precision & gold interpretations covered, queries matching a gold result & \cref{sec:ambrosia} \\
pass@$\K$, per-chain accuracy & any chain correct, fraction of chains correct & \cref{sec:gsm8k} \\
naive DPP, $\K$-means, \algnameLLM & pool baselines & \cref{sec:experiments} \\
DivBS, DetBS, SBS, & decoding and reranking baselines & \cref{sec:experiments} \\
\quad CPSBS, MBR, MMR & & \\
ESS & effective sample size of the weights & \cref{app:weights} \\
pool coverage & fraction of gold interpretations in a pool & \cref{app:pool-ceiling} \\
plain, instructed kernel & embedder without, with a task instruction & \cref{app:embedder} \\
\bottomrule
\end{longtable}
\endgroup

\clearpage
\section*{NeurIPS Paper Checklist}

\begin{enumerate}

\item {\bf Claims}
    \item[] Question: Do the main claims made in the abstract and introduction accurately reflect the paper's contributions and scope?
    \item[] Answer: \answerYes{}
    \item[] Justification: The abstract and \cref{sec:intro} state three claims, each substantiated in the body: the pool-based DPP sampler with two inference modes (\cref{sec:algorithm}), a finite-sample $\bigo(1/N)$ KL bound with an exact variance identity (\cref{thm:zfree}), and quality--diversity frontier gains on FineWeb (from \algnameMAP) and Ambrosia, together with higher diversity at about the same pass@$\K$ on GSM8K, with and without an instructed embedder (\cref{sec:experiments,app:embedder}).

\item {\bf Limitations}
    \item[] Question: Does the paper discuss the limitations of the work performed by the authors?
    \item[] Answer: \answerYes{}
    \item[] Justification: The Conclusion collects the limitations in one paragraph, and each is discussed where it arises. The LM-hidden-state embedding variant does not consistently improve over off-the-shelf embedders (\cref{sec:dpp}). On GSM8K, neither the baseline nor the instructed embedder lets either \algname mode improve pass@$\K$ at matched temperature (\cref{sec:gsm8k,app:embedder}). The weights of \algnameSamp have an effective sample size below $\K$ for most prompts on FineWeb and GSM8K away from $T{=}1$ (\cref{sec:ablations}). The bound is evaluated only on instances built from our pools, where it is vacuous on GSM8K at the pool size we run (\cref{sec:analysis}). Achievable diversity is bounded by the pool's coverage, and the MAP mode is a greedy approximation to the pool's mode, not the population's. The dominant cost is pool construction: drawing $N$ sequences from the LM is much more expensive than the subsequent selection (\cref{tab:efficiency-fixed}). The paper's scope is set at the sequence level, to ground a principled diversity objective with finite-sample guarantees; we leave further extensions to future work.

\item {\bf Theory assumptions and proofs}
    \item[] Question: For each theoretical result, does the paper provide the full set of assumptions and a complete (and correct) proof?
    \item[] Answer: \answerYes{}
    \item[] Justification: \cref{thm:var-kl-mixture}, \cref{prop:PoolDPP-well-behaved}, \cref{thm:zfree}, and \cref{cor:kl-sum} state their assumptions inline; complete proofs are in \cref{app:PoolDPP-proofs}, and machine-checked Lean proofs are described in \cref{app:lean}.

    \item {\bf Experimental result reproducibility}
    \item[] Question: Does the paper fully disclose all the information needed to reproduce the main experimental results of the paper to the extent that it affects the main claims and/or conclusions of the paper (regardless of whether the code and data are provided or not)?
    \item[] Answer: \answerYes{}
    \item[] Justification: Pseudocode for both inference modes is provided in \cref{alg:dual-kdpp,alg:greedy-map}; all language models, embedding models, and datasets used are publicly available; pool sizes, subset sizes, temperature settings, and per-method parameter sweeps are reported in \cref{sec:experiments,sec:ablations}, and the code and the commands that reproduce each table and figure are released (next item).

\item {\bf Open access to data and code}
    \item[] Question: Does the paper provide open access to the data and code, with sufficient instructions to faithfully reproduce the main experimental results, as described in supplemental material?
    \item[] Answer: \answerYes{}
    \item[] Justification: All datasets used (FineWeb, Ambrosia, GSM8K) and all language and embedding models (Gemma-2-9B-Instruct, Qwen3-8B, Llama-3.1-8B-Instruct, MiniLM-L6, Qwen3-Embedding-0.6B) are publicly available under their original licenses. Our implementation of \algname, the experiment scripts, the outputs quoted in the paper, and the Lean proofs are released at \url{https://github.com/rycolab/pool-dpp}, whose README lists the command that produces each table and figure; pseudocode for the algorithms is also provided in \cref{alg:dual-kdpp,alg:greedy-map}.

\item {\bf Experimental setting/details}
    \item[] Question: Does the paper specify all the training and test details (e.g., data splits, hyperparameters, how they were chosen, type of optimizer) necessary to understand the results?
    \item[] Answer: \answerYes{}
    \item[] Justification: \cref{sec:experiments} specifies the language models, embedding models, datasets, subset size $\K$, pool size $N$, and metric definitions; the per-method parameter sweep traced by each Pareto curve is described inline (\cref{sec:intrinsic,sec:ambrosia,sec:gsm8k}); ablations over $N$, $\K$, the embedding/LM choice, and the global temperature $T$ are in \cref{sec:ablations}.

\item {\bf Experiment statistical significance}
    \item[] Question: Does the paper report error bars suitably and correctly defined or other appropriate information about the statistical significance of the experiments?
    \item[] Answer: \answerYes{}
    \item[] Justification: All Pareto figures display 95\% $t$-intervals over independently sampled pools ($10$ on FineWeb, $100$ on Ambrosia and GSM8K); the pool count, interval type, and quantity are stated in each caption and in \cref{sec:experiments}. Comparisons that support claims use paired sign-flip permutation tests over questions and over pools (\cref{app:significance}).

\item {\bf Experiments compute resources}
    \item[] Question: For each experiment, does the paper provide sufficient information on the computer resources (type of compute workers, memory, time of execution) needed to reproduce the experiments?
    \item[] Answer: \answerYes{}
    \item[] Justification: \cref{tab:efficiency-fixed} reports per-stage wall-clock time for $N{=}400$ and $\K{=}10$ on a single A100 GPU, and \cref{tab:efficiency-scaling} reports selection-time scaling with the pool size. Pool construction via vLLM dominates total runtime; the DPP selection step itself is two orders of magnitude cheaper.

\item {\bf Code of ethics}
    \item[] Question: Does the research conducted in the paper conform, in every respect, with the NeurIPS Code of Ethics \url{https://neurips.cc/public/EthicsGuidelines}?
    \item[] Answer: \answerYes{}
    \item[] Justification: The research uses publicly released language models and public-domain or appropriately licensed datasets, and it involves no human subjects. We have reviewed the NeurIPS Code of Ethics and find no conflict.

\item {\bf Broader impacts}
    \item[] Question: Does the paper discuss both potential positive societal impacts and negative societal impacts of the work performed?
    \item[] Answer: \answerNA{}
    \item[] Justification: This is a foundational, technical paper on inference-time sampling from pretrained language models. It has no direct path to a specific application or deployment, and the impacts would be inherited from whatever downstream system uses it. The paper does not introduce new societal risks beyond those of the underlying base models.

\item {\bf Safeguards}
    \item[] Question: Does the paper describe safeguards that have been put in place for responsible release of data or models that have a high risk for misuse (e.g., pre-trained language models, image generators, or scraped datasets)?
    \item[] Answer: \answerNA{}
    \item[] Justification: The paper introduces an inference-time sampling algorithm that operates on top of publicly released language models. It does not release a new pretrained model, scraped dataset, or generative system that would carry distinct misuse risks beyond those of the underlying base models.

\item {\bf Licenses for existing assets}
    \item[] Question: Are the creators or original owners of assets (e.g., code, data, models), used in the paper, properly credited and are the license and terms of use explicitly mentioned and properly respected?
    \item[] Answer: \answerYes{}
    \item[] Justification: All language models (Gemma-2-9B-Instruct, Qwen3-8B, Llama-3.1-8B-Instruct), embedding models (MiniLM-L6, Qwen3-Embedding-0.6B), datasets (FineWeb, Ambrosia, GSM8K), and baseline implementations (DivBS, DetBS) are cited with their original references in \cref{sec:experiments}. We use each asset under its original license.

\item {\bf New assets}
    \item[] Question: Are new assets introduced in the paper well documented and is the documentation provided alongside the assets?
    \item[] Answer: \answerYes{}
    \item[] Justification: The new assets are our code, the outputs quoted in the paper, and the Lean formalization of the theorems, released at \url{https://github.com/rycolab/pool-dpp}. The README documents installation and the command for each experiment, and the \texttt{lean/} and \texttt{baselines/detbs\_ssk/} folders have their own READMEs. The paper releases no new dataset or pretrained model; pseudocode and analysis for the proposed sampler are in \cref{alg:dual-kdpp,alg:greedy-map} and \cref{sec:analysis}.

\item {\bf Crowdsourcing and research with human subjects}
    \item[] Question: For crowdsourcing experiments and research with human subjects, does the paper include the full text of instructions given to participants and screenshots, if applicable, as well as details about compensation (if any)?
    \item[] Answer: \answerNA{}
    \item[] Justification: The paper does not involve crowdsourcing or research with human subjects.

\item {\bf Institutional review board (IRB) approvals or equivalent for research with human subjects}
    \item[] Question: Does the paper describe potential risks incurred by study participants, whether such risks were disclosed to the subjects, and whether Institutional Review Board (IRB) approvals (or an equivalent approval/review based on the requirements of your country or institution) were obtained?
    \item[] Answer: \answerNA{}
    \item[] Justification: The paper does not involve research with human subjects.

\item {\bf Declaration of LLM usage}
    \item[] Question: Does the paper describe the usage of LLMs if it is an important, original, or non-standard component of the core methods in this research? Note that if the LLM is used only for writing, editing, or formatting purposes and does \emph{not} impact the core methodology, scientific rigor, or originality of the research, declaration is not required.
    \item[] Answer: \answerYes{}%
    \item[] Justification: Two analyses use LLMs as components: Qwen3-8B is the LLM-based selection baseline (\cref{sec:ablations}), and Gemini 3.6 Flash is the judge that labels pairs of GSM8K chains as differing in answer, derivation, or surface only (\cref{app:embedder}); both are described with their settings and validated in the appendix. Our use of AI agents to edit and draft text, to double-check proofs, and to help write the Lean formalization is described in the AI-use disclosure at the end of the Conclusion. Pretrained language models are otherwise the object of study, not a component of the method.

\end{enumerate}

\end{document}